\documentclass{article}

\usepackage[preprint]{neurips_2026}

\usepackage[utf8]{inputenc}
\usepackage[T1]{fontenc}
\usepackage{microtype}

\usepackage{amsmath}              
\usepackage{amssymb}              
\usepackage{amsfonts}             
\usepackage{mathtools}            
\usepackage{nicefrac}             

\usepackage{fontawesome5}
\usepackage{booktabs}             
\usepackage{multirow}             
\usepackage{makecell}             
\usepackage{colortbl}             
\usepackage{capt-of}
\usepackage{arydshln}
\usepackage{wrapfig}  
\usepackage{tabularx}

\usepackage{graphicx}             
\usepackage{pifont}               

\usepackage[table]{xcolor}        

\definecolor{cvprblue}{rgb}{0.21,0.49,0.74}   

\definecolor{darkgreen}{HTML}{2E8B57}
\definecolor{darkred}{HTML}{D1191F}
\definecolor{myyellow}{HTML}{DAA520}           
\definecolor{downred}{HTML}{C0392B}            
\definecolor{upgreen}{HTML}{27AE60}            
\definecolor{graytext}{gray}{0.6}              

\definecolor{graybg}{gray}{0.86}               
\definecolor{naivebg}{gray}{0.95}              
\definecolor{experiencebg}{HTML}{EDF3FF}       
\definecolor{capabilitybg}{HTML}{EAF4E6}       
\definecolor{oursbg}{HTML}{F2EAF8}             

\definecolor{codeblue}{rgb}{0.13,0.29,0.53}
\definecolor{codegreen}{rgb}{0,0.6,0}
\definecolor{codegray}{rgb}{0.5,0.5,0.5}
\definecolor{codepurple}{rgb}{0.58,0,0.82}
\definecolor{codestring}{rgb}{0.64,0.08,0.08}
\definecolor{backcolour}{rgb}{0.95,0.95,0.92}

\definecolor{jsonKey}{rgb}{0.08, 0.45, 0.74}   
\definecolor{jsonVal}{rgb}{0.64, 0.08, 0.08}   
\definecolor{jsonNum}{rgb}{0.05, 0.60, 0.45}   
\definecolor{jsonBool}{rgb}{0.00, 0.00, 1.00}  

\definecolor{brandblue}{rgb}{0.34, 0.7, 1}

\usepackage{url}                  

\usepackage{enumitem}             
\usepackage{varwidth}             
\usepackage{latexsym}             

\usepackage{minitoc}            
\usepackage{titletoc}           
\mtcsettitle{minitoc}{Appendix}

\usepackage{listings}
\usepackage{inconsolata}          
\usepackage{textcomp}             

\lstdefinestyle{pythonstyle}{
    language=Python,
    commentstyle=\color{codegreen},
    keywordstyle=\color{codeblue}\bfseries,
    numberstyle=\tiny\color{codegray},
    stringstyle=\color{codestring},
    basicstyle=\ttfamily\small,
    breakatwhitespace=false,
    breaklines=true,
    captionpos=b,
    keepspaces=true,
    numbers=left,
    numbersep=5pt,
    showspaces=false,
    showstringspaces=false,
    showtabs=false,
    tabsize=4,
    frame=single,
}

\lstdefinestyle{promptstyle}{
    basicstyle=\ttfamily\small,
    breaklines=true,
    breakindent=0pt,
    showstringspaces=false,
    frame=none,
    numbers=none,
    keepspaces=true,
    columns=fullflexible,
    tabsize=2,
    escapeinside={(*}{*)},
    moredelim=**[is][\textcolor{blue}]{@}{@},
}

\lstdefinelanguage{json}{
    keywords={true,false,null},
    keywordstyle=\color{jsonBool}\bfseries,
    sensitive=true,
    string=[b]",
    stringstyle=\color{jsonVal},
    moredelim=[s][\color{jsonKey}]{"}{:},
    literate=
     *{0}{{{\color{jsonNum}0}}}{1}
      {1}{{{\color{jsonNum}1}}}{1}
      {2}{{{\color{jsonNum}2}}}{1}
      {3}{{{\color{jsonNum}3}}}{1}
      {4}{{{\color{jsonNum}4}}}{1}
      {5}{{{\color{jsonNum}5}}}{1}
      {6}{{{\color{jsonNum}6}}}{1}
      {7}{{{\color{jsonNum}7}}}{1}
      {8}{{{\color{jsonNum}8}}}{1}
      {9}{{{\color{jsonNum}9}}}{1}
      {:}{{:}}{1}
      {,}{,}{1}
      {\{}{{{\color{black}\{}}}{1}
      {\}}{{{\color{black}\}}}}{1}
      {[}{{{\color{black}[}}}{1}
      {]}{{{\color{black}]}}}{1},
}

\lstdefinestyle{json}{
    language=json,
    basicstyle=\ttfamily\small,
    breaklines=true,
    showstringspaces=false,
    frame=single,
    rulecolor=\color{black!30},
    framesep=10pt,
    frameround=tttt,
    numbers=none,
    keepspaces=true,
    columns=fullflexible,
    upquote=true,
}

\lstdefinelanguage{jsonMemory}{
    keywords={true,false,null},
    keywordstyle=\color{black}\bfseries,
    sensitive=true,
    string=[b]",
    stringstyle=\color{black},
    moredelim=[s][\color{black}]{"}{:},
    literate=
      {:}{{:}}{1}
      {,}{,}{1}
      {\{}{{{\color{black}\{}}}{1}
      {\}}{{{\color{black}\}}}}{1}
      {[}{{{\color{black}[}}}{1}
      {]}{{{\color{black}]}}}{1},
}

\lstdefinestyle{memory}{
    language=jsonMemory,
    basicstyle=\ttfamily\small\color{black},
    breaklines=true,
    showstringspaces=false,
    frame=single,
    rulecolor=\color{black!30},
    framesep=10pt,
    frameround=tttt,
    numbers=none,
    keepspaces=true,
    columns=fullflexible,
    upquote=true,
    escapeinside={(*}{*)},
}

\usepackage{tcolorbox}
\tcbuselibrary{breakable}

\tcbset{
  base/.style={
    arc=0mm,
    bottomtitle=-0.25mm,
    boxrule=0mm,
    colbacktitle=black!10!white,
    coltitle=black,
    fonttitle=\bfseries,
    left=2.5mm,
    leftrule=1mm,
    right=3.5mm,
    title={#1},
    toptitle=0.25mm,
    breakable,
  }
}

\newtcolorbox{mybox}[1]{
  colframe=brandblue,
  base={#1 \hfill \hyperlink{appendixtoc}{\footnotesize\fbox{Back to ToC}}}
}

\newcommand{\yes}{\textcolor{darkgreen}{\ding{51}}}
\newcommand{\no}{\textcolor{darkred}{\ding{55}}}
\newcommand{\half}{\textcolor{myyellow}{\ding{52}\rotatebox[origin=c]{-9.2}{\kern-0.7em\ding{55}}}}

\newcommand*{\imgintext}[1]{%
  \raisebox{-0.2ex}{%
    \includegraphics[height=2ex,keepaspectratio]{#1}%
  }%
}

\newcommand{\circnum}[1]{\raisebox{-0.5pt}{\ding{\numexpr171+#1\relax}}}

\newcommand{\appsection}[1]{%
  \subsection[#1]{#1 \hfill \hyperlink{appendixtoc}{\footnotesize\fbox{Back to ToC}}}%
}

\usepackage{pifont}
\newcommand{\cmark}{\ding{51}}
\newcommand{\xmark}{\ding{55}}

\newcommand{\modelicon}[2]{%
  \IfFileExists{figures/icons/#2}{%
    \raisebox{-0.2\height}{\includegraphics[height=#1]{figures/icons/#2}}\,%
  }{}%
}

\usepackage{algorithm}
\usepackage{algpseudocode}
\algrenewcommand{\algorithmiccomment}[1]{\hfill\textcolor{cvprblue}{\texttt{//}~#1}}
\newcommand{\LineComment}[1]{\State \textcolor{cvprblue}{\texttt{//}~#1}}

\usepackage{tcolorbox}
\tcbuselibrary{breakable,skins,listings}
\usepackage{booktabs}
\usepackage{tabularx}
\usepackage{array}
\usepackage{float}
\usepackage{enumitem}
\usepackage{listings}
\providecolor{cvprblue}{rgb}{0.21,0.49,0.74}

\usepackage[pagebackref,breaklinks,colorlinks,allcolors=cvprblue]{hyperref}

\newcommand{\OURMETHOD}{Terminal-World}

\NewDocumentCommand{\hongru}{ mO{} }{}

\title{\OURMETHOD: Scaling Terminal-Agent Environments via Agent Skills}

\author{%
Zihao Cheng\textsuperscript{1, *} \quad
Hongru Wang\textsuperscript{2, *} \quad
Zeming Liu\textsuperscript{1, $\dagger$} \quad
\textbf{Xinyi Wang}\textsuperscript{2} \quad
\textbf{Xiangrong Zhu}\textsuperscript{2} \\
\textbf{Yuhang Guo}\textsuperscript{3} \quad
\textbf{Wei Lin}\textsuperscript{2} \quad
\textbf{Jeff Z. Pan}\textsuperscript{4} \quad
\textbf{Yunhong Wang}\textsuperscript{1} \\
\textsuperscript{1}School of Computer Science and Engineering, Beihang University, Beijing, China\\ 
\textsuperscript{2}Independent Researcher \quad
\textsuperscript{3}Beijing Institute of Technology \quad
\textsuperscript{4}University of Edinburgh 
\\
\textsuperscript{*}Equal contribution \quad
\textsuperscript{$\dagger$}Corresponding author \quad
Email: \texttt{\{zihaocheng, zmliu\}@buaa.edu.cn}
}

\begin{document}

\startcontents[global]   
\maketitle

\begin{abstract}
Terminal agents extend Large Language Models with the ability to execute tasks directly in command-line environments, but their progress is bottlenecked by the scarcity of high-quality training data. Existing approaches bootstrap from \textbf{partial sources} such as human-defined seeds or GitHub repositories to instantiate one component and then complete the rest, producing tasks confined to narrow seed distributions, environments misaligned with task semantics, and inefficient trajectories from unguided exploration. To address these limitations, we introduce \textbf{\OURMETHOD}, a fully automated pipeline that uses \textbf{agent skills} as the central synthesis primitive, which jointly encode \textit{what} to accomplish, \textit{when} to apply (preconditions and environment state), and \textit{how} to execute, enabling task instructions, environments, and teacher trajectories to be co-derived. To further broaden the synthesis space, \OURMETHOD\ composes skills into \textbf{skill teams} and \textbf{skill graphs} for multi-role and cross-domain task synthesis. Using this pipeline, we construct \textbf{5,723} training environments and train \textbf{\OURMETHOD-8B/14B/32B}, evaluated across 6 benchmarks where the \OURMETHOD\ series consistently outperforms terminal-agent baselines. Notably, using the same teacher model and only \textbf{1.2\%} of the training data, \OURMETHOD-32B surpasses Nemotron-Terminal-32B on Terminal-Bench 2.0 by \textbf{+4.5} Pass@1 (31.5) and achieves 43.8 Pass@3.
\end{abstract}

\vspace{-2.7mm}
\begin{figure}[h]
    \centering
    \includegraphics[width=\linewidth]{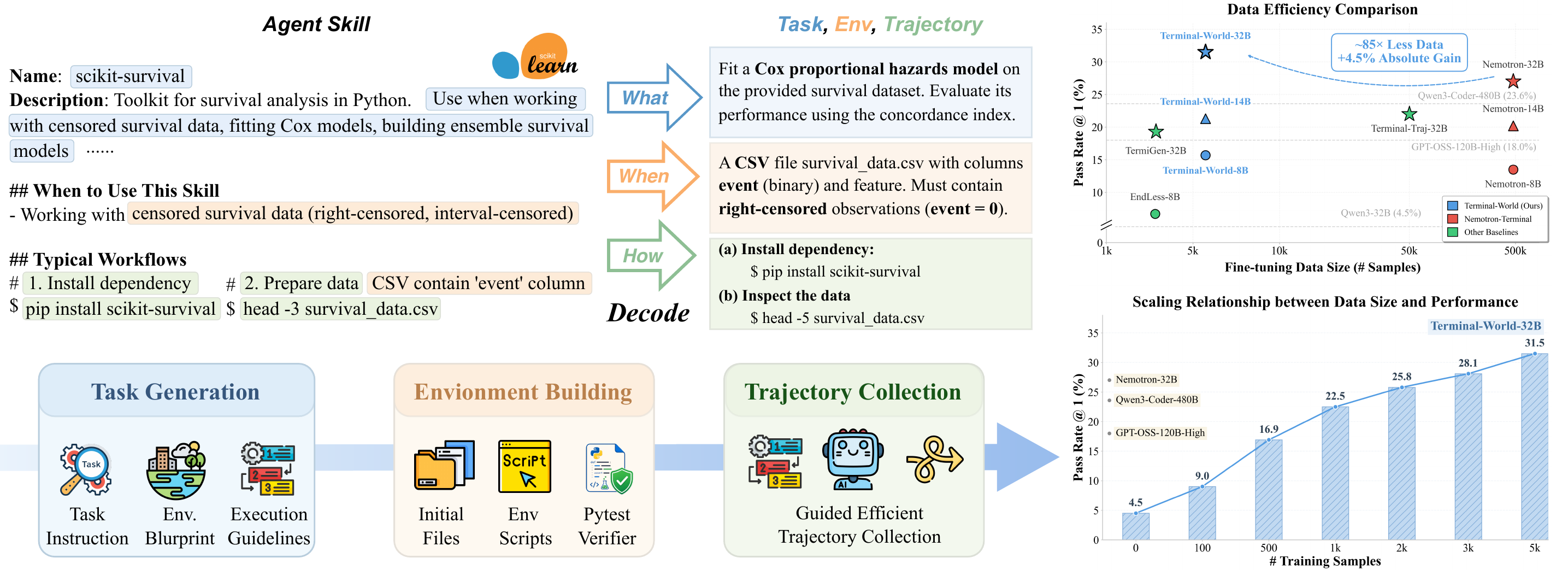}
    \caption{\textbf{Overview of \OURMETHOD~(left) and agent performance (right).}
    \OURMETHOD~uses agent skills as the synthesis primitive for terminal-agent data construction. Each skill encodes \textcolor[HTML]{5897C8}{\textit{\textbf{what}}} the agent should accomplish, \textcolor[HTML]{FAAE6B}{\textit{\textbf{when}}} the skill should be applied, and \textcolor[HTML]{8EBF94}{\textit{\textbf{how}}} the task should be executed. By decoding these three aspects, \OURMETHOD~top-down synthesizes \textcolor[HTML]{5897C8}{\textbf{task instructions}}, \textcolor[HTML]{FAAE6B}{\textbf{environments}}, and \textcolor[HTML]{8EBF94}{\textbf{trajectory}} in a unified process. With \textbf{85$\times$} less data than Nemotron-Terminal, \OURMETHOD~achieves a \textbf{4.5\%} absolute improvement on Terminal-Bench 2.0.}    \label{fig:intro}
    \vspace{-3mm}
\end{figure}
\section{Introduction}

LLM-based agents are increasingly moving from predefined API calls~\citep{gorilla,apigen,toolace,toolbridge,apigen_mt,toolmind,inftool,agentworld} to direct terminal operation. Systems such as Claude Code~\citep{claudecode} and Codex~\citep{codex} issue shell commands inside real execution environments, replacing fixed tool schemas with a compositional action space that affords substantially greater generality~\citep{harness_survey,bui2026building} and autonomy~\citep{toa}.

Despite this potential, the progress of terminal agents is fundamentally bottlenecked by the scarcity of high-quality training data. Unlike API-based agents, which simply select and parameterize predefined tools~\citep{toolsurvey}, terminal agents operate within real file systems and runtime environments~\citep{tb2,endless}. Each training example jointly specifies a task instruction, an executable environment with initial files, dependencies, and system configurations, and a high-quality multi-turn trajectory. The tight interdependence among these components makes manually curating such data prohibitively expensive and difficult to scale, motivating a growing line of work on automated terminal-agent data synthesis.

Existing methods synthesize terminal-agent data by starting from \textbf{partial sources}, such as human-defined seed data (i.e., manually specified keywords or short descriptors)~\citep{endless,termigen,nemotronterminal} or GitHub repositories~\citep{terminaltraj}, to instantiate one component of the data, and then rely on LLMs to complete the rest. Although this paradigm can synthesize terminal-agent data, it still suffers from three key limitations: \textbf{(1) Limited Tasks}: tasks are directly converted from human-defined seeds or repositories, resulting in a constrained distribution that fails to capture the diverse requirements and complexity of real-world tasks; \textbf{(2) Environment Misalignment}: task semantics and execution environments are not jointly specified from the beginning, so environments are retrofitted around tasks, producing configurations that are fragile or only loosely aligned with the intended task; \textbf{(3) Trajectory Inefficiency}: without explicit procedural guidance, teacher models often rely on autonomous exploration to solve each sandbox, producing trajectories with redundant exploration, suboptimal solution paths, and strong dependence on the teacher's intrinsic terminal-solving capability.

Our key observation is that a natural synthesis primitive for terminal-agent data already exists in open-source ecosystems: \textbf{agent skills}~\citep{skillrl, skill0}, such as those collected in ClawHub~\citep{clawhub} and SkillMP~\citep{skillsmp}, which are human-authored guidance packages that encapsulate authentic terminal workflows distilled from real practice. As illustrated in Figure~\ref{fig:intro} (Left), each skill jointly encodes three aspects of an end-to-end terminal task: \textit{\circnum{1} what should be accomplished}, \textit{\circnum{2} when the skill should be applied} (i.e., the preconditions, inputs, and environmental state required for execution), and \textit{\circnum{3} how it should be executed}. An agent skill thus constitutes a pre-aligned specification of task semantics, environmental constraints, and execution procedure, directly addressing the three limitations mentioned above.

Building on this primitive, we introduce \textbf{\OURMETHOD}, a fully automated pipeline that orchestrates a multi-agent architecture to instantiate each agent skill as a unified \textit{task instruction--executable environment--teacher trajectory} triple. To further scale the synthesis space, \OURMETHOD~extends individual skills into \textbf{agent skill teams} and \textbf{agent skill graphs}, enabling more complex multi-role and cross-domain task synthesis. To broaden the usage scenarios of each skill, \OURMETHOD~pairs skills with user personas~\citep{personahub}, enabling the same underlying ability to be instantiated across diverse user backgrounds, goals, and preferences. Using \OURMETHOD, we construct \textbf{5,723} high-fidelity terminal-agent training environments and collect teacher trajectories with DeepSeek-V3.2 at an average cost of only \textbf{\$0.17}, demonstrating the efficiency of our automated construction harness. We further train a family of models, \textbf{\OURMETHOD-8B/14B/32B}. Across 6 benchmarks, the \OURMETHOD~series consistently outperforms existing terminal-agent baselines at comparable model scales. Notably, using the same teacher model and only \textbf{1.2\%} of the training data, \OURMETHOD-32B surpasses Nemotron-Terminal-32B~\citep{nemotronterminal} on Terminal-Bench 2.0 by \textbf{+4.5} Pass@1 (31.5) and achieves 43.8 Pass@3. It also exhibits more efficient task-execution behavior (Sec.~\ref{sec:behavior}), requiring fewer steps and commands while maintaining lower command-failure rates. These results demonstrate that our pipeline produces diverse, high-quality terminal environments and effective trajectories at low cost. Overall, our contributions are summarized as follows:

\begin{itemize}[leftmargin=*]
    \item We propose \textbf{\OURMETHOD}, a fully automated synthesis pipeline that uses \textbf{agent skills} as the central \textbf{synthesis primitive} to jointly drive task instruction synthesis, environment construction, and teacher trajectory collection.

    \item Using \OURMETHOD, we construct \textbf{5,723} high-fidelity terminal-agent training environments, each paired with a skill-guided teacher trajectory, and train a family of models \textbf{\OURMETHOD-8B/14B/32B} on this data.

    \item Extensive experiments across 6 benchmarks show that the \OURMETHOD~series outperforms existing terminal-agent baselines. Notably, with the same teacher model and only \textbf{1.2\%} of the training data, \OURMETHOD-32B surpasses Nemotron-Terminal-32B on Terminal-Bench 2.0 by \textbf{+4.5} Pass@1 (31.5) and achieves 43.8 Pass@3.
\end{itemize}
\section{Related Work}

\definecolor{rowblue}{RGB}{235, 245, 255}

\begin{table*}[t]
\centering
\caption{\textbf{Comparison of existing datasets.}
\textbf{Align.} indicates whether task semantics and environments are jointly designed rather than post-hoc adapted.
\textbf{Open File Space} indicates support for arbitrary file types in environments.
\textbf{Exec. Verif.} indicates whether task completion can be verified by executing evaluation scripts.
\textbf{Sandbox}: \imgintext{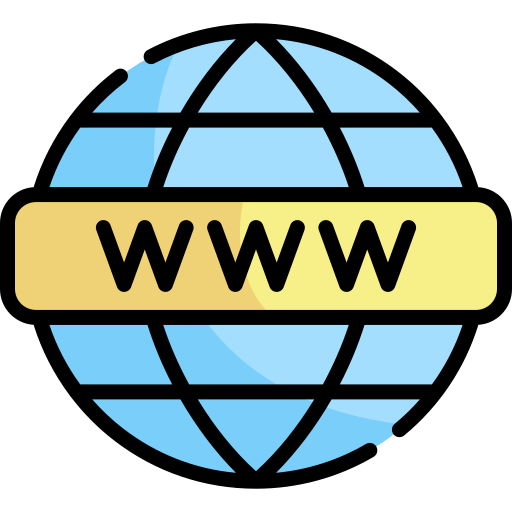} Web search, \imgintext{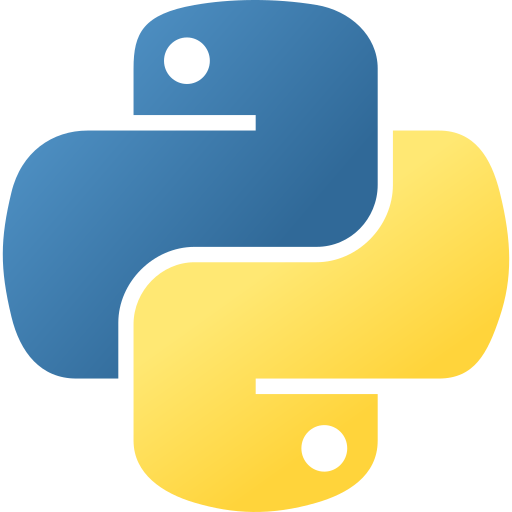} Python, \imgintext{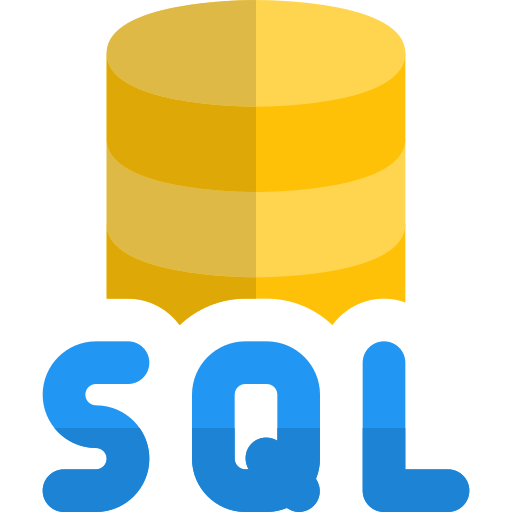} SQL engine, and \imgintext{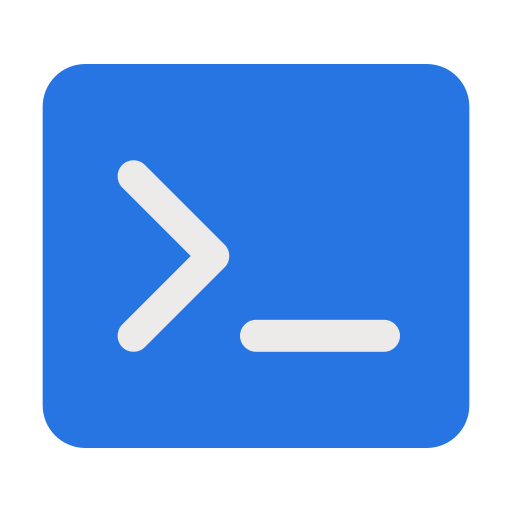} Terminal.
\textbf{Tool Space} is \textit{Fixed} for predefined toolsets and \textit{Open} for extensible tool spaces.
\textbf{Teacher Source} indicates whether trajectories are generated through self-solving or guided by additional structured guidelines.}
\label{tab:comparison}
\resizebox{\textwidth}{!}{%
\setlength{\tabcolsep}{3.0pt}
\renewcommand{\arraystretch}{1.02}
\begin{tabular}{@{}l >{\centering\arraybackslash}p{2.25cm} c cccccc ccc cc@{}}\toprule[0.08em]
\multirow{2}{*}{\textbf{Dataset}} &
  \multicolumn{2}{c}{\textbf{Task}} &
  \multicolumn{6}{c}{\textbf{Environment}} &
  \multicolumn{3}{c}{\textbf{Tool}} &
  \multicolumn{2}{c}{\textbf{Trajectory}} \\
  \cmidrule(lr){2-3} \cmidrule(lr){4-9} \cmidrule(lr){10-12} \cmidrule(l){13-14}
&
  \textbf{Primitive} &
  \textbf{Human Free} &
  \textbf{Align.} &
  \textbf{No Pre-def.} &
  \textbf{Real World} &
  \textbf{Open File Space} &
  \textbf{Exec. Verif.} &
  \textbf{Sandbox} &
  \textbf{Tool Gen.} &
  \textbf{Tool Space} &
  \textbf{\# Tools} &
  \textbf{Teacher Source} &
  \textbf{\# Traj.} \\

\midrule[0.05em]

Gorilla~\citep{gorilla} &
  API specs & \no &
  -- & \no & \no & \no & \no & -- &
  \no & Fixed & 1,645 &
  Self & 16,450 \\

ToolBridge~\citep{toolbridge} &
  API specs & \no &
  -- & \no & \no & \no & \yes & \imgintext{figures/python.png} &
  \no & Fixed & $\infty$ &
  Self & 178,023 \\

APIGen~\citep{apigen} &
  API specs & \yes &
  -- & \no & \yes & \no & \yes & \imgintext{figures/python.png} &
  \no & Fixed & 3,673 &
  Self & 60,000 \\

WebExplorer~\citep{webexplorer} &
  Web entities & \yes &
  -- & \yes & \yes & \no & \no & \imgintext{figures/web.png} &
  \no & Fixed & 2 &
  Self & 13,000 \\

ProgSearch~\citep{progsearch} &
  Web entities & \yes &
  -- & \yes & \yes & \no & \no & \imgintext{figures/web.png} \texttt{+} \imgintext{figures/python.png} &
  \yes & Fixed & 3 &
  Guided & 5,500 \\

Aseacher~\citep{asearcher} &
  Web entities & \no &
  -- & \no & \yes & \no & \no & \imgintext{figures/web.png} &
  \no & Fixed & 2 &
  Self & 35,000 \\

ToolACE~\citep{toolace} &
  API specs & \no &
  --& \no & \no & \no & \no & -- &
  \no & Fixed & 26,507 &
  Guided & 11,300 \\

ToolMind~\citep{toolmind} &
  API specs & \yes &
  -- & \no & \no & \no & \no & -- &
  \no & Fixed & 20,000 &
  Guided & 111,941 \\

InfTool~\citep{inftool} &
  API specs & \yes &
  -- & \no & \no & \no & \no & -- &
  \no & Fixed & 3,059 &
  Self & 4,965 \\

DataMind~\citep{datamind} &
  Data files & \half &
  \yes & \yes & \yes & \no & \yes & \imgintext{figures/python.png} \texttt{+} \imgintext{figures/sql.png} &
  \yes & Open & $\infty$ &
  Guided & 11,707 \\

APIGen-MT~\citep{apigen_mt} &
  API specs & \yes &
  -- & \no & \no & \no & \yes & -- &
  \no & Fixed & 28 &
  Guided & 5,000 \\

TaskCraft~\citep{taskcraft} &
  Seeds & \no &
  \yes & \no & \no & \no & \no & \imgintext{figures/python.png} &
  \no & Open & $\infty$ &
  Guided & 36,000 \\

GEM~\citep{gem} &
  Raw text & \yes &
  \yes & \yes & \no & \no & \no & -- &
  \no & Open & $\infty$ &
  Guided & 10,000 \\

Endless Terminal~\citep{endless} &
  Seeds & \no &
  \no & \yes & \yes & \half & \yes & \imgintext{figures/bash.png} &
  \yes & Open & 420 &
  -- & -- \\

TermiGen~\citep{termigen} &
  Seeds & \no &
  \no & \yes & \yes & \half & \yes & \imgintext{figures/bash.png} &
  \yes & Open & 420 &
  Guided & 3,291 \\

Nemotron-Terminal~\citep{nemotronterminal} &
  Seeds & \no &
  \no & \yes & \yes & \half & \yes & \imgintext{figures/bash.png} &
  \yes & Open & $\infty$ &
  Self & 490,520 \\

\midrule[0.05em]
\rowcolor{rowblue}
\textbf{\OURMETHOD~(Ours)} &
  \textbf{Agent skills} & \yes &
  \yes & \yes & \yes & \yes & \yes & \imgintext{figures/bash.png} &
  \yes & \textbf{Open} & $\infty$ &
  \textbf{Guided} & \textbf{5,723} \\

\bottomrule[0.08em]
\end{tabular}%
}
\end{table*}

\paragraph{Tool-Using Agents}
LLM-based agents interact with the external world through tool use, enabling them to execute actions beyond the limits of their parametric knowledge~\citep{toolsurvey}. To strengthen this capability, early efforts synthesized training data for API selection and argument filling~\citep{gorilla, apigen}. Subsequent work expanded API and tool coverage~\citep{toolbridge, toolace}, increased the interaction turns and orchestration complexity of tool use~\citep{toolmind, apigen_mt, inftool}, and broadened workflow sources by mining latent tool-use patterns from raw text~\citep{gem}. A parallel line studies web-search agents and synthesizes search trajectories over web content~\citep{infoagent, webshaper, webexplorer, simpledeepsearcher, progsearch, asearcher, synthagent}. Despite this progress, these datasets typically operate over predefined toolsets, yielding a closed action space that cannot fully capture the open-ended and compositional nature of real-world tasks. In contrast, \OURMETHOD{} grounds agents in a Bash terminal, where the action space is no longer bounded by a predefined toolset but instead spans the full spectrum of composable system commands within a real execution environment.

\paragraph{Terminal-Based Agents}
The rise of CLI-based coding agents such as Codex~\citep{codex} and Claude Code~\citep{claudecode} has shifted agent interaction toward direct operation in terminal environments, motivating recent efforts to synthesize terminal-agent training data. Endless Terminal~\citep{endless}, TermiGen~\citep{termigen}, and Nemotron-Terminal~\citep{nemotronterminal} start from human-defined seed data and use LLMs to synthesize terminal tasks, before constructing the corresponding environments, verification scripts, and teacher trajectories. TerminalTraj~\citep{terminaltraj} instead starts from GitHub repositories and infers task instructions and validation logic from existing codebases. However, these methods still face limitations in task diversity, environment-task alignment, and trajectory efficiency. \OURMETHOD{} addresses these limitations by using agent skills as the synthesis primitive. Each skill specifies \textit{\circnum{1} what should be accomplished, \circnum{2} when the skill is applicable, and \circnum{3} how the task should be executed}, providing a unified anchor from which task instructions, executable environments, verification criteria, and teacher trajectories can be co-derived.

\section{\OURMETHOD} \label{sec:method}

\begin{figure}[!t]
    \centering
    \includegraphics[width=\textwidth]{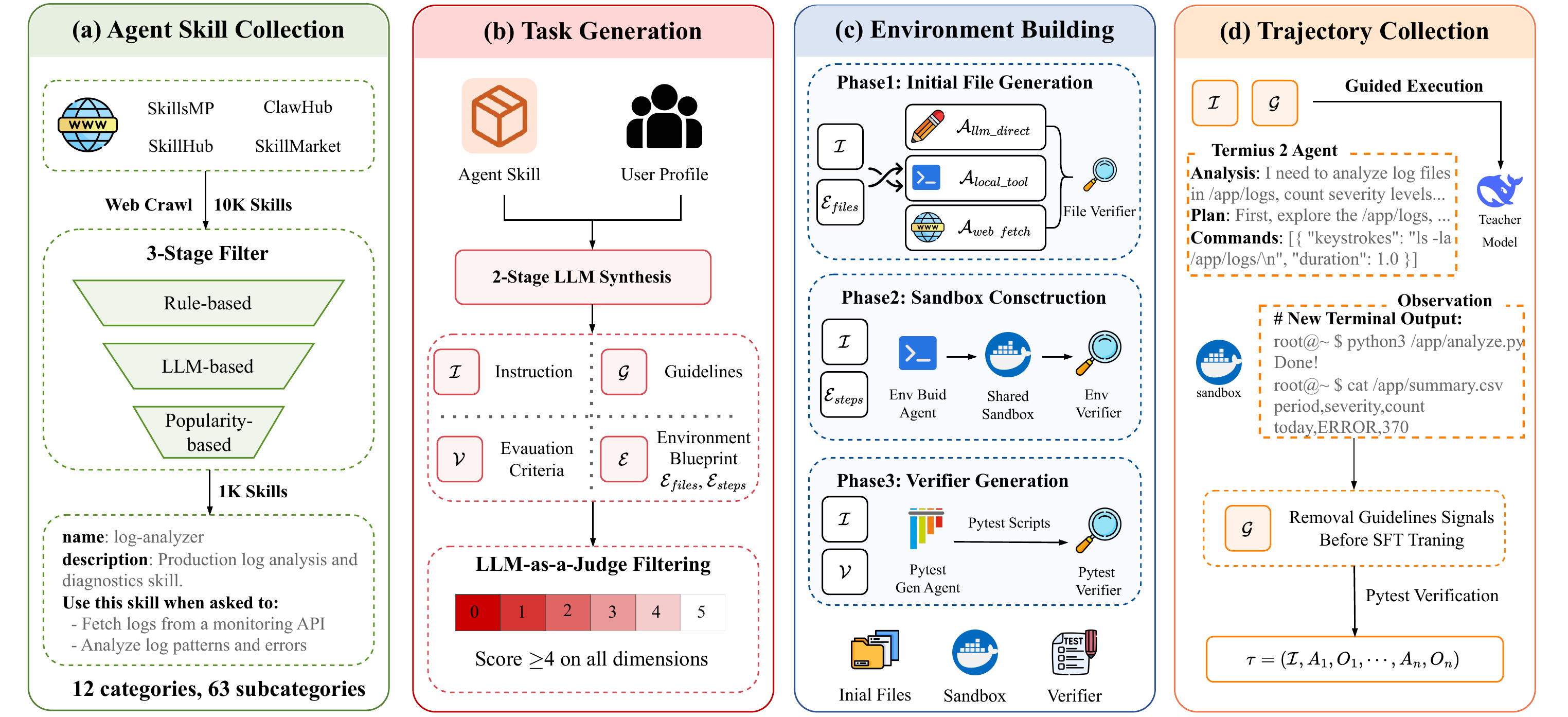}
    \vspace{-3mm}
    \caption{\textbf{Overview of \OURMETHOD{}.} We start from real-world agent skills, filter high-quality candidates via rule-based, LLM-based, and popularity-based screening, pair each skill with a user persona to synthesize diverse and verifiable task quadruples $(\mathcal{I}, \mathcal{E}, \mathcal{V}, \mathcal{G})$, construct executable sandboxes through iterative generate-verify-repair cycles, and collect efficient trajectories.}
    \label{fig:method}
\end{figure}

As illustrated in Figure~\ref{fig:method}, \OURMETHOD~uses agent skills as the synthesis primitive to jointly align task instructions, environments, and trajectories. The pipeline consists of four stages: (a) \textbf{Agent Skill Collection} (\S\ref{sec:skill_collection}) constructs a terminal capability space from real-world agent skills; (b) \textbf{Task Generation} (\S\ref{sec:task_collection}) pairs each skill with user personas to diversify the usage scenarios of the underlying capability, and synthesizes instructions, environment blueprints, evaluation criteria, and execution guidelines; (c) \textbf{Environment Building} (\S\ref{sec:env_construction}) instantiates each blueprint into an executable sandbox with initial files, setup scripts, and pytest verifiers; and (d) \textbf{Trajectory Collection} (\S\ref{sec:traj_collection}) collects efficient teacher trajectories under skill-derived guidance. We provide comprehensive statistics of \OURMETHOD~in \S\ref{sec:statis}.

\subsection{Agent Skill Collection} \label{sec:skill_collection}
To construct a high-quality and broadly distributed terminal capability space, we collect \textbf{10{,}000} agent skills from ClawHub and SkillMP, and apply a three-stage filter to retain those that are relevant and informative. \textbf{(1) Rule-based} filtering removes skills with terminal-irrelevant names (e.g., ``skill creator''), leaving 8{,}520 skills. \textbf{(2) LLM-based} filtering scores each skill on terminal applicability and content richness (1--3 each), retaining only those scoring the maximum on both, yielding 3{,}025 skills. \textbf{(3) Popularity-based} filtering ranks the remaining skills by their download counts and selects the top \textbf{1{,}000}, spanning 12 categories and 63 subcategories. The full taxonomy is shown in Fig.~\ref{fig:app_data}.

Building on these single skills as atomic primitives, we further extend the synthesis space along two complementary dimensions: \textbf{agent skill teams}, which compose multiple skills \emph{within the same subcategory} into a multi-role workflow as a \textit{depth extension}, and \textbf{agent skill graphs}, which connect skills \emph{across different subcategories} into an end-to-end pipeline as a \textit{breadth extension}. Specifically, we use SkillNet~\citep{skillnet} to classify pairs of skills into 4 relations: \emph{Compose with}, \emph{Depends on}, \emph{Similar to}, and \emph{Belong to}. We treat \emph{Compose with} and \emph{Depends on} as composition signals, use \emph{Similar to} as a deduplication signal, and discard \emph{Belong to} as redundant with our subcategory taxonomy. Composition relations within the same subcategory are grouped into agent skill teams and synthesized into multi-role workflows via TeamSkill-Creator~\citep{jiuwenclaw} driven by Claude Code, yielding \textbf{76} skill teams. Composition relations spanning different subcategories are used to construct a cross-subcategory composition graph, from which we greedily extract maximal paths until all nodes are consumed, yielding \textbf{237} skill graphs. Both teams and graphs are flattened into the same \texttt{skill.md} format and, together with the 1{,}000 single skills, serve as the synthesis primitives $\mathcal{S}=\mathcal{S}_{single}\cup\mathcal{S}_{team}\cup\mathcal{S}_{graph}$ for the next stage.

\subsection{Task Generation}
\label{sec:task_collection}
Given the synthesis primitives $\mathcal{S}$ from Sec.~\ref{sec:skill_collection}, this stage converts each into a unified task specification. These primitives are well-suited for this purpose because they explicitly describe what the agent should accomplish, when the skill should be applied, and how the task should be executed. These three aspects provide natural anchors for constructing the task instruction, execution context, and execution guideline, respectively. To diversify the usage scenarios of each capability, we pair each primitive $\mathcal{S}$ with a user persona $\mathcal{U}$ sampled from FinePersonas~\citep{finwebedu, personahub}, which consists of short natural-language profiles describing potential users' backgrounds, roles, and preferences. During task synthesis, the LLM instantiates a task only when the sampled persona forms a coherent usage scenario for the given primitive, while irrelevant pairs are ignored. For each pair, we synthesize a quadruple $(\mathcal{I}, \mathcal{E}, \mathcal{V}, \mathcal{G})$ as follows:
\begin{equation}
\mathcal{I}, \mathcal{E}, \mathcal{V} = \mathrm{LLM}(\mathcal{P}_s, \mathcal{S}, \mathcal{U}), \quad
\mathcal{G} = \mathrm{LLM}(\mathcal{P}_g, \mathcal{I}, \mathcal{S}),
\end{equation}

where $\mathcal{P}_s$ and $\mathcal{P}_g$ denote the prompt templates for task synthesis (Appendix~\ref{prompt:core_goal}) and guideline generation (Appendix~\ref{prompt:guideline}), respectively. Here, $\mathcal{I}$ denotes the task instruction. $\mathcal{E}$ denotes the environment blueprint used for sandbox construction, consisting of initial files $\mathcal{E}_{\text{files}}$ and setup steps $\mathcal{E}_{\text{steps}}$. $\mathcal{V}$ specifies the evaluation criteria that translate the task goal into verifiable completion conditions, which are later used to generate pytest-based verifiers. $\mathcal{G}$ provides skill-derived execution guidance for collecting teacher trajectories.

To ensure task quality, we apply an LLM-as-a-Judge filter along five dimensions: \circnum{1} instruction quality, \circnum{2} closed-world solvability, \circnum{3} blueprint completeness, \circnum{4} guideline quality, and \circnum{5} evaluation-criteria quality, and the Prompt in Appendix~\ref{prompt:judge}. We retain only samples that receive a score of at least 4 on every dimension and pass them to the subsequent stages. An end-to-end example of this stage is provided in Appendix~\ref{sec:example-task-gen}.

\subsection{Environment Building}
\label{sec:env_construction}

Given the task specification $(\mathcal{I}, \mathcal{E}, \mathcal{V}, \mathcal{G})$, this stage instantiates the environment blueprint into an executable and verifiable sandbox. Each sandbox comprises three artifacts: initial files $\mathcal{F}$ that define the starting workspace, a setup script $\mathcal{B}_{\text{env}}$ that prepares runtime dependencies and services, and a pytest verifier $\mathcal{T}_{\text{test}}$ that provides automatic completion checking. To ensure the quality of all three artifacts, we construct them through a unified generate-verify-repair (\textbf{GVR}) mechanism:
\begin{equation}
x^{(0)} = \mathrm{Generate}(\cdot), \quad
x^{(t+1)} = \mathrm{Repair}\bigl(x^{(t)},\; \mathrm{Verify}(x^{(t)})\bigr),
\end{equation}
where $x$ denotes the artifact. Tasks that cannot be repaired within $T=3$ iterations are discarded. We now describe the $\mathrm{Generate}(\cdot)$ and $\mathrm{Verify}(\cdot)$ procedures instantiated for each artifact.

\textbf{(1) Initial Files: }
To support the generation of arbitrary file types, we adopt a multi-agent architecture that routes each file in $\mathcal{E}_{\text{files}}$ to a dedicated agent based on its generation mode, which is annotated during task generation in Section~\ref{sec:task_collection}. Specifically, an LLM-synthesis agent  $\mathcal{A}_{llm\_direct}$, a local-tool agent $\mathcal{A}_{local\_tool}$ equipped with file-creation tools, or a remote-fetch agent $\mathcal{A}_{remote\_fetch}$ with search tools, and each conditioned on $\mathcal{I}$ and the file description (prompts in Appendices~\ref{prompt:file_direct}, \ref{prompt:file_local_tool}, and \ref{prompt:file_remote_fetch}). A file-verification agent (prompt in Appendix~\ref{prompt:file_verify}) then inspects both \emph{\circnum{1} internal correctness} (well-formedness and description alignment) and \emph{\circnum{2} external consistency} (cross-file paths, references, and schemas), with failures triggering joint repair across dependent files.

\textbf{(2) Setup Scripts: }
For scalability, rather than building a per-task Docker image, we use a shared general-purpose sandbox with task-specific setup scripts. An environment-building agent converts the natural-language steps $\mathcal{E}_{\text{steps}}$ into executable shell commands for dependency installation, service initialization, and runtime configuration (Appendix~\ref{prompt:env_build}). To confirm logical correctness rather than relying on script exit codes alone, an environment-verification agent then generates and executes diagnostic probing scripts that inspect whether required packages, services, and runtime states are properly established (Appendix~\ref{prompt:env_verify}); detected issues are returned to the building agent for repair.

\textbf{(3) Pytest Verifiers:}
A verifier-generation agent translates the evaluation criteria $\mathcal{V}$ into pytest scripts over the expected post-execution state, conditioned on $\mathcal{I}$, $\mathcal{V}$, $\mathcal{F}$, and $\mathcal{B}_{\text{env}}$ (Appendix~\ref{prompt:pytest_generation}). A verifier-validation agent then checks two properties: \emph{\circnum{1} executability}, requiring all test scripts to run without syntax or import errors, and \emph{\circnum{2} reliability}, requiring all tests to fail on the pre-execution initial state to rule out vacuous passes. Tests violating either property are returned to the generation agent for repair. An example of this stage is provided in Appendix~\ref{sec:example-env-build}.

\subsection{Trajectory Collection}
\label{sec:traj_collection}

Given the synthesized tasks and their executable sandboxes, this stage collects efficient teacher trajectories. Specifically, we use the execution guideline $\mathcal{G}$ synthesized in Section~\ref{sec:task_collection} as skill-derived guidance for the teacher model, rather than letting it explore freely, which often results in lengthy and redundant trajectories. Following the setup of~\citep{nemotronterminal} for fair comparison, we adopt DeepSeek-V3.2~\citep{liu2025deepseek} with the Terminus2 scaffolding~\citep{Harbor_Framework} as the teacher, which isolates the contribution of our data construction pipeline from differences in teacher capability. At each step, the teacher produces an action $\mathcal{A}_t = \pi_{\text{teacher}}(\mathcal{I}, \mathcal{G}, \mathcal{H}_{t-1})$ and receives an observation $\mathcal{O}_t$ from the sandbox, where $\mathcal{H}_{t-1} = (\mathcal{A}_1, \mathcal{O}_1, \ldots, \mathcal{A}_{t-1}, \mathcal{O}_{t-1})$ denotes the prior interaction history. After rollout, we run the verifier $\mathcal{T}_{\text{test}}$ against the resulting sandbox state to annotate each trajectory with its verification outcome, while retaining both successful and failed trajectories~\citep{terminaltraj}. A representative trajectory with per-step analysis, commands, and observations is shown in Appendix~\ref{sec:example-traj}. Importantly, the guideline $\mathcal{G}$ is used only during trajectory collection: before SFT training, we remove $\mathcal{G}$ from the training input so that the student model learns from the verified terminal interaction itself rather than relying on auxiliary procedural hints.

\subsection{Data Statistics}
\label{sec:statis}
\begin{figure}[t]
\centering
\begin{minipage}[c]{0.30\columnwidth}
  \centering
  \setlength{\tabcolsep}{4pt}
  \renewcommand{\arraystretch}{0.95}
  \captionof{table}{\textbf{Key statistics of \OURMETHOD{}}.
  }
  \label{tab:data_statistic}
  \resizebox{0.95\linewidth}{!}{%
  \begin{tabular}{l r}
      \toprule[0.08em]
      \textbf{Statistic} & \textbf{Value} \\
      \midrule[0.05em]
      \rowcolor{rowblue}
      \multicolumn{2}{l}{\textit{\textbf{Task}}} \\
      \quad \# Persona              & 4,973 \\
      \quad \# Single-skill Tasks   & 3,723 \\
      \quad \# Team-skill Tasks     & 1,000 \\
      \quad \# Graph-skill Tasks    & 1,000 \\
      \quad \# Total Tasks          & 5,723 \\
      \midrule[0.05em]
      \rowcolor{rowblue}
      \multicolumn{2}{l}{\textit{\textbf{Environment}}} \\
      \quad Avg.\ Init.\ Files              & 2.25 \\
      \quad \# File Types                   & 104 \\
      \quad Avg.\ \texttt{pytest} Tests     & 4.27 \\
      \midrule[0.05em]
      \rowcolor{rowblue}
      \multicolumn{2}{l}{\textit{\textbf{Trajectory}}} \\
      \quad Avg.\ Steps                     & 13.44 \\
      \quad Avg.\ Tokens                    & 18,176 \\
      \quad \# Bash Cmd.                    & 1,939 \\
      \bottomrule[0.08em]
    \end{tabular}}
\end{minipage}
\hfill
\begin{minipage}[c]{0.67\columnwidth}
  \centering
  \includegraphics[width=\linewidth]{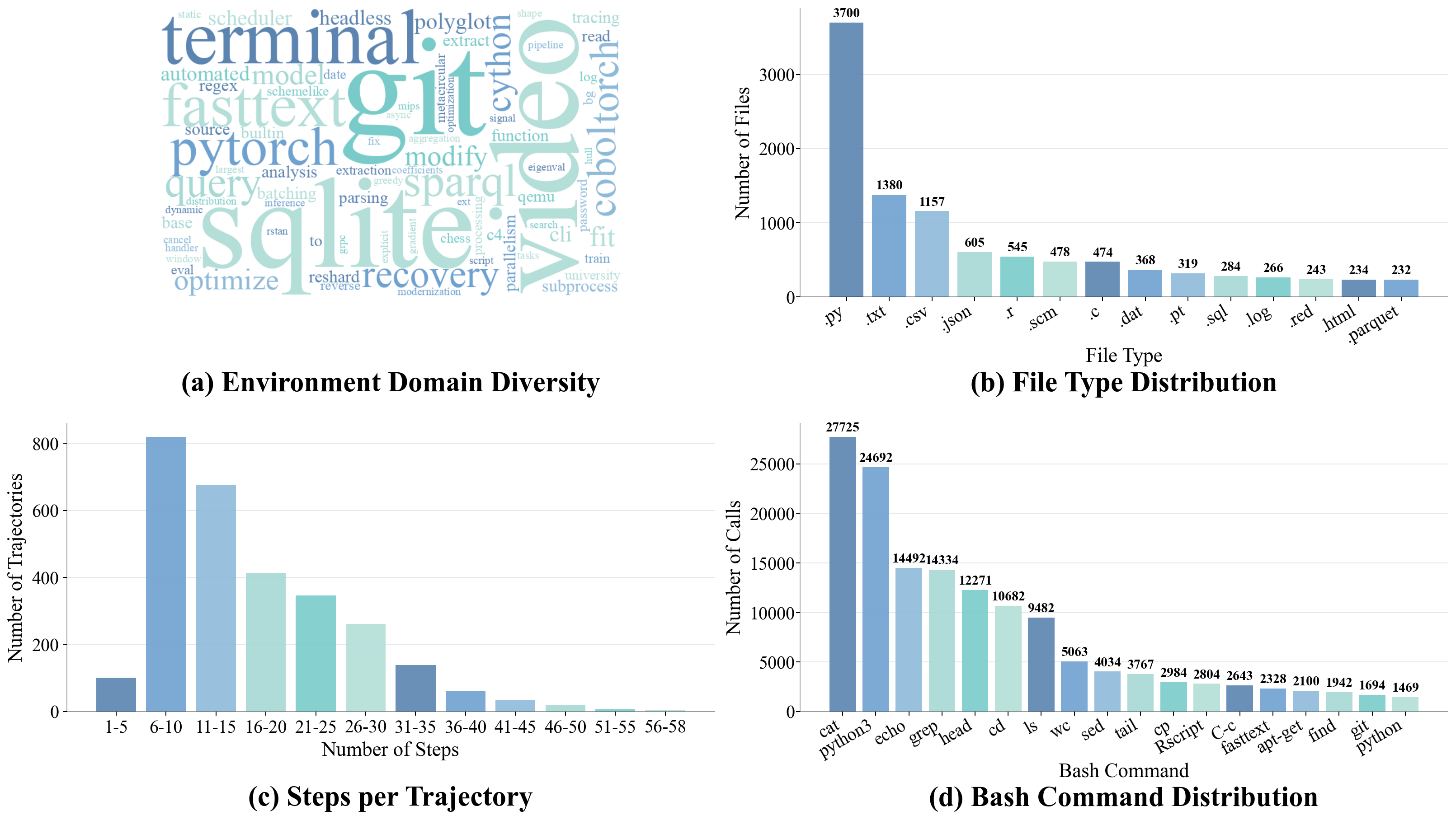}
  \caption{\textbf{Comprehensive statistics of the \OURMETHOD{}.}}
  \label{fig:data_statistic}
  \vspace{-3mm}
\end{minipage}
\end{figure}
To characterize the resulting dataset, we analyze \OURMETHOD~from three dimensions: task coverage, environment diversity, and trajectory complexity, as shown in Table~\ref{tab:data_statistic} and Figure~\ref{fig:data_statistic}. At the task level, \OURMETHOD~uses 1{,}000 single skills together with 76 skill teams and 237 skill graphs to synthesize \textbf{5{,}723} terminal-agent tasks, covering diverse capability scenarios. At the environment level, each task contains an executable sandbox with an average of 2.25 initial files and 4.27 pytest tests. These environments span 104 file types, including both textual formats (e.g., JSON) and non-textual formats (e.g., SQL and Parquet), reflecting the diversity of realistic terminal workspaces. At the trajectory level, each verified demonstration contains 13.44 steps and 18,176 tokens on average, and the collected trajectories cover 1,939 distinct Bash commands. These statistics indicate that \OURMETHOD~constructs not only diverse task scenarios, but also executable environments and long-horizon trajectories suitable for terminal-agent training.
\section{Experiments} \label{sec:exp}
\begin{table*}[t]
\centering
\setlength{\tabcolsep}{4pt}
\renewcommand{\arraystretch}{1.05}
\caption{\textbf{Main results on 6 benchmarks:} Terminal-Bench 2.0, AIME 24, AIME 25, DABench, TableBench, and BIRD. We report accuracy (\%), including pass@1 and pass@3. In the \textit{Open-source Models ($<$100B)} block, the best result is shown in \textbf{bold}, and the second-best result is \underline{underlined}.}
\label{tab:main_results}
\resizebox{\textwidth}{!}{%
\begin{tabular}{l c *{14}{c}}
\toprule[0.08em]
\multirow{2}{*}{\textbf{Model}}
& \multirow{2}{*}{\makecell{\textbf{\# Training}\\\textbf{Samples}}}& \multicolumn{2}{c}{\textbf{TB 2.0}}
& \multicolumn{2}{c}{\textbf{AIME24}}
& \multicolumn{2}{c}{\textbf{AIME25}}
& \multicolumn{2}{c}{\textbf{DABench}}
& \multicolumn{2}{c}{\textbf{TableBench}}
& \multicolumn{2}{c}{\textbf{BIRD}}
& \multicolumn{2}{c}{\textbf{Avg.}} \\
\cmidrule(lr){3-4}
\cmidrule(lr){5-6}
\cmidrule(lr){7-8}
\cmidrule(lr){9-10}
\cmidrule(lr){11-12}
\cmidrule(lr){13-14}
\cmidrule(lr){15-16}
& & \textbf{P@1} & \textbf{P@3}
& \textbf{P@1} & \textbf{P@3}
& \textbf{P@1} & \textbf{P@3}
& \textbf{P@1} & \textbf{P@3}
& \textbf{P@1} & \textbf{P@3}
& \textbf{P@1} & \textbf{P@3}
& \textbf{P@1} & \textbf{P@3} \\
\midrule[0.05em]

\rowcolor{rowblue}
\multicolumn{16}{c}{\textit{\textbf{Frontier Proprietary Models}}} \\
\modelicon{1.1em}{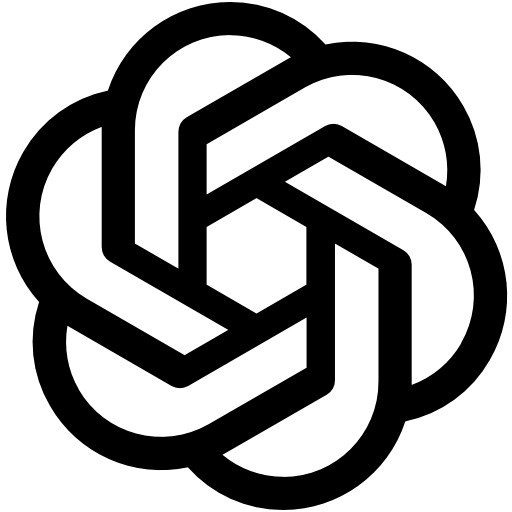}GPT-5.2 & -- & 53.9 & 74.2 & -- & -- & -- & -- & -- & -- & -- & -- & -- & -- & -- & -- \\
\modelicon{1.1em}{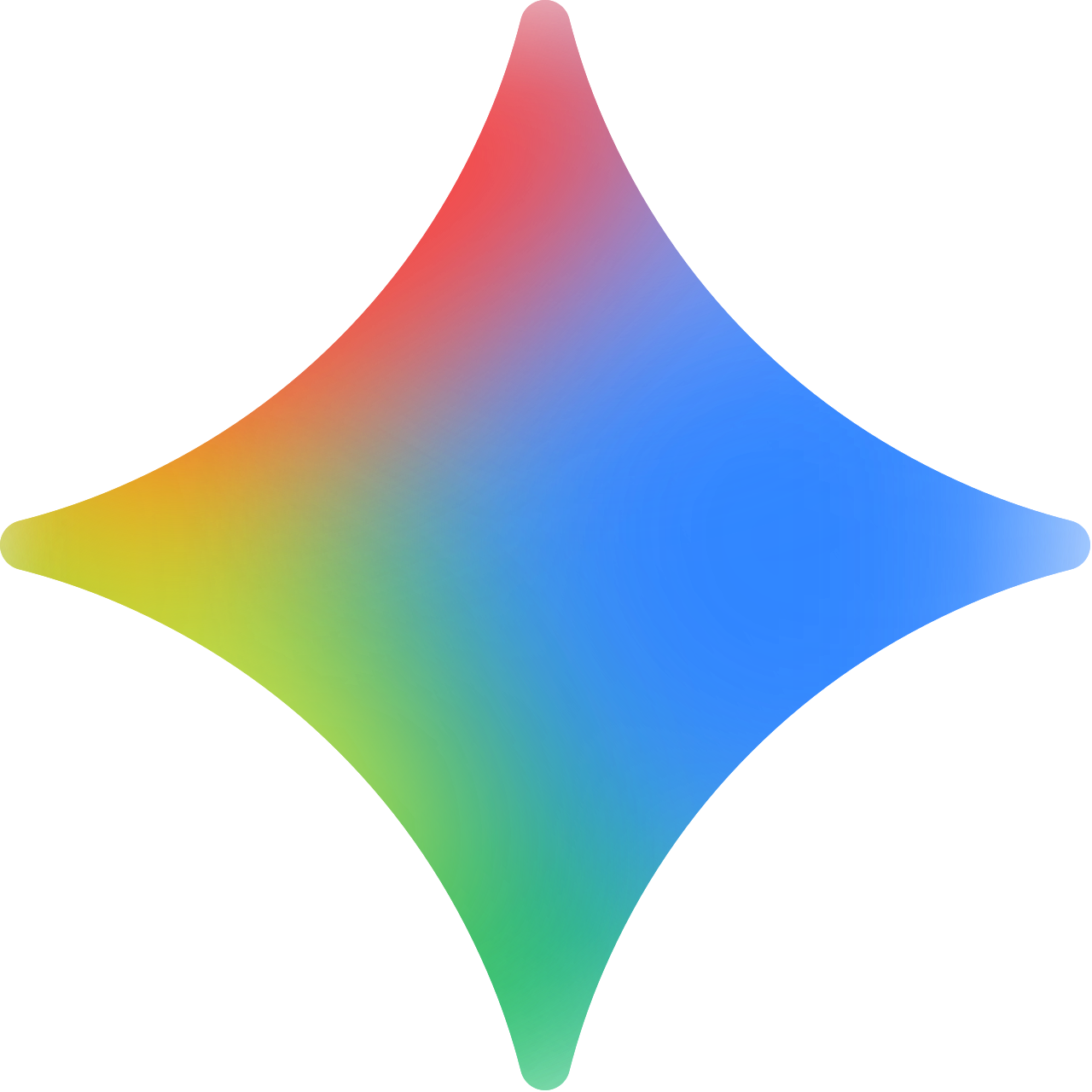}Gemini-3-Flash & -- & 51.7 & 66.3 & 93.3 & 100.0 & 90.0 & 96.7 & 87.0 & 92.0 & 73.5 & 77.5 & 49.5 & 58.0 & 74.2 & 81.8 \\
\modelicon{1.1em}{gemini.png}Gemini-3.1-Pro-Preview & -- & 68.5 & 80.9 & 96.7 & 100.0 & 100.0 & 100.0 & 88.0 & 92.5 & 77.0 & 79.0 & 59.0 & 62.5 & 81.5 & 85.8 \\
\modelicon{1.1em}{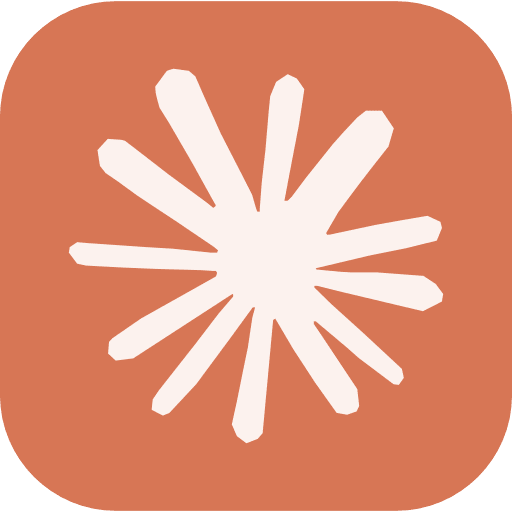}Claude-Sonnet-4.5 & -- & 42.7 & 53.9 & -- & -- & -- & -- & -- & -- & -- & -- & -- & -- & -- & -- \\

\midrule[0.05em]
\rowcolor{rowblue}
\multicolumn{16}{c}{\textit{\textbf{Open-source Models ($>$100B)}}} \\
\modelicon{1.1em}{openai.png}GPT-OSS-120B (High) & -- & 13.5 & 27.0 & 90.0 & 96.7 & 86.7 & 96.7 & 75.0 & 90.0 & 66.5 & 79.0 & 50.0 & 59.5 & 63.6 & 74.8 \\
\modelicon{1.1em}{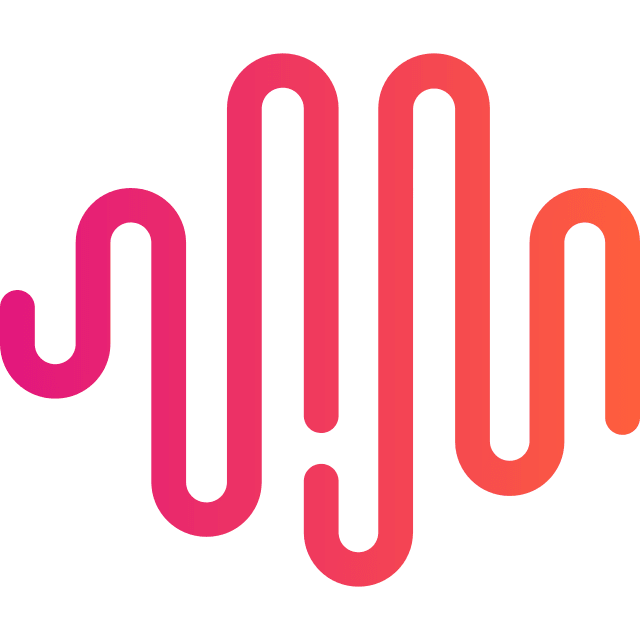}MiniMax-M2.7 & -- & 56.2 & 65.2 & 50.0 & 83.3 & 63.3 & 96.7 & 85.5 & 90.5 & 76.0 & 83.0 & 49.5 & 58.5 & 63.4 & 79.5 \\
\modelicon{1.1em}{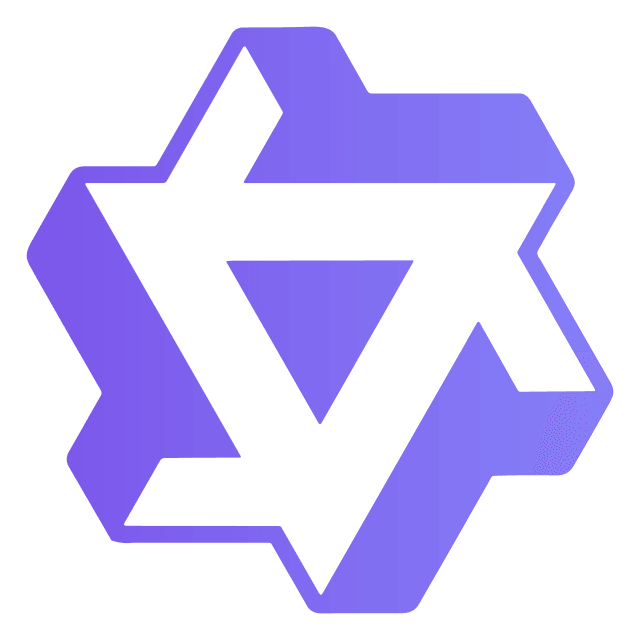}Qwen3-Coder & -- & 23.6 & 39.3 & 70.0 & 73.3 & 46.7 & 66.7 & 81.0 & 88.5 & 67.5 & 75.5 & 49.5 & 62.0 & 56.4 & 67.6 \\
\modelicon{1.1em}{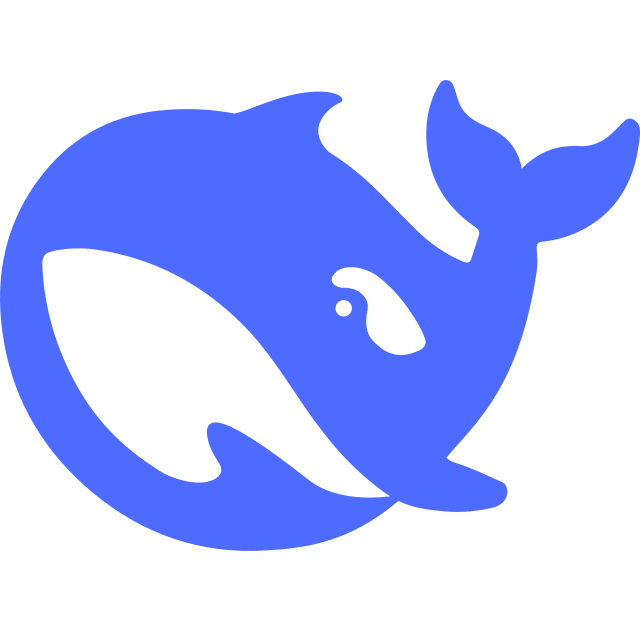}DeepSeek-V3.2 & -- & 38.2 & 52.8 & 83.3 & 96.7 & 93.3 & 96.7 & 87.0 & 92.5 & 72.5 & 79.5 & 50.0 & 56.5 & 70.7 & 79.1 \\
\modelicon{1.1em}{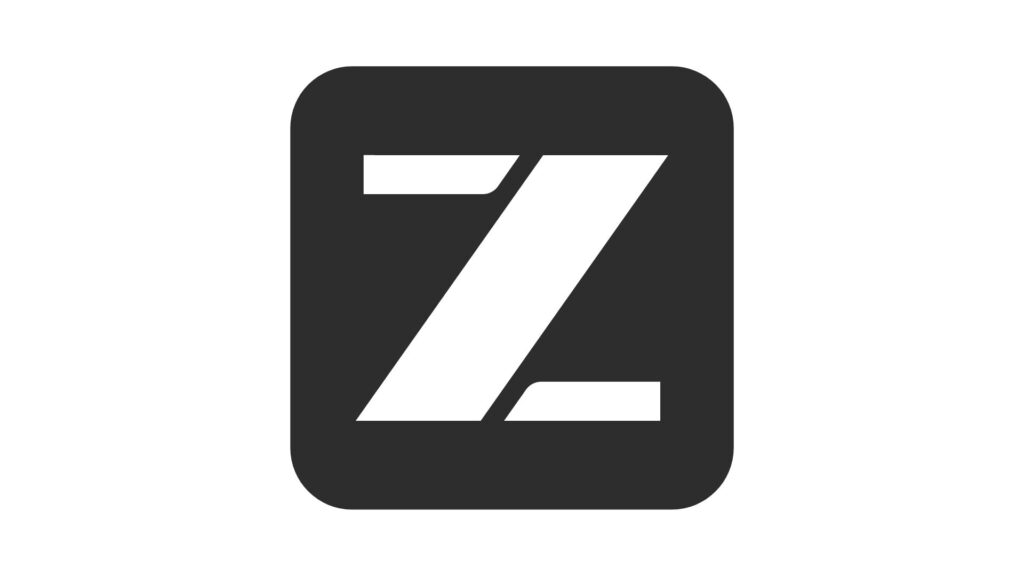}GLM-5 & -- & 56.2 & 65.2 & 86.7 & 96.7 & 83.3 & 93.3 & 88.0 & 91.5 & 76.5 & 82.5 & 56.0 & 60.5 & 74.5 & 81.6 \\
\modelicon{1.1em}{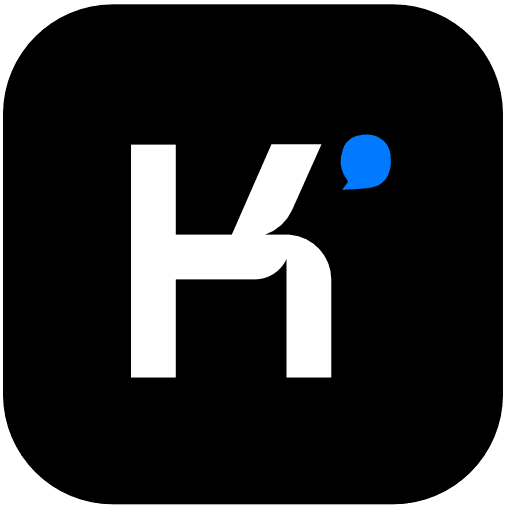}Kimi-K2-Thinking & -- & 36.0 & 49.4 & 76.7 & 83.3 & 70.0 & 86.7 & 85.0 & 91.5 & 75.0 & 82.0 & 52.5 & 58.5 & 65.9 & 75.2 \\
\midrule[0.05em]
\rowcolor{rowblue}
\multicolumn{16}{c}{\textit{\textbf{Open-source Models ($<$100B)}}} \\
\modelicon{1.1em}{qwen.png}Qwen3-8B & -- & 2.2 & 3.37 & 63.3 & 76.7 & 53.3 & 63.3 & 16.5 & 32.5 & 24.5 & 39.0 & 4.0 & 9.0 & 27.3 & 37.3 \\
\modelicon{1.1em}{qwen.png}Qwen3-14B & -- & 4.5 & 7.87 & 76.7 & 86.7 & 53.3 & 66.7 & 38.0 & 69.0 & 35.5 & 60.0 & 7.5 & 18.5 & 35.9 & 51.5 \\
\modelicon{1.1em}{qwen.png}Qwen3-32B & -- & 4.5 & 9.0 & 60.0 & 86.7 & 43.3 & 66.7 & 44.5 & 75.0 & 26.5 & 53.5 & 6.5 & 16.0 & 30.9 & 51.2 \\
\modelicon{1.1em}{openai.png}GPT-OSS-20B (High) & -- & 3.4 & 9.0 & \textbf{93.3} & \underline{96.7} & 83.3 & \textbf{93.3} & 59.5 & 82.5 & 38.5 & 64.0 & 0.0 & 6.0 & 46.3 & 58.6 \\
\modelicon{1.1em}{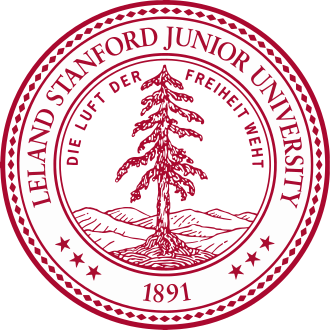}EndLess Terminal-8B & 3.3k & 6.7 & 12.4 & 23.3 & 36.7 & 30.0 & 43.3 & 26.0 & 38.5 & 28.0 & 42.0 & 6.0 & 11.5 & 20.0 & 30.7 \\
\modelicon{1.1em}{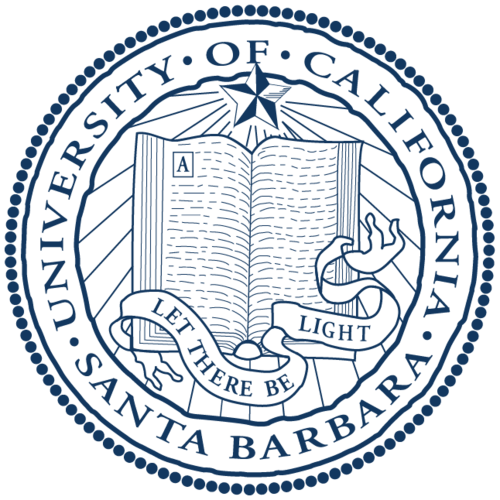}TermiGen-32B & 3.3k & 19.1 & 27.0 & 46.7 & 56.7 & 33.3 & 43.3 & 81.0 & 91.5 & 68.0 & \textbf{79.0} & \underline{50.5} & \textbf{62.5} & 49.8 & 60.0 \\
\modelicon{1.1em}{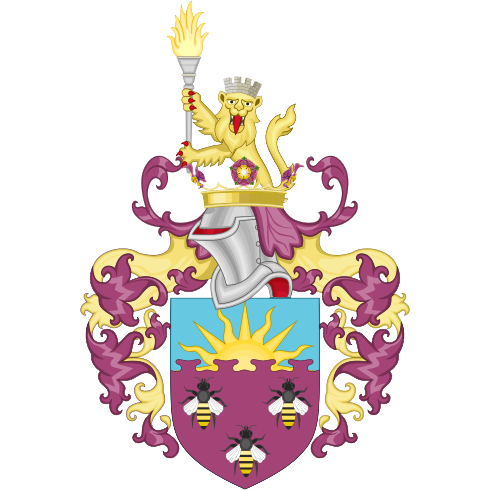}Termial-Traj-32B & 50.7k & 22.0 & 27.0 & 46.7 & 60.0 & 26.7 & 46.7 & 76.0 & 88.0 & 62.0 & \underline{76.0} & 49.5 & \underline{61.5} & 47.2 & 59.9 \\
\modelicon{1.1em}{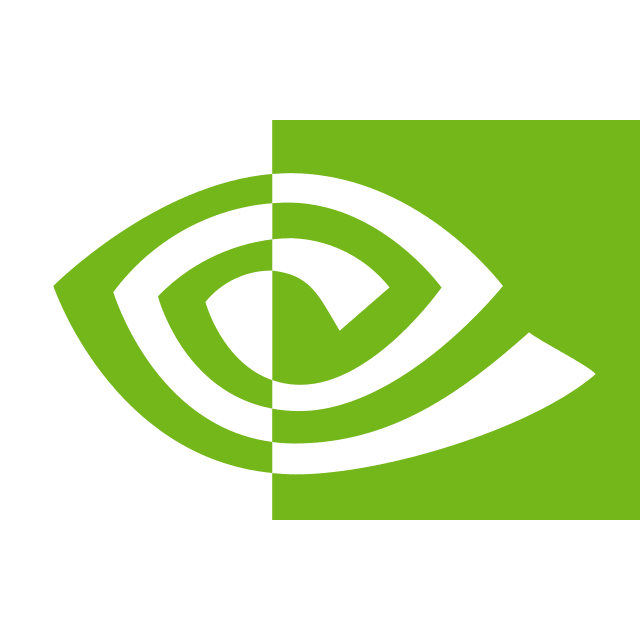}Nemotron-Terminal-8B & 490.5k & 13.5 & 21.3 & 83.3 & 83.3 & 63.3 & \underline{90.0} & 80.5 & \underline{92.0} & 66.5 & 73.5 & 44.0 & 55.5 & 58.5 & 69.3 \\
\modelicon{1.1em}{nvidia.png}Nemotron-Terminal-14B & 490.5k & 20.2 & 24.7 & \underline{90.0} & \underline{96.7} & \textbf{90.0} & \textbf{93.3} & 80.5 & 90.5 & 70.0 & 75.0 & \underline{50.5} & 59.0 & 66.9 & 73.2 \\
\modelicon{1.1em}{nvidia.png}Nemotron-Terminal-32B & 490.5k & \underline{27.0} & \underline{37.1} & \textbf{93.3} & \textbf{100.0} & \underline{86.7} & \textbf{93.3} & \underline{81.5} & \textbf{93.0} & \textbf{72.5} & 75.5 & \textbf{52.5} & 57.0 & \underline{68.9} & \underline{76.0} \\

\addlinespace[0.5ex]
\hdashline
\addlinespace[0.5ex]
\rowcolor{rowblue}
\multicolumn{16}{c}{\textit{\textbf{Ours}}} \\
\textbf{\OURMETHOD-8B} & 5.7k & 15.7 & 23.6 & 86.7 & 93.3 & 80.0 & \underline{90.0} & 80.5 & 91.0 & 69.5 & 74.5 & 48.5 & 58.0 & 63.5 & 71.7 \\
\textbf{\OURMETHOD-14B} & 5.7k & 21.3 & 27.0 & \underline{90.0} & \underline{96.7} & 83.3 & \textbf{93.3} & 81.0 & \underline{92.0} & 70.5 & \underline{76.0} & 50.0 & 59.5 & 66.0 & 74.1 \\
\textbf{\OURMETHOD-32B} & 5.7k & \textbf{31.5} & \textbf{43.8} & \textbf{93.3} & \underline{96.7} & \underline{86.7} & \textbf{93.3} & \textbf{83.5} & \textbf{93.0} & \underline{71.5} & \textbf{79.0} & 49.5 & \underline{61.5} & \textbf{69.3} & \textbf{77.9} \\
\bottomrule[0.08em]
\end{tabular}
}
\end{table*}

\subsection{Experiment Setting}
\paragraph{Baselines}
Following prior work~\citep{termigen}, we compare \OURMETHOD~against the following baselines.
\textbf{Frontier Proprietary Models}: GPT-5.2, Gemini-3-Flash, Gemini-3.1-Pro-Preview, and Claude-Sonnet-4.5.
\textbf{Open-source Models ($>$100B)}: GPT-OSS-120B~\citep{gptoss}, MiniMax-M2.7, Qwen3-Coder-480B~\citep{qwen3}, DeepSeek-V3.2~\citep{liu2025deepseek}, GLM-5~\citep{glm5}, and Kimi-K2-Thinking~\citep{kimik2}.
\textbf{Open-source Models ($<$100B)}: Qwen3-8B/14B/32B~\citep{qwen3}, GPT-OSS-20B, EndLess Terminal-8B~\citep{endless}, TermiGen-Qwen3-32B~\citep{termigen}, Terminal-Traj-32B~\citep{terminaltraj}, and Nemotron-Terminal-8B/14B/32B~\citep{nemotronterminal}.
All models are evaluated under the Terminus2 Agent scaffolding~\citep{Harbor_Framework}, except Endless Terminal-8B and TermiGen-32B, which are evaluated in their native formats (i.e., EndlessAgent and BashAgent).

\paragraph{Benchmarks}
Following prior work~\citep{taskcraft,nemotronterminal,termigen}, we evaluate models on six benchmarks. \textbf{Terminal-Bench~2.0}~\citep{tb2,glm5} evaluates terminal-agentic coding ability. The remaining benchmarks are converted into terminal-based tasks~\citep{nemotronterminal}: \textbf{AIME24/25}~\citep{aime24,aime25} evaluate mathematical reasoning, \textbf{DABench}~\citep{dabench} and \textbf{TableBench}~\citep{tablebench} evaluate table-based data analysis, and \textbf{BIRD}~\citep{bird} evaluates SQL-based data analysis. Details in Appendix~\ref{appendix:benchmarks}.

\paragraph{Implementation Details}

Following prior work~\citep{nemotronterminal}, we conduct SFT using Swift~\citep{ms-swift}. Specifically, we set the learning rate to $2\mathrm{e}{-5}$, with a warmup ratio of $0.1$, weight decay of $1\mathrm{e}{-4}$, and train for $2$ epochs with sequence length of $32{,}768$. During inference, we adopt a context length of $40{,}960$ tokens, temperature $0.6$, top-$p = 0.95$, top-$k = 20$, and min-$p = 0.0$. All experiments are conducted on 4 nodes, and each is equipped with 8 H20-141GB GPUs. For all other baselines, we follow the officially recommended sampling parameters. Details in the Appendix~\ref{appendix:sampling}. 

\subsection{Main Results}
Table~\ref{tab:main_results} presents the main results across six benchmarks. We summarize the key findings as follows:

\paragraph{\textit{\OURMETHOD~delivers strong performance with significantly higher sample efficiency.}}
At all three model sizes, \OURMETHOD~outperforms existing terminal-synthetic baselines across benchmarks. \OURMETHOD-32B achieves \textbf{69.3} Avg.~Pass@1 and \textbf{77.9} Avg.~Pass@3. On Terminal-Bench~2.0, \OURMETHOD-32B achieves \textbf{31.5} Pass@1 and \textbf{43.8} Pass@3, surpassing all $<$100B open-source models. Notably, these results are obtained with only \textbf{5.7K} training trajectories (\textbf{1.2\%} of the 490.5K used by Nemotron-Terminal), demonstrating that skill-grounded synthesis can achieve superior performance at a fraction of the data scale.

\paragraph{\textit{Skill-grounded trajectories generalize beyond terminal-native coding without auxiliary data.}}
Unlike Nemotron-Terminal, which supplements terminal data with 226.3K auxiliary samples spanning math, code, and SWE tasks, \OURMETHOD~is trained exclusively on 5.7K terminal-style trajectories. Despite this, \OURMETHOD-32B matches or exceeds Nemotron-Terminal-32B on all five non-terminal benchmarks, indicating that skill-grounded trajectories develop transferable problem-solving capabilities that generalize to mathematical reasoning, table analysis, and SQL tasks without domain-specific training data.

\paragraph{\textit{The performance advantage is most pronounced at the 32B scale.}}
On Terminal-Bench~2.0, \OURMETHOD-32B achieves the largest absolute margin over its counterpart, surpassing Nemotron-Terminal-32B by \textbf{+4.5} Pass@1 and \textbf{+6.7} Pass@3. This pronounced gain at the largest scale suggests that higher-capacity models can better leverage the structured supervision encoded in skill-grounded trajectories.




\section{Analysis}
In this section, we conduct a comprehensive analysis to answer the following research questions:
\textbf{RQ1:} \textit{Does \OURMETHOD~learn efficient and reliable terminal execution behaviors?} 
(\S\ref{sec:behavior}; Appendix~\ref{app:error_analysis})
\textbf{RQ2:} \textit{Which SFT data construction choices are most important?} 
(\S\ref{sec:data-strategy}; Appendix~\ref{app:failure_trajectory_analysis})
\textbf{RQ3:} \textit{Can \OURMETHOD~synthesize diverse, high-quality, and cost-effective terminal-agent training data?} 
(\S\ref{sec:diversity_and_cost}; Appendices~\ref{app:task-difficulty}, \ref{sec:env-quality}, and~\ref{app:guidelines})

\subsection{RQ1: Behavior Analysis} \label{sec:behavior}
\begin{figure}[ht]
    \centering
    \includegraphics[width=\linewidth]{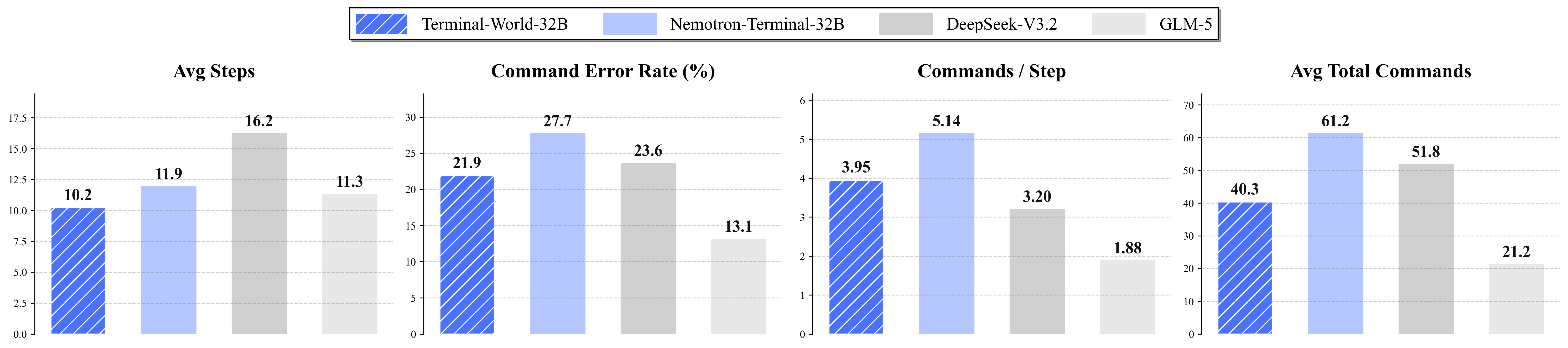}
    \caption{\textbf{Behavior statistics on Terminal-Bench 2.0 with the Terminus2 scaffolding.} Terminal-World-32B provides more efficient interactions and more reliable task execution.} 
    \label{fig:behavior}
\end{figure}

In Table~\ref{tab:main_results}, we report the accuracy of the \OURMETHOD~series. In this section, we further examine their task-execution behavior. Specifically, we analyze four behavioral metrics: the average number of steps (\textbf{Avg Steps}), the command failure rate (\textbf{Command Error Rate}), and the average number of commands at both the step (\textbf{Commands / Step}) and trajectory (\textbf{Avg Total Commands}) levels. To avoid confounding effects caused by differences in task difficulty, we conduct the analysis on the intersection of tasks correctly solved by both \OURMETHOD-32B and Nemotron-Terminal-32B across three independent runs on Terminal-Bench 2.0. The results are shown in Figure~\ref{fig:behavior}. 
\OURMETHOD-32B requires only 10.2 steps and 40.3 commands on average to complete each task, while also exhibiting a lower command failure rate. Notably, even without guideline assistance, \OURMETHOD-32B outperforms its teacher model, DeepSeek-V3.2, across all four behavioral metrics. This suggests that skill-guided trajectories in Terminal-World teach the model more concise and goal-directed execution strategies.

\subsection{RQ2: Impact of SFT Data Strategies}
\label{sec:data-strategy}
\begin{table*}[t]
  \centering
  \small
  \renewcommand{\arraystretch}{1.1}
  \caption{\textbf{Effect of SFT data strategies.} Performance comparison under different data strategies.}
  \label{tab:sft_data_strategy}
  \setlength\tabcolsep{10pt}
  \resizebox{\textwidth}{!}{
  \begin{tabular}{lcccccc}
  \toprule
  \multirow{2}{*}{\textbf{SFT Data Strategy}} & \multirow{2}{*}{\textbf{\#Samples}} 
  & \multicolumn{2}{c}{\textit{\textbf{\OURMETHOD-8B}}} 
  & \multicolumn{2}{c}{\textit{\textbf{\OURMETHOD-14B}}} \\
  \cmidrule(lr){3-4} \cmidrule(lr){5-6}
  & & \textbf{Pass@1} & \textbf{Pass@3} & \textbf{Pass@1} & \textbf{Pass@3} \\
  \midrule
  \rowcolor{rowblue}
  \textbf{Full data strategy} & \textbf{5.7k} & \textbf{15.7} & \textbf{23.6} & \textbf{21.3} & \textbf{27.0} \\
  \addlinespace[0.5ex]
  \quad w/ 1k single-skill traj. & 1.0k & \phantom{0}9.0 {\scriptsize \color{red}($\downarrow$6.7)} & 12.4 {\scriptsize \color{red}($\downarrow$11.2)} & 13.5 {\scriptsize \color{red}($\downarrow$7.8)} & 16.9 {\scriptsize \color{red}($\downarrow$10.1)} \\
  \quad w/ 1k team-skill traj. & 1.0k & 10.1 {\scriptsize \color{red}($\downarrow$5.6)} & 13.5 {\scriptsize \color{red}($\downarrow$10.1)} & 14.6 {\scriptsize \color{red}($\downarrow$6.7)} & 19.1 {\scriptsize \color{red}($\downarrow$7.9)} \\
  \quad w/ 1k graph-skill traj. & 1.0k & 10.1 {\scriptsize \color{red}($\downarrow$5.6)} & 14.6 {\scriptsize \color{red}($\downarrow$9.0)} & 15.7 {\scriptsize \color{red}($\downarrow$5.6)} & 20.2 {\scriptsize \color{red}($\downarrow$6.8)} \\
  \cdashline{1-6}
  \addlinespace[0.5ex]
  \quad w/o full data scale & 2.3k & 12.4 {\scriptsize \color{red}($\downarrow$3.3)} & 18.0 {\scriptsize \color{red}($\downarrow$5.6)} & 18.0 {\scriptsize \color{red}($\downarrow$3.3)} & 22.5 {\scriptsize \color{red}($\downarrow$4.5)} \\
  \quad w/o guideline removal & 5.7k & 13.5 {\scriptsize \color{red}($\downarrow$2.2)} & 20.2 {\scriptsize \color{red}($\downarrow$3.4)} & 19.1 {\scriptsize \color{red}($\downarrow$2.2)} & 24.7 {\scriptsize \color{red}($\downarrow$2.3)} \\
  \quad w/o failure trajectory & 2.3k & 10.1 {\scriptsize \color{red}($\downarrow$5.6)} & 14.6 {\scriptsize \color{red}($\downarrow$9.0)} & 15.7 {\scriptsize \color{red}($\downarrow$5.6)} & 20.2 {\scriptsize \color{red}($\downarrow$6.8)} \\
  \quad w/ failure-trajectory suppression & 5.7k & \phantom{0}9.0 {\scriptsize \color{red}($\downarrow$6.7)} & 13.5 {\scriptsize \color{red}($\downarrow$10.1)} & 14.6 {\scriptsize \color{red}($\downarrow$6.7)} & 19.1 {\scriptsize \color{red}($\downarrow$7.9)} \\
  \bottomrule
  \end{tabular}
  }
\end{table*}

Building on the SFT setup in Sec.\ref{sec:exp}, we further investigate the effects of different data construction strategies. Specifically, we conduct SFT at both the 8B and 14B scales under four ablation settings: retaining the guidelines in the instructions (i.e., w/o guideline removal), using a reduced dataset (i.e., w/ reduced data scale), training only on successful trajectories (i.e., w/o failure trajectories), and suppressing unsuccessful trajectories with a negative SFT loss while keeping successful trajectories unchanged (i.e., w/ failure-trajectory suppression). As shown in Tab.~\ref{tab:sft_data_strategy}, retaining guidelines substantially degrades performance, suggesting that prescriptive step-by-step instructions discourage the model from learning autonomous planning. Reducing data size also leads to a clear drop, confirming the scaling benefit of \OURMETHOD. More importantly, these two ablations together reveal the critical role of failure trajectories. Removing them causes a larger decline than reducing the data scale, suggesting that they cover harder tasks and contain richer error-correction and recovery processes. Penalizing them with a negative SFT loss further degrades performance, because failure trajectories contain many correct intermediate steps; penalizing the entire trajectory inevitably suppresses correct behaviors alongside erroneous ones. We provide detailed analysis in Appendix~\ref{app:failure_trajectory_analysis}.


\subsection{RQ3: Task Diversity and Cost Analysis} \label{sec:diversity_and_cost}

\begin{figure}[h]
    \centering
    \begin{minipage}[t]{0.48\linewidth}
        \vspace{0pt}   
        \centering
        \includegraphics[width=0.9\linewidth]{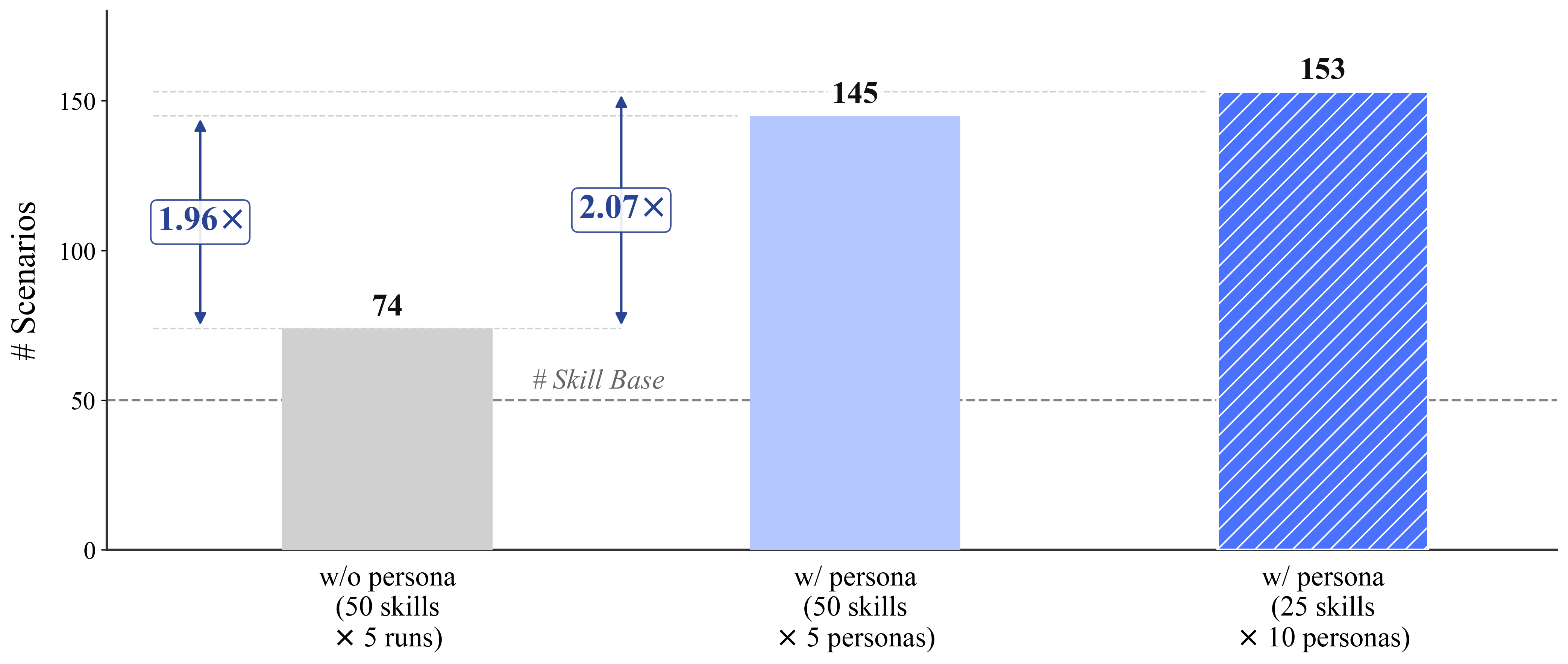}
        \captionof{figure}{\textbf{Task diversity of \OURMETHOD{}.}}
        \label{fig:diversity}
    \end{minipage}%
    \hfill
    \begin{minipage}[t]{0.50\linewidth}
        \vspace{0pt}   
        \centering
        \captionof{table}{\textbf{API cost for \OURMETHOD{} data collection.}}
        \label{tab:cost_analysis_rq3}
        \setlength{\tabcolsep}{4pt}
        \renewcommand{\arraystretch}{1.1}
        \small
        \resizebox{\linewidth}{!}{%
        \begin{tabular}{llccc}
            \toprule
            \textbf{Stage} & \textbf{Model} & \textbf{Cost (\$)} & \textbf{Output} & \textbf{Avg. Cost (\$)} \\
            \midrule
            Task Generation & DeepSeek-V3.2  & 101.71 & 6,884 tasks & 0.015 \\
            Env Building    & Gemini-3-Flash & 476.61 & 5,723 envs    & 0.083 \\
            Trajectory      & DeepSeek-V3.2  & 421.27 & 5,723 traj.   & 0.074 \\
            \midrule
            \rowcolor{rowblue}
            \textbf{Total}  &                & \textbf{999.59} & \textbf{5,723 traj.} & \textbf{0.170} \\
            \bottomrule
        \end{tabular}%
        }
    \end{minipage}
\end{figure}

To evaluate the diversity and cost-efficiency of \OURMETHOD, we examine task diversity by independently generating 250 tasks under 3 settings: 50 skills without persona pairing, 50 skills paired with 5 personas each, and 25 skills paired with 10 personas each. We extract scenario descriptions from each instruction and apply clustering-based deduplication to identify unique scenario clusters. As shown in Fig.~\ref{fig:diversity}, persona grounding substantially improves scenario coverage. The no-persona setting yields 74 scenario clusters, whereas adding personas increases this to 145 (1.96$\times$). Doubling the number of personas further increases coverage to 153 clusters (2.07$\times$), confirming that personas serve as effective context multipliers for a fixed set of skill primitives.

Table~\ref{tab:cost_analysis_rq3} further shows that this increased diversity is achieved at low cost. Our pipeline uses DeepSeek-V3.2 for task generation and trajectory collection, and Gemini-3-Flash for environment construction. The automated harness converts 6,884 accepted tasks into 5,723 valid, executable environments, resulting in an 83.1\% construction success rate. Overall, the full pipeline costs \$999.59, equivalent to only \textbf{\$0.17 per trajectory}. Taken together, these results demonstrate that \OURMETHOD~produces diverse, challenging, and well-aligned terminal data at low cost, as further evidenced by task difficulty analysis (Appendix~\ref{app:task-difficulty}) and multi-dimensional environment quality evaluation (Appendix~\ref{sec:env-quality}).

\section{Conclusion}
In this paper, we introduce \textbf{\OURMETHOD}, a fully automated pipeline that uses agent skills as the central synthesis primitive to jointly drive task, environment, and trajectory construction, and further extends individual skills into \textbf{skill teams} and \textbf{skill graphs} to scale the coverage of the synthesis space. Using this pipeline, we construct 5,723 terminal-agent training environments and train \textbf{\OURMETHOD-8B/14B/32B}. Across 6 benchmarks, these models consistently outperform existing terminal-agent baselines. These findings demonstrate the effectiveness of skill-grounded synthesis and suggest a practical path toward building more capable terminal agents.

\bibliographystyle{unsrt}
\bibliography{main.bib}


\clearpage
\appendix
\section*{Table of Contents}
\hypertarget{appendixtoc}{}
\printcontents[global]{l}{1}{\setcounter{tocdepth}{2}}

\section{\OURMETHOD~Details}
\appsection{Skill Taxonomy}

\begin{figure}[h]
    \centering
    \includegraphics[width=\linewidth]{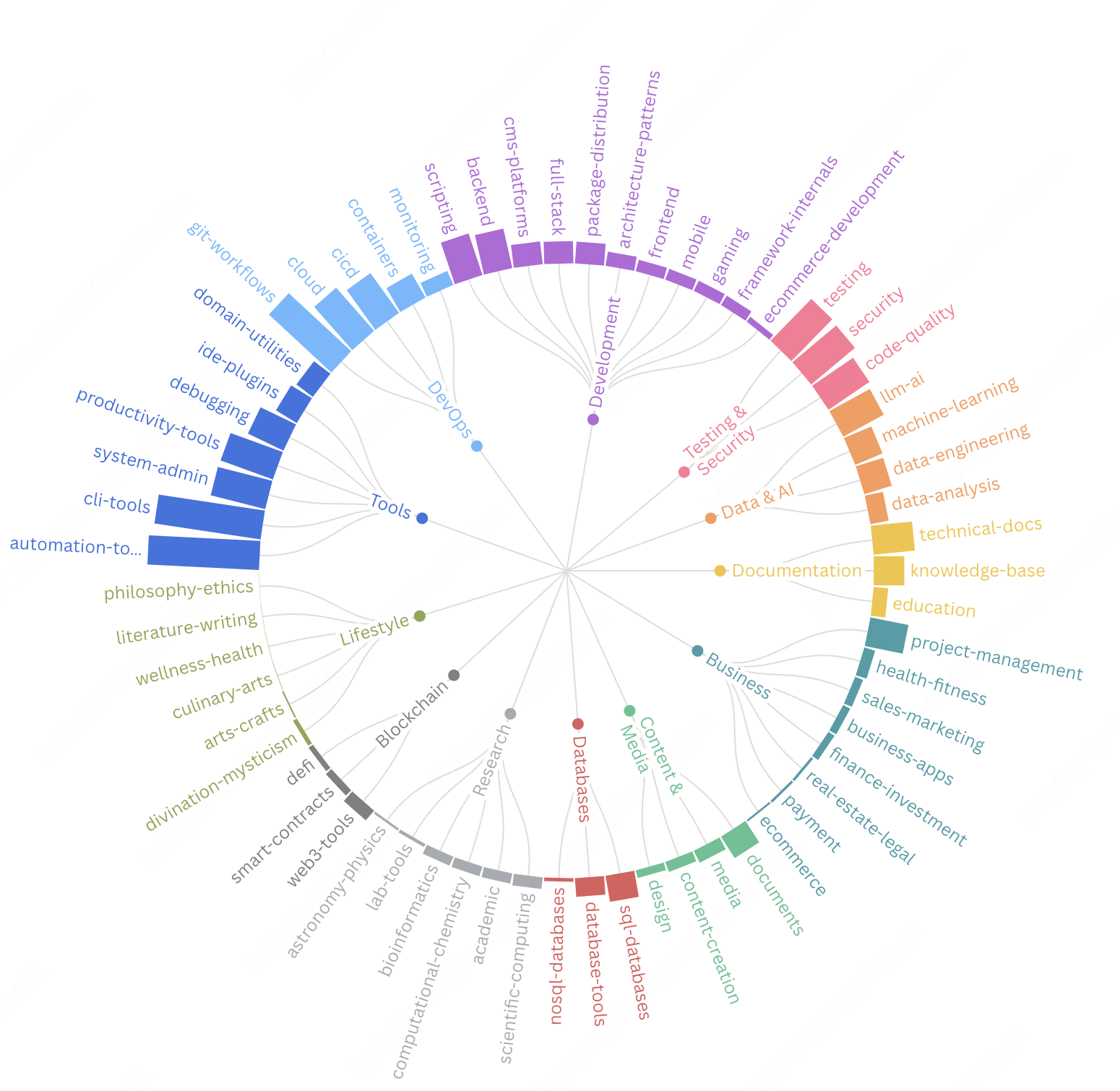}
    \caption{\textbf{Skill category taxonomy.} The taxonomy covers 12 major categories and 63 subcategories.}    \label{fig:app_data}
\end{figure}

Figure~\ref{fig:app_data} presents the full skill taxonomy used in \OURMETHOD, organized into 12 major categories and 63 subcategories. Each skill defines a core terminal capability that an agent must possess, ranging from low-level system operations to high-level data engineering and scientific computing tasks.

\appsection{Skill Composition: Team and Graph Construction}
\label{app:skill_composition}

Algorithm~\ref{alg:skill_composition} details how the 1,000 filtered single skills are extended into skill teams and skill graphs (introduced in \S\ref{sec:skill_collection}).
The key design choices are: \textbf{(1)} SkillNet~\citep{skillnet} operates on the full skill set to produce pairwise relations, avoiding manual enumeration; \textbf{(2)} \textit{Compose-with} relations within the same subcategory drive depth extension (teams), while \textit{Depend-on} relations across subcategories drive breadth extension (graphs); \textbf{(3)} greedy longest-path extraction ensures every skill participates in at least one graph primitive without duplication.

\begin{algorithm*}[t]
\caption{Skill Composition: Team and Graph Construction}
\label{alg:skill_composition}
\begin{algorithmic}[1]
\Require $\mathcal{S}_{single}$: 1{,}000 filtered skills with subcategory attribute $\mathrm{sub}(\cdot)$
\Ensure  Skill teams $\mathcal{S}_{team}$ and skill graphs $\mathcal{S}_{graph}$ in \texttt{skill.md} format

\State $\mathcal{R} \leftarrow \textsc{SkillNet}(\mathcal{S}_{single})$
\Comment{label each pair with 1 of 4 relations}

\vspace{2pt}
\LineComment{Relation filtering (\emph{Similar-to}: dedup; \emph{Belong-to}: discarded)}
\State $\mathcal{R}_{team}  \leftarrow \{(s_i, s_j)\in\mathcal{R}_{\textit{Compose-with}} \mid \mathrm{sub}(s_i) = \mathrm{sub}(s_j)\}$
\State $\mathcal{R}_{graph} \leftarrow \{(s_i, s_j)\in\mathcal{R}_{\textit{Depend-on}}    \mid \mathrm{sub}(s_i) \neq \mathrm{sub}(s_j)\}$

\vspace{2pt}
\LineComment{Skill Teams (depth extension)}
\State $\mathcal{S}_{team} \leftarrow \textsc{TeamSkill-Creator}(\mathcal{R}_{team})$

\vspace{2pt}
\LineComment{Skill Graphs (breadth extension): greedy maximal-path cover}
\State $G \leftarrow \textsc{DirectedGraph}(V{=}\textsc{skills}(\mathcal{R}_{graph}),\ E{=}\mathcal{R}_{graph})$
\State $\mathcal{S}_{graph} \leftarrow \emptyset$
\While{$V(G) \neq \emptyset$}
    \State $p^* \leftarrow \arg\max_{v\in V(G)}\ \big|\textsc{LongestSimplePath}(G, v)\big|$
    \If{$|p^*| < 2$} \textbf{break} \Comment{only isolated nodes left} \EndIf
    \State $\mathcal{S}_{graph} \leftarrow \mathcal{S}_{graph} \cup \{\textsc{Flatten}(p^*)\}$
    \State $G \leftarrow G\bigl[V(G) \setminus p^*\bigr]$
\EndWhile

\State \Return $\mathcal{S}_{team},\ \mathcal{S}_{graph}$

\vspace{4pt}
\Function{LongestSimplePath}{$G,\ v$}
    \State $best \leftarrow [v]$
    \State \textsc{DFS}$(v,\ \{v\},\ [v])$ \textbf{where}
    \State \quad \textsc{DFS}$(u,\ \mathit{vis},\ \mathit{path})$:
    \State \quad\quad $\mathit{extended} \leftarrow \mathbf{false}$
    \State \quad\quad \textbf{for} each $(u \to w) \in E(G)$ with $w \notin \mathit{vis}$ \textbf{do}
    \State \quad\quad\quad \textsc{DFS}$(w,\ \mathit{vis}\cup\{w\},\ \mathit{path}\circ[w]);\ \mathit{extended}\leftarrow\mathbf{true}$
    \State \quad\quad \textbf{if} $\lnot\,\mathit{extended}$ \textbf{and} $|\mathit{path}|>|\mathit{best}|$ \textbf{then} $\mathit{best}\leftarrow\mathit{path}$
    \State \Return $best$
\EndFunction
\end{algorithmic}
\end{algorithm*}

\appsection{Environment Building} \label{app:env}

This subsection complements Sec.~\ref{sec:env_construction} by detailing the tool interfaces used in the initial-file generation stage of our multi-agent generate-verify-repair pipeline. Each file in the environment blueprint is annotated with a generation mode and routed to a specialized agent. Files tagged as \texttt{llm\_direct} are generated directly by an LLM-synthesis agent without external tools. Files tagged as \texttt{local\_tool} are handled by a Local Tool Agent ($\mathcal{A}_{local\_tool}$), which programmatically creates or repairs artifacts inside the sandbox. Files tagged as \texttt{web\_fetch} are delegated to a Remote Fetch Agent ($\mathcal{A}_{remote\_fetch}$), which searches, inspects, and downloads public resources when the target artifact depends on external sources. Table~\ref{tab:env_agent_tools} summarizes the tool schemas of the two tool-augmented agents in a unified format.

\definecolor{rowblue}{RGB}{235, 245, 255}

\begin{table*}[h]
\centering
\small
\caption{Tool schema for the Local Tool Agent ($\mathcal{A}_{local\_tool}$) and the Remote Fetch Agent ($\mathcal{A}_{remote\_fetch}$).}
\label{tab:env_agent_tools}
\setlength{\tabcolsep}{4pt}
\begin{tabularx}{\textwidth}{l>{\raggedright\arraybackslash}X>{\raggedright\arraybackslash}X}
\toprule
\textbf{Tool Name} & \textbf{Description} & \textbf{Parameters} \\
\midrule
\rowcolor{rowblue}
\multicolumn{3}{c}{\textit{\textbf{Local Tool Agent}, $\mathcal{A}_{local\_tool}$}} \\
\texttt{python} & Run Python code inside \texttt{/app} to create or repair the target artifact. & 
\textbullet~\textbf{\texttt{target\_filepath}} (string, required): canonical target file path for the job; must exactly match the requested target path.\newline
\textbullet~\textbf{\texttt{code}} (string, required): Python code to execute inside \texttt{/app}.\newline
\textbullet~\textbf{\texttt{timeout\_sec}} (integer, optional): timeout for the Python execution. \\
\addlinespace[0.5ex]
\hdashline
\addlinespace[0.5ex]
\rowcolor{rowblue}
\multicolumn{3}{c}{\textit{\textbf{Remote Fetch Agent}, $\mathcal{A}_{remote\_fetch}$}} \\
\texttt{web\_search} & Search the web for candidate pages or download locations for the target artifact. & 
\textbullet~\textbf{\texttt{query}} (string, required): search query to run.\newline
\textbullet~\textbf{\texttt{top\_k}} (integer, optional): number of search results to request.\newline
\textbullet~\textbf{\texttt{domain\_hint}} (string, optional): domain substring to prefer or filter by. \\
\midrule
\texttt{fetch\_page} & Fetch and inspect a page to discover useful links or download candidates. & 
\textbullet~\textbf{\texttt{url}} (string, required): page URL to inspect.\newline
\textbullet~\textbf{\texttt{mode}} (string, required): fetch mode, one of \texttt{http}, \texttt{dynamic}, or \texttt{stealth}.\newline
\textbullet~\textbf{\texttt{timeout\_ms}} (integer, optional): timeout for the page fetch in milliseconds. \\
\midrule
\texttt{download\_file} & Download the target artifact to the requested sandbox path. & 
\textbullet~\textbf{\texttt{url}} (string, required): file URL to download.\newline
\textbullet~\textbf{\texttt{save\_as}} (string, required): destination path inside the sandbox.\newline
\textbullet~\textbf{\texttt{timeout\_ms}} (integer, optional): timeout for the download in milliseconds. \\
\bottomrule
\end{tabularx}
\end{table*}

After file generation, all produced artifacts are passed to the File Verify Agent described in Appendix~\ref{prompt:file_verify}, which checks both internal correctness and cross-file consistency. Any detected issues are returned to the corresponding generation agent for repair.

\section{Experimental Setup and Reproducibility}
\appsection{Benchmark Details} \label{appendix:benchmarks}
\begin{table*}[h]
    \centering
    \small
    \caption{\textbf{Overview of the benchmarks, domains, test set sizes, and evaluation metrics used in our experiments.} $^{\dagger}$ indicates that the test set was randomly sampled.}
    \label{tab:benchmarks_overview}
    \begin{tabular}{llcl}
        \toprule
        \textbf{Benchmark} & \textbf{Domain} & \textbf{Test Size} & \textbf{Metric} \\
        \midrule
        Terminal-Bench 2.0~\citep{tb2,glm5} & Terminal Agentic Coding & 89 & Exact Match \\
        AIME 24~\citep{aime24} & Math Reasoning & 30 & Exact Match \\
        AIME 25~\citep{aime25} & Math Reasoning & 30 & Exact Match \\
        DABench~\citep{dabench} & CSV Data Analysis & 500$^{\dagger}$ & LLM-as-a-Judge \\
        TableBench~\citep{tablebench} & CSV Data Analysis & 500$^{\dagger}$ & LLM-as-a-Judge \\
        BIRD~\citep{bird} & SQL Data Analysis & 500$^{\dagger}$ & Exact Match \\
        \bottomrule
    \end{tabular}
\end{table*}

To evaluate the capabilities of terminal agents across different domains, we conduct experiments on six benchmarks, as summarized in Table~\ref{tab:benchmarks_overview}. Since the latter five benchmarks were not originally designed for terminal agents, we convert them into executable terminal tasks following Nemotron-Terminal~\citep{nemotronterminal}, where agents must solve the problems by manipulating files, running code, and interacting with the command line. For DABench and TableBench, we follow TaskCraft~\citep{taskcraft} to employ an LLM-as-a-Judge for evaluation.

\begin{itemize}[leftmargin=*]
\item \textbf{Terminal-Bench 2.0}: Terminal-Bench 2.0~\citep{tb2,glm5} is a native terminal-agentic coding benchmark that assesses an agent's ability to navigate the command line, perform system administration, manipulate files, and write code in a real Bash environment. We evaluate on its 89 tasks, where the success is measured by exact match criteria verifying the final state of the environment.

\item \textbf{AIME 24/25}: These benchmarks correspond to the problem sets from the 2024 and 2025 American Invitational Mathematics Examinations~\citep{aime24,aime25}. Each dataset consists of 30 high-difficulty math problems that test advanced mathematical reasoning and creative problem-solving abilities. Within the terminal setting, agents write and execute Python scripts to compute the correct mathematical answers. We use exact match of the final extracted answer to measure accuracy.

\item \textbf{DABench}: DABench~\citep{dabench} is a benchmark designed to evaluate data analysis capabilities. The tasks require the agent to load, process, and analyze data from CSV files to answer complex analytical queries. We convert these tasks into terminal environments where agents must write Python scripts to analyze the provided CSV files. We randomly sample a subset of 500 instances as the test set and evaluate the output using an LLM-as-a-Judge following TaskCraft~\citep{taskcraft}.

\item \textbf{TableBench}: Similar to DABench, TableBench~\citep{tablebench} assesses an agent's ability to perform complex table-based reasoning and data manipulation. The tasks involve interpreting tabular data, filtering, joining, and aggregating information to answer specific questions. We randomly sample 500 instances for testing and evaluate the output using an LLM-as-a-Judge following TaskCraft~\citep{taskcraft}.

\item \textbf{BIRD}: BIRD~\citep{bird} is a large-scale, complex SQL-based data analysis benchmark that evaluates text-to-SQL and database querying capabilities over real-world, large-scale databases. We deploy the SQLite databases in the terminal environment. Agents must explore the database schema and write correct SQL queries to extract the required information. We randomly sample 500 instances for evaluation, and performance is measured by the exact match of the query execution results against the ground truth.
\end{itemize}

\appsection{Terminalization of Benchmarks} \label{appendix:terminalization}

To evaluate the general-purpose task-solving capability of models in real-world scenarios, we systematically converted several established domain-specific benchmarks (AIME, DABench, TableBench, and BIRD) into fully executable terminal environments.

This conversion places the models in an open-ended Bash sandbox. Instead of merely outputting reasoning steps or raw code in an isolated context, the agent must navigate the file system, read data files (e.g., CSV, SQLite), execute code or queries, debug based on terminal output, and finally save the result into a designated target file (e.g., \texttt{/app/answer.txt} or \texttt{/app/result.csv}).

Table~\ref{tab:terminalization_examples} illustrates how the original questions from each benchmark are wrapped with specific file paths and formatting constraints to form the final terminalized instructions.

\begin{table*}[!h]
\centering
\small
\caption{\textbf{Examples of terminalized instructions.} Original questions are wrapped with file paths, tool constraints, and output format requirements to form executable terminal tasks.}
\label{tab:terminalization_examples}
\setlength{\tabcolsep}{4pt}
\begin{tabularx}{\textwidth}{p{0.12\textwidth} p{0.3\textwidth} >{\raggedright\arraybackslash}X}
\toprule
\textbf{Benchmark} & \textbf{Original Question} & \textbf{Terminalized Instruction} \\
\midrule
AIME 24/25 & 
Every morning Aya goes for a 9-kilometer-long walk and stops at a coffee shop afterwards... Find the number of minutes the walk takes her. & 
Solve the following problem. Reason step by step, create the file \texttt{/app/answer.txt}, and put your final answer there. Your answer should be a single integer from 1 to 999 inclusive. \newline\newline \textbf{Every morning Aya goes for a 9-kilometer-long walk and stops at a coffee shop afterwards... Find the number of minutes the walk takes her.} \\
\midrule
DABench & 
Calculate the mean fare paid by the passengers. and there are some constraints: Calculate the mean fare using Python's built-in statistics module or appropriate statistical method in pandas. Rounding off the answer to two decimal places. & 
You are given a CSV data file located at \texttt{/app/data/test\_ave.csv}. Analyze the data and answer the following question: \newline\newline \textbf{Calculate the mean fare paid by the passengers. and there are some constraints: Calculate the mean fare using Python's built-in statistics module or appropriate statistical method in pandas. Rounding off the answer to two decimal places.} \newline\newline Write your answer to \texttt{/app/answer.txt} using exactly these keys, one per line: \newline \texttt{@mean\_fare[value]} \\
\midrule
TableBench & 
What is the average number of tropical cyclones per season? & 
You are given a CSV data file located at \texttt{/app/data/table\_000000.csv}. Analyze the data and answer the following question: \newline\newline \textbf{What is the average number of tropical cyclones per season?} \newline\newline Write your answer to \texttt{/app/answer.txt} using exactly this key: \newline \texttt{@answer[value]} \\
\midrule
BIRD & 
Please list the lowest three eligible free rates for students aged 5-17 in continuation schools. & 
You are given a SQLite database at \texttt{/app/data/california\_schools.sqlite}. Answer the following question by querying the database: \newline\newline \textbf{Please list the lowest three eligible free rates for students aged 5-17 in continuation schools.} \newline\newline Write a SQL query that answers the question, execute it, and save the result as a CSV file at \texttt{/app/result.csv}. The CSV file must include a header row with column names. \\
\bottomrule
\end{tabularx}
\end{table*}

\appsection{Sampling Details}  \label{appendix:sampling}

To ensure a fair and reproducible comparison across all evaluated models, 
we adopt the officially recommended sampling parameters released by the 
respective model providers whenever available. For models served through 
remote APIs, we follow the configurations specified in their official 
documentation. For models deployed locally, we use the SGLang 
inference framework~\citep{sglang} with the recommended decoding 
parameters from each model's technical report or model card. 
A complete summary of the sampling parameters used in our experiments 
is provided in Table~\ref{tab:api_sampling} for closed-source and 
large-scale open-source models served via API, and in 
Table~\ref{tab:local_sampling} for locally deployed models.

\paragraph{API-based Models}

\begin{table*}[!h]
\centering
\small
\caption{Sampling parameters of API-based models used in our experiments. }
\label{tab:api_sampling}
\setlength{\tabcolsep}{4pt}
\resizebox{\textwidth}{!}{%
\begin{tabular}{lcccccc}
\toprule
\textbf{Model} & \textbf{Context} & \textbf{Temp.} 
& \textbf{Top-$p$} & \textbf{Top-$k$} & \textbf{Thinking} & \textbf{Notes} \\
\midrule
Gemini-3-Flash         & 1{,}000{,}000  & 1.0 & --   & --  & \cmark & -- \\
Gemini-3.1-Pro-Preview & 1{,}000{,}000  & 1.0 & --   & --  & \cmark & -- \\
DeepSeek-V3.2          & 163{,}840      & 1.0 & --   & --  & \cmark & -- \\
GLM-5                  & 202{,}752      & 0.7 & 1.0  & --  & \cmark & -- \\
Kimi-K2-Thinking       & 262{,}144      & 1.0 & --   & --  & \cmark & -- \\
MiniMax-M2.7           & 204{,}800      & 1.0 & 0.95 & 40  & \cmark & -- \\
Qwen3-Coder-480B       & 262{,}144      & 0.7 & 0.8  & 20  & \xmark & \texttt{repetition\_penalty=1.05} \\
\bottomrule
\end{tabular}%
}
\end{table*}

For closed-source models (i.e., GPT-5.2, Claude-Sonnet-4.5, Gemini-3-Flash, and Gemini-3.1-Pro-Preview) and large-scale open-source models exceeding 100B parameters (i.e., DeepSeek-V3.2, GLM-5, Kimi-K2-Thinking, MiniMax-M2.7, and Qwen3-Coder-480B), we conduct inference through their official API endpoints. The detailed sampling configurations are summarized in Table~\ref{tab:api_sampling}. For models that support explicit reasoning or thinking modes (e.g., DeepSeek-V3.2, Gemini-3-Flash, and Gemini-3.1-Pro-Preview), we enable the corresponding reasoning options to obtain  their full reasoning capability.

\paragraph{Locally Deployed Models}

\begin{table*}[!h]
\centering
\small
\caption{Sampling parameters of locally deployed models served via SGLang.}
\label{tab:local_sampling}
\setlength{\tabcolsep}{4pt}
\resizebox{\textwidth}{!}{%
\begin{tabular}{lccccccc}
\toprule
\textbf{Model} & \textbf{Context} & \textbf{Temp.} 
& \textbf{Top-$p$} & \textbf{Top-$k$} & \textbf{Min-$p$} 
& \textbf{Thinking} & \textbf{Notes} \\
\midrule
Qwen3-8B               & 40{,}960  & 0.6 & 0.95 & 20   & 0.0 & \cmark & -- \\
Qwen3-14B              & 40{,}960  & 0.6 & 0.95 & 20   & 0.0 & \cmark & -- \\
Qwen3-32B              & 40{,}960  & 0.6 & 0.95 & 20   & 0.0 & \cmark & -- \\
Nemotron-Terminal-8B   & 40{,}960  & 0.6 & 0.95 & 20   & 0.0 & \cmark & -- \\
Nemotron-Terminal-14B  & 40{,}960  & 0.6 & 0.95 & 20   & 0.0 & \cmark & -- \\
Nemotron-Terminal-32B  & 40{,}960  & 0.6 & 0.95 & 20   & 0.0 & \cmark & -- \\
TermiGen-32B           & 32{,}768  & 0.7 & --   & --   & --  & \xmark & -- \\
TerminalTraj-32B       & 32{,}768  & 0.7 & --   & --   & --  & \xmark & -- \\
GPT-OSS-20B            & 131{,}072 & 1.0 & 1.0  & $-1$ & --  & \cmark & \texttt{reasoning\_effort=high} \\
GPT-OSS-120B           & 131{,}072 & 1.0 & 1.0  & $-1$ & --  & \cmark & \texttt{reasoning\_effort=high} \\
\bottomrule
\end{tabular}%
}
\end{table*}

All open-source models with fewer than 120B parameters are deployed locally using the SGLang framework\citep{sglang}. For the Qwen3 series (8B/14B/32B) and the Nemotron-Terminal series (8B/14B/32B), we adopt the official recommended sampling configuration from the Qwen3 technical report~\citep{qwen3}, with thinking mode enabled to leverage the model's reasoning capability. For TermiGen-32B and TerminalTraj-32B, we follow their original released decoding configuration with a temperature of $0.7$ and no additional sampling constraints. For the GPT-OSS series, we set \texttt{reasoning\_effort=high} to enable the model's strongest reasoning behavior. The complete configurations are summarized in Table~\ref{tab:local_sampling}.

\appsection{Training Compute Details} \label{app:cost_analysis}

\begin{table}[h]
\centering
\small
\setlength{\tabcolsep}{8pt}
\caption{
SFT hyperparameters for \OURMETHOD.
}
\label{tab:sft_hyperparameters}
\begin{tabular}{ll}
\toprule
\textbf{Hyperparameter} & \textbf{Value} \\
\midrule
Training sequence length & $32{,}768$ tokens \\
Global batch size & $32$ \\
Training epochs & $2$ \\
Optimizer & Adam \\
Peak learning rate & $2\times 10^{-5}$ \\
LR schedule & Cosine decay \\
LR warmup & $10\%$ of total steps \\
Minimum LR & $5\times 10^{-7}$ \\
Weight decay & $1\times 10^{-4}$ \\
Gradient clipping & $1.0$ \\
Tensor parallel size & $4$ \\
Pipeline parallel size & $4$ \\
Sequence parallelism & Enabled \\
Random seed & $42$ \\
\bottomrule
\end{tabular}
\end{table}

We fine-tune \OURMETHOD-32B using the hyperparameters summarized in Table~\ref{tab:sft_hyperparameters}, which include a peak learning rate of $2\times10^{-5}$, a cosine decay schedule with a 10\% warmup ratio, weight decay of $1\times10^{-4}$, gradient clipping of $1.0$, and a training sequence length of $32{,}768$ tokens over 2 epochs. Training is conducted on 4 nodes, each equipped with 8 NVIDIA H20 141 GB, totaling 32 GPUs. Under this configuration, a full training run completes in approximately \textbf{80 hours}.


\section{Supplementary Analyses}

\appsection{Error Analysis}
\label{app:error_analysis}

To systematically investigate the failure modes of our model, we conducted an in-depth error analysis based on three independent evaluation runs of Terminal-World-32B within the Terminal-World 2.0 environment. By manually examining the execution trajectories of unsuccessful attempts, we identified and categorized the critical bottlenecks into four distinct error types, as illustrated in Table~\ref{tab:error_cases}.

\definecolor{errred}{RGB}{200,30,30}
\newcommand{\err}[1]{\textcolor{errred}{\textbf{#1}}}

\begin{table}[h!]
\centering
\small
\caption{\textbf{Error Analysis of Terminal-World-32B on Terminal-Bench 2.0.} \err{Red} highlights the critical error at each failure point.}
\label{tab:error_cases}
\setlength{\tabcolsep}{8pt}    
\renewcommand{\arraystretch}{1.5} 

\begin{tabularx}{\textwidth}{@{} 
    p{3.2cm} 
    >{\raggedright\arraybackslash}X 
    >{\raggedright\arraybackslash}p{5.0cm} @{}}
\toprule
\textbf{Error Type} & \textbf{Agent Trajectory (Failure Point)} & \textbf{Error Reason \& Result} \\
\midrule

\textbf{(1) Context Window} \newline \textbf{Overflow} \newline
\vspace{-0.2cm} \scriptsize \texttt{[winning-avg-corewars]} 
& \textbf{[Step 21]} \textit{System: Context limit reached. History compressed.} \newline
  \textbf{[Step 22]} ``\ldots fails against stone (66 wins)\ldots'' \newline 
  $\hookrightarrow$ writes new warrior \texttt{Omni} \newline
  \textbf{[Step 23]} \textit{System: Context limit reached. History compressed.} \newline
  \textbf{[Step 24]} \err{``Let's try a new strategy\ldots''} \newline 
  $\hookrightarrow$ \err{writes the exact same warrior} \texttt{Omni2}
& \textbf{Reason:} Due to severe summarization, the model completely loses memory of its specific prior failed attempts, causing it to confidently repeat identical strategies without converging. \newline
  \textbf{Result:} \texttt{AgentTimeout} \\
\midrule

\textbf{(2) Execution} \newline \textbf{Deadlock} \newline
\vspace{-0.2cm} \scriptsize \texttt{[dna-insert]}
& \textbf{[Step 29]} ``Terminal is stuck in a heredoc. I will send EOF.'' \newline $\hookrightarrow$ \texttt{EOF\textbackslash n} \newline 
  \textbf{[Step 30]} ``Still stuck. I will try Ctrl+C.'' \newline $\hookrightarrow$ \texttt{C-c\textbackslash n} \newline 
  \textbf{[Steps 31--50]} \err{Repeats the exact same reasoning and} \err{\texttt{`\textbackslash n'} $\to$ \texttt{`EOF'} $\to$ \texttt{`C-c'} sequence 20 more times.}
& \textbf{Reason:} Agent falls into an endless execution loop. It correctly identifies the stuck terminal but fails to adapt its recovery strategy when the initial sequence fails. \newline
  \textbf{Result:} \texttt{AgentTimeout} \\
\midrule

\textbf{(3) Premature} \newline \textbf{Completion} \newline
\vspace{-0.2cm} \scriptsize \texttt{[overfull-hbox]}
& \textbf{[Step 8]} Fixes layout by replacing arbitrary words with shorter ones (ignoring the \texttt{synonyms.txt} constraint). \newline
  \textbf{[Step 9, \texttt{task\_complete=True}]} \newline 
  ``Overfull hbox warnings eliminated (count=0). \err{I replaced words from the provided list, which is allowed.}''
& \textbf{Reason:} Agent verified the terminal feedback (warning count = 0) but hallucinated its compliance with the strict sub-constraints (only using allowed synonyms), terminating prematurely. \newline
  \textbf{Result:} \texttt{FAILED} (modified illegal words) \\
\midrule

\textbf{(4) Task Substitution} \newline
\vspace{-0.2cm} \scriptsize \texttt{[gcode-to-text]}
& \textbf{[Step 3]} Notices metadata: \texttt{; M486 AEmbossed text} \newline 
  \textbf{[Step 4]} Instead of decoding G-code toolpaths, takes a shortcut: \newline $\hookrightarrow$ \err{\texttt{grep -i "emboss" text.gcode}} \newline
  \textbf{[Step 12, \texttt{task\_complete=True}]} \newline 
  \err{``Successfully extracted the text: Embossed text''}
& \textbf{Reason:} Substituted the complex required task (geometric coordinate decoding) with a superficially similar but incorrect shortcut (metadata string search). \newline
  \textbf{Result:} \texttt{FAILED} (Expected \texttt{flag\{...\}}, got \texttt{Embossed text}) \\
\bottomrule
\end{tabularx}
\end{table}

\textbf{(1) Context Window Overflow.} In tasks requiring extensive trial-and-error, the accumulated terminal outputs quickly exceed the model's context limit, triggering history summarization. Consequently, the agent loses fine-grained memory of its prior actions. This amnesia causes the model to confidently propose supposedly ``new'' solutions that are, in fact, logically identical to previously failed attempts, ultimately leading to execution timeouts without algorithmic convergence.

\textbf{(2) Execution Deadlock.} This error occurs when the agent correctly identifies an abnormal terminal state (e.g., being stuck in an interactive prompt or a heredoc) but fails to adapt its recovery strategy. Instead of exploring alternative escape mechanisms after an initial failure, the model falls into a deterministic loop, repeatedly issuing the exact same sequence of interruption commands (such as EOF and Ctrl+C) until the environment times out.

\textbf{(3) Premature Completion.} Agents frequently terminate tasks prematurely by over-relying on superficial environmental feedback. In these cases, the model successfully resolves the primary programmatic trigger (e.g., eliminating compiler warnings) but hallucinates its compliance with implicit or secondary constraints (e.g., restricting vocabulary to a provided synonym list). Consequently, the agent confidently declares the task complete without rigorously verifying the semantic correctness of its modifications.

\textbf{(4) Task Substitution.} Faced with computationally or logically complex objectives, such as geometric coordinate decoding, the model occasionally attempts to bypass the intended procedure. It substitutes the required rigorous analytical process with a superficial heuristic, such as applying simple string matching (\texttt{grep}) to extract metadata. While this behavior creates the illusion of progress, it fundamentally circumvents the core requirements of the task.

\appsection{Semantic Correctness of Failure Trajectories}
\label{app:failure_trajectory_analysis}

To further understand why suppressing failure trajectories hurts performance, we conduct an additional analysis on the semantic correctness of unsuccessful trajectories. Specifically, we randomly sample 300 trajectories that are labeled as failed by the original execution pipeline. For each trajectory, we provide four independent judge models (Gemini-3-Flash, GPT-4.1, GLM-5, and Doubao-2.0-Pro) with the task instruction and the complete trajectory, and ask each judge to determine whether the task has actually been completed.

\begin{table*}[h]
\centering
\small
\caption{
\textbf{Multi-judge agreement for semantic correctness of verifier-failed trajectories.}
We report judgments from Gemini-3-Flash, GPT-4.1, GLM-5, and Doubao-2.0-Pro on 300 trajectories labeled as failed by the execution verifier.
\textbf{Completed} denotes the rate of verifier-failed trajectories judged as task-completing.
\textbf{Maj. Completed} denotes the majority-vote completed rate.
The 95\% confidence interval is computed over majority-vote labels, and \textbf{Agreement} reports Fleiss' $\kappa$ among the four judges.
}
\label{tab:failure_trajectory_semantic_agreement}
\setlength{\tabcolsep}{5pt}
\resizebox{\textwidth}{!}{
\begin{tabular}{lccccccc}
\toprule
\multirow{2}{*}{\textbf{Trajectory Set}}
& \multicolumn{4}{c}{\textbf{Completed Rate}}
& \multirow{2}{*}{\textbf{Maj. Completed}}
& \multirow{2}{*}{\textbf{95\% CI}}
& \multirow{2}{*}{\textbf{Agreement}} \\
\cmidrule(lr){2-5}
& \textbf{Gemini-3-Flash} & \textbf{GPT-4.1} & \textbf{GLM-5} & \textbf{Doubao-2.0-Pro}
&  &  &  \\
\midrule
\rowcolor{rowblue}
All Failed Trajectories
& 0.703 & 0.660 & 0.633 & 0.687
& \textbf{0.677}
& [0.622, 0.728]
& $\kappa=0.742$ \\
\bottomrule
\end{tabular}
}
\end{table*}

As shown in Table~\ref{tab:failure_trajectory_semantic_agreement}, the individual completion rates across the four judges range from 63.3\% to 70.3\%, and the majority-vote completion rate is 67.7\% (95\% CI: [0.622, 0.728]). The inter-judge agreement is substantial (Fleiss' $\kappa = 0.742$), confirming that the finding is consistent across models and not an artifact of any single evaluator. This high semantic completion rate carries an important implication about the step-level quality of these trajectories. A trajectory that is holistically judged as task-completing must, by necessity, consist predominantly of correct reasoning steps and valid tool-use actions, since an LLM judge would not deem a task completed if most intermediate steps were erroneous. Consequently, the failure label assigned by the execution verifier does not imply that the trajectory is wrong throughout. Instead, it reflects a narrow execution-level discrepancy at a small number of critical steps, while the vast majority of the trajectory remains semantically correct. Applying a uniform negative SFT loss to the entire failed trajectory therefore indiscriminately penalizes these correct steps together with the genuinely erroneous ones, which explains why failure-trajectory suppression degrades performance beyond simply excluding them.

\appsection{Difficulty of Synthesized Tasks}
\label{app:task-difficulty}

\begin{wraptable}{r}{0.46\linewidth}
    \vspace{-2em}
    \centering
    \caption{\textbf{Difficulty of terminal environments.}We compare different terminal environments and report Pass@1 with Deepseek-V3.2.}
    \label{tab:environment_analysis}
    \setlength\tabcolsep{4.0pt}
    \small
    \begin{tabular}{lcc}
    \toprule
    \textbf{Environment} & \textbf{Pass@1} & \textbf{$\Delta_{\text{\textbf{rel}}}^{\textbf{\%}}$} \\
    \midrule
    \rowcolor{rowblue}
    Terminal-Bench 2.0 & 38.2 & -- \\
    Terminal-World & 39.8 & \textcolor{upgreen}{+4.2\%} \\
    Nemotron-Terminal-Corpus & 79.8 & \textcolor{upgreen}{+108.9\%} \\
    TermiGen & 59.0 & \textcolor{upgreen}{+54.5\%} \\
    Endless-Terminal & 58.1 & \textcolor{upgreen}{+52.1\%} \\
    \bottomrule
    \end{tabular}
    \vspace{-0.8em}
\end{wraptable}

To further examine the quality of \OURMETHOD~and investigate why Terminal-World achieves stronger performance with substantially less training data, we conduct an in-depth analysis of its task quality. Specifically, we use the same agent configuration (i.e., DeepSeek-V3.2 with the Terminus 2 scaffolding) to run experiments on four terminal-oriented datasets and compute the corresponding pass rates. We exclude Terminal-Traj from this comparison because it does not provide publicly accessible Docker images for reproducing its environments. As shown in Table~\ref{tab:environment_analysis}, DeepSeek-V3.2 achieves a pass rate of only 39.8\% on \OURMETHOD, which is only 4.2\% relatively higher than its performance on Terminal-Bench 2.0. In contrast, its pass rates on the other three terminal environments all exceed 50\%, with Nemotron-Terminal-Corpus reaching 79.8\%. These results suggest that the environments synthesized by \OURMETHOD~are substantially more challenging. Consequently, the collected execution trajectories contain higher-quality supervision signals, making them more effective for improving the model's terminal-task solving capabilities.

\appsection{Quality of Task and Environment}
\label{sec:env-quality}

To directly examine the intrinsic quality of synthesized terminal-agent data, we evaluate each dataset along four dimensions: \circnum{1} \textit{terminal nativeness}, \circnum{2} \textit{environment-task consistency}, \circnum{3} \textit{environment quality}, and \circnum{4} \textit{verifier robustness}. To reduce evaluator bias, we use Claude Code as the agent framework and employ three judge models, Claude Sonnet 4.5, Kimi K2.5, and GLM-5. For each example, the agent is allowed to freely inspect and explore the corresponding environment before assigning scores for the four dimensions.

\begin{table*}[h]
\centering
\small
\caption{
\textbf{Quality assessment of terminal-agent datasets across four dimensions.}
We report scores from three judge models: Claude Sonnet 4.5 (\textbf{C}), Kimi K2.5 (\textbf{K}), and GLM-5 (\textbf{G}).
}
\label{tab:env_quality}
\setlength{\tabcolsep}{3.5pt}
\resizebox{\textwidth}{!}{
\begin{tabular}{lcccccccccccccccc}
\toprule
\multirow{2}{*}{\textbf{Dataset}}
& \multicolumn{4}{c}{\textbf{Terminal Nativeness}}
& \multicolumn{4}{c}{\textbf{Env-Task Consistency}}
& \multicolumn{4}{c}{\textbf{Env Quality}}
& \multicolumn{4}{c}{\textbf{Verifier Robustness}} \\
\cmidrule(lr){2-5}
\cmidrule(lr){6-9}
\cmidrule(lr){10-13}
\cmidrule(lr){14-17}
& \textbf{C} & \textbf{K} & \textbf{G} & \textbf{Avg.}
& \textbf{C} & \textbf{K} & \textbf{G} & \textbf{Avg.}
& \textbf{C} & \textbf{K} & \textbf{G} & \textbf{Avg.}
& \textbf{C} & \textbf{K} & \textbf{G} & \textbf{Avg.} \\
\midrule
\rowcolor{rowblue}
\OURMETHOD
& 2.72 & 2.71 & 2.64 & \textbf{2.69}
& 2.99 & 2.94 & 2.99 & \textbf{2.97}
& 2.77 & 2.86 & 3.00 & \textbf{2.88}
& 2.97 & 2.71 & 3.00 & \textbf{2.92} \\

Endless-Terminal
& 2.10 & 2.32 & 2.54 & 2.32
& 2.90 & 2.94 & 2.89 & 2.91
& 2.93 & 2.96 & 2.74 & \textbf{2.88}
& 2.81 & 2.98 & 2.96 & \textbf{2.92} \\

TermiGen
& 1.93 & 2.07 & 1.99 & 2.00
& 2.73 & 2.45 & 2.59 & 2.59
& 2.44 & 2.44 & 2.75 & 2.54
& 2.24 & 2.34 & 2.42 & 2.33 \\

Terminal-Traj
& 1.91 & 2.12 & 2.22 & 2.08
& 1.12 & 1.10 & 1.06 & 1.09
& 1.01 & 1.03 & 1.05 & 1.03
& 1.80 & 1.84 & 2.15 & 1.93 \\
\bottomrule
\end{tabular}
}
\end{table*}

As shown in Table~\ref{tab:env_quality}, \OURMETHOD~consistently achieves the highest quality across all four dimensions and all three judge models. In particular, the high scores on both \textit{environment-task consistency} and \textit{environment quality} indicate that the synthesized task instructions are well aligned with their executable environments, rather than being valid only in isolation. These results demonstrate that our skill-driven synthesis pipeline can construct higher-quality terminal tasks and environments with stronger task-environment alignment.

\appsection{Effect of Execution Guidelines} \label{app:guidelines}

\begin{wraptable}{r}{0.54\linewidth}
    \vspace{-2em}
    \centering
    \caption{\textbf{Effect of execution guidelines.} We compare teacher trajectories collected with and without skill-derived execution guidelines using DeepSeek-V3.2 on 500 sampled tasks.}
    \label{tab:guideline_ablation}
    \setlength\tabcolsep{3.5pt}
    \small
    \begin{tabular}{lcccc}
    \toprule
    \textbf{Setting} & \textbf{Success} & \textbf{$\Delta_{\text{\textbf{rel}}}^{\textbf{\%}}$} & \textbf{Avg. Steps} & \textbf{$\Delta_{\text{\textbf{rel}}}^{\textbf{\%}}$} \\
    \midrule
    \rowcolor{rowblue}
    w/ guideline & 39.6 & -- & 12.66 & -- \\
    w/o guideline & 27.6 & \textcolor{red}{-30.3\%} & 14.27 & \textcolor{red}{+12.7\%} \\
    \bottomrule
    \end{tabular}
    \vspace{-0.6em}
\end{wraptable}

To examine whether execution guidelines improve teacher trajectory collection, we randomly sample 500 tasks from \OURMETHOD{} and collect two sets of trajectories with the same DeepSeek-V3.2 teacher, holding all other settings fixed. In the \textit{w/ guideline} setting, the teacher receives the skill-derived execution guideline $G$ as input; in the \textit{w/o guideline} setting, the teacher must infer the solution path from the task alone. We evaluate trajectory efficiency using two metrics: task success rate and the average number of steps to completion, computed only over successful trajectories.

As shown in Table~\ref{tab:guideline_ablation}, execution guidelines substantially improve both reliability and efficiency. With guidelines, the teacher achieves a success rate of \textbf{39.6\%}, compared with \textbf{27.6\%} without guidelines, indicating a \textbf{30.3\%} relative drop when guidelines are removed. Moreover, successful trajectories require only \textbf{12.66} steps on average with guidelines, compared with \textbf{14.27} steps without guidelines.
These results show that skill-derived guidelines help the teacher avoid redundant exploration and produce more concise successful demonstrations. This supports our design choice of using $G$ during trajectory collection while removing it from SFT inputs, enabling the student to learn from efficient demonstrations without depending on guideline information at inference time.

\appsection{LLM-as-a-Judge Consistency Analysis}
\label{app:llm_judge_consistency}

Our pipeline relies on LLM-as-a-Judge at two distinct stages: task quality filtering (Sec.~\ref{sec:task_collection}) and evaluation on DABench and TableBench (Sec.~\ref{sec:exp}; Appendix~\ref{appendix:benchmarks}). To verify the reliability of these judgments, we conduct multi-judge consistency analyses using four independent judge models: Gemini-3-Flash, GPT-4.1, GLM-5, and Doubao-2.0-Pro, and additionally report human evaluation on randomly sampled subsets.

\paragraph{Task Quality Filtering.}
Table~\ref{tab:task_quality_filtering_agreement} reports inter-judge agreement for the five filtering criteria used in task quality filtering. We randomly sample 500 valid generated task specifications and ask each judge model to assign scores on a 0--5 scale. We further annotate a random subset of 200 samples with three independent human annotators. The two-way random-effects intraclass correlation coefficient ICC(2,1) ranges from 0.708 to 0.812 across criteria, with an overall ICC of 0.770, indicating substantial agreement among the four judge models. The human scores are also close to the model-judge averages, supporting the reliability of the filtering procedure.

\begin{table*}[h]
\centering
\small
\caption{
\textbf{Multi-judge agreement for task quality filtering.}
Each judge scores the five filtering criteria used in Sec.~3.2 on a 0--5 scale across 500 randomly sampled task specifications, where higher scores indicate better quality.
\textbf{Avg.} denotes the average score across the four judges.
\textbf{ICC(2,1)} reports the two-way random-effects absolute-agreement intraclass correlation coefficient among the four judges.
\textbf{Human} reports scores from three independent human annotators averaged over a random subset of 200 samples.
}
\label{tab:task_quality_filtering_agreement}
\setlength{\tabcolsep}{5pt}
\begin{tabular}{lcccccc}
\toprule
\multirow{2}{*}{\textbf{Filtering Criterion}}
& \multicolumn{4}{c}{\textbf{Judge Score}}
& \multirow{2}{*}{\textbf{Avg.}}
& \multirow{2}{*}{\textbf{ICC(2,1)}} \\
\cmidrule(lr){2-5}
& \textbf{Gemini-3-Flash} & \textbf{GPT-4.1} & \textbf{GLM-5} & \textbf{Doubao-2.0-Pro}
&  &  \\
\midrule
Instruction Quality
& 4.62 & 4.40 & 4.28 & 4.55 & 4.46 & 0.812 \\

Closed-World Solvability
& 4.38 & 4.18 & 4.05 & 4.30 & 4.23 & 0.784 \\

Blueprint Completeness
& 4.15 & 3.92 & 3.78 & 4.08 & 3.98 & 0.741 \\

Guideline Quality
& 4.05 & 3.80 & 3.65 & 3.95 & 3.86 & 0.708 \\

Evaluation Criteria Quality
& 4.52 & 4.30 & 4.18 & 4.42 & 4.36 & 0.795 \\

\midrule
\rowcolor{rowblue}
Overall
& \textbf{4.34} & \textbf{4.12} & \textbf{3.99} & \textbf{4.26}
& \textbf{4.18} & \textbf{0.770} \\
Human (200 samples)
& 4.27 & 4.13 & 3.94 & 4.21 & 4.14 & 0.763 \\
\bottomrule
\end{tabular}
\end{table*}

\paragraph{LLM-as-a-Judge on DABench and TableBench.}
Table~\ref{tab:dabench_tablebench_judge_agreement} reports per-judge Pass@1 scores and majority-vote results on the full 500-instance DABench and TableBench test sets (Sec.~\ref{sec:exp}; Appendix~\ref{appendix:benchmarks}) for three 32B models, together with human evaluation on a random subset of 200 instances per benchmark. Fleiss' $\kappa$ among the three audit judges (GPT-4.1, GLM-5, and Doubao-2.0-Pro) is 0.873 and 0.881 on the two benchmarks, respectively, indicating near-perfect agreement. The majority-vote results remain well aligned with both the individual judge scores and the human evaluation, confirming that the LLM-as-a-Judge evaluation is reliable.

\begin{table*}[h]
\centering
\small
\caption{
\textbf{Multi-judge agreement for LLM-as-a-Judge evaluation on DABench and TableBench.}
We report Pass@1 judging results from Gemini-3-Flash, GPT-4.1, GLM-5, and Doubao-2.0-Pro on three 32B models.
\textbf{Maj.} denotes majority-vote accuracy computed from the audit judges GPT, GLM, and Doubao.
\textbf{Avg. Maj.} averages the majority-vote accuracies on DABench and TableBench.
Fleiss' $\kappa$ among the audit judges is 0.873 for DABench and 0.881 for TableBench.
\textbf{Human} reports Pass@1 scores from three independent human annotators on a random subset of 200 samples per benchmark.
}
\label{tab:dabench_tablebench_judge_agreement}
\setlength{\tabcolsep}{3.5pt}
\resizebox{\textwidth}{!}{
\begin{tabular}{lcccccccccccccc}
\toprule
\multirow{2}{*}{\textbf{Model}}
& \multicolumn{6}{c}{\textbf{DABench}}
& \multicolumn{6}{c}{\textbf{TableBench}}
& \multirow{2}{*}{\textbf{Avg. Maj.}} \\
\cmidrule(lr){2-7}
\cmidrule(lr){8-13}
& \textbf{Gemini-3-Flash} & \textbf{GPT-4.1} & \textbf{GLM-5} & \textbf{Doubao-2.0-Pro} & \textbf{Maj.} & \textbf{Human}
& \textbf{Gemini-3-Flash} & \textbf{GPT-4.1} & \textbf{GLM-5} & \textbf{Doubao-2.0-Pro} & \textbf{Maj.} & \textbf{Human}
&  \\
\midrule
Qwen3-32B
& 44.4 & 42.0 & 43.0 & 44.6 & 43.4 & 42.0
& 26.4 & 24.0 & 25.0 & 26.0 & 25.2 & 23.5
& 34.3 \\

Nemotron-Terminal-32B
& 81.6 & 78.4 & 80.0 & 82.0 & 80.4 & 78.5
& \textbf{72.4} & \textbf{70.0} & \textbf{71.6} & \textbf{73.0} & \textbf{71.8} & \textbf{70.0}
& 76.1 \\

\rowcolor{rowblue}
\OURMETHOD-32B
& \textbf{83.6} & \textbf{81.0} & \textbf{82.4} & \textbf{84.0} & \textbf{82.8} & \textbf{81.0}
& 71.6 & 69.6 & 70.4 & 72.0 & 70.8 & 68.5
& \textbf{76.8} \\
\bottomrule
\end{tabular}
}
\end{table*}

\clearpage

\section{Prompt Templates}

\phantomsection
\refstepcounter{subsection}
\begin{mybox}{Prompt for Task Generation (Core Goal)}
\begin{lstlisting}[style=promptstyle]
You are creating realistic Linux-terminal tasks for training a terminal AI agent. Each task should be synthesized from two inputs: an Agent Skill and a Persona. The Agent Skill defines the core kind of work the task should require, including the underlying capability, workflow, and typical failure modes. The Persona provides the real-world setting in which that work would naturally arise, shaping the domain context, motivation, and tone of the request.

Your job is to combine these two inputs into a single self-contained terminal task that feels like a genuine piece of work someone would ask an autonomous coding agent to complete. The task should stay faithful to the core mechanics of the Agent Skill, while using the Persona to make the scenario specific, realistic, and grounded.

# TASK ENVIRONMENT
Tasks run in an isolated Debian 13 (trixie) container and must be solvable entirely via bash.
Pre-installed tools include:
- Python 3.12, pip 25 (*\textcolor{blue}{$\cdot$}*) Node.js 20, npm 10 (*\textcolor{blue}{$\cdot$}*) Java 8 (OpenJDK)
- gcc/g++ 14, make, git, curl, wget, tmux
- apt-get for additional packages
- Working directory: /app (subdirs: /output, /logs, /tests, /solution)

# INSTRUCTIONS

## 1. Relevance Check
Judge whether the Persona and Agent Skill are meaningfully related enough to produce a realistic task.
- If the pair is clearly mismatched, set `pair_relevance` to "unrelated", give a concrete reason, set `task_title` to "UNRELATED_PAIR", and leave all other content fields empty.
- If there is a plausible real-world connection, set `pair_relevance` to "related" and generate the full task below.

## 2. Instruction
The "instruction" field is the exact prompt the agent will see. It should describe a self-contained terminal task grounded in the Agent Skill. Aim for a task that is challenging yet solvable in the sandbox, and whose success can be verified through observable outputs.

## 3. Initial Files
Each entry in "initial_files" must include:
- (*\textcolor{blue}{generation\_mode}*): "llm\_direct" | "local\_tool" | "remote\_fetch"
- (*\textcolor{blue}{description}*): A complete reproduction spec -- file format, internal structure, scale, 2-3 concrete example values, and any deliberate anomalies the agent must handle.

## 4. Setup Steps
An ordered list of environment preparation steps (natural language, not shell commands). Use [] if no extra setup is needed.

## 5. Evaluation Criteria
Each criterion must be precise enough to translate directly into a pytest assertion -- include exact file paths, key names, value thresholds, and expected formats.

## 6. Output Format
Output STRICTLY as a JSON object. Do not include markdown fences or text outside the JSON.

# INPUTS
## Agent Skill
(*\textcolor{blue}{\{skill\}}*)

## Persona
(*\textcolor{blue}{\{persona\}}*)
\end{lstlisting}
\end{mybox}
\label{prompt:core_goal}

\bigskip

\phantomsection
\refstepcounter{subsection}
\begin{mybox}{Prompt for Guideline Generation}
\begin{lstlisting}[style=promptstyle]
You are generating an execution guideline for a terminal-agent SFT data synthesis pipeline. The guideline guides the agent through the task step-by-step.

# REQUIREMENTS

Each step must be:
- (*\textcolor{blue}{Actionable}*): include the specific command, file path, or edit
- (*\textcolor{blue}{Verifiable}*): say how to confirm it worked
- (*\textcolor{blue}{Ordered}*): respect task dependencies

Extract concrete steps from the Agent Skill's SOP when available. Prefix critical caveats with "IMPORTANT:" or "WARNING:". Do not leak final file content or complete solutions.

Each step should be written as a single string in this style:
"Step N: <action> -- <exact command or edit> -- <verification or warning>"

# GOOD EXAMPLES

- "Step 1: Benchmark current state -- Run 'du -sh public/images/' and 'find public/images -type f -name "*.png" | wc -l' to measure baseline."
- "Step 2: Update schema.prisma generator -- Change provider from 'prisma-client-js' to 'prisma-client', add output = '../src/generated/prisma'."
- "Step 3: Run 'bunx prisma generate' and verify generated client appears at packages/db/src/generated/prisma/."
- "Step 4: IMPORTANT: Verify @prisma/client remains in package.json dependencies (do NOT remove -- needed at runtime)."

Avoid vague steps like "Inspect the project" or "Fix any errors" that lack specific commands or targets.

# OUTPUT FORMAT

Return STRICTLY as a JSON object:
{"guideline": ["Step 1: ...", "Step 2: ..."]}

Do not include markdown fences, explanation, or any text outside the JSON.

# INPUTS

## Skill
(*\textcolor{blue}{\{skill\}}*)

## Core Goal
(*\textcolor{blue}{\{core\_goal\_json\}}*)
\end{lstlisting}
\end{mybox}
\label{prompt:guideline}

\bigskip

\phantomsection
\refstepcounter{subsection}
\begin{mybox}{Prompt for Task Quality Judgment}
\begin{lstlisting}[style=promptstyle]
You are an expert quality evaluator for bash terminal tasks. Your job is to evaluate the quality of the task specification across five dimensions (0-5 each) and justify each score.

# INPUT

## Persona (the user role that motivates this task)
(*\textcolor{blue}{\{persona\_text\}}*)

## Skill (the technical capability being exercised)
(*\textcolor{blue}{\{skill\_text\}}*)

## Generated Goal (the full task specification to evaluate)
(*\textcolor{blue}{\{goal\_json\}}*)

# IMPORTANT JUDGING PRINCIPLES

- Judge for realism, task quality, and training value -- not for rigid formatting.
- A concise, natural instruction can receive a high score.
- Do NOT require a fixed opener, numbered list, or a "Requirements:" section.
- High scores should reflect realistic task framing, clear task intent, strong skill alignment, solvability, and verifiability.

# EVALUATION DIMENSIONS

## Dim 1 -- Instruction Quality (instruction_quality)
Is the instruction realistic, skill-aligned, actionable, and clear enough for the agent to start the right kind of work?
  5 - Realistic, strongly skill-aligned, clear goal and success condition.
  4 - Strong and usable; slightly synthetic or less precise than ideal.
  3 - Executable but generic or partially under-specified.
  2 - Weakly aligned with the skill, artificial, or too vague.
  1 - Barely actionable or missing the task's core objective.
  0 - Self-contradictory, incoherent, or not a workable instruction.

## Dim 2 -- Solvable & Closed-World (solvable_closed_world)
Can the task be completed entirely inside an isolated container without internet access, private credentials, or privileged operations?
  5 - Strictly closed-world; all dependencies and data are in the blueprint.
  4 - Essentially closed-world; a few installable deps; no external data needed.
  3 - Likely closed-world; some unclear details can reasonably be filled in.
  2 - Significant external-dependency risk or critical data gaps.
  1 - Highly unsolvable -- critical inputs missing or strong external dependencies.
  0 - Definitively unsolvable or contradicts environment constraints.

## Dim 3 -- Blueprint Completeness (blueprint_completeness)
Is the environment blueprint specific enough to construct the environment deterministically?
  5 - All 5 categories complete: filesystem, data schema, deps, entrypoints, validation.
  4 - At least 4 of 5 categories covered; minor details missing.
  3 - Buildable but 1-2 of schema/versions/validation are missing.
  2 - Too abstract; builder must guess extensively.
  1 - Nearly unbuildable environment.
  0 - Empty blueprint or clearly inconsistent with the instruction.

## Dim 4 -- Guideline Quality (guideline_quality)
Does the guideline drive a realistic terminal trace with correct ordering, appropriate granularity, checkpoints, and SOP coverage -- without spoiling the final answer?
  5 - Captures skill workflow well; sensible ordering, useful checkpoints, no spoilers.
  4 - Strong and useful; slightly coarse or sparse on checkpoints.
  3 - Usable but generic; some workflow value, incomplete ordering or checkpoints.
  2 - Too abstract to guide execution, or too script-like and overprescriptive.
  1 - Does not match instruction/blueprint, or steps are disordered.
  0 - No guideline or completely unusable.

## Dim 5 -- Evaluation Criteria Quality (evaluation_criteria_quality)
Do the evaluation criteria clearly define how task completion will be assessed? Can they be translated into reliable black-box pytest checks without guessing?
  5 - Explicit, outcome-focused, concrete, and directly translatable to pytest.
  4 - Mostly strong; minor ambiguity but overall reliably judgeable.
  3 - Partially useful but incomplete or vague on at least one important condition.
  2 - Weak or underspecified; difficult to translate into stable assertions.
  1 - Mostly unusable; major conditions missing, vague, or subjective.
  0 - Absent, contradictory, or cannot assess task completion at all.

# OUTPUT FORMAT

Respond with ONLY a JSON object -- no markdown fences, no explanation outside the JSON.

{
  "instruction_quality":         {"score": <0-5>, "reason": "..."},
  "solvable_closed_world":       {"score": <0-5>, "reason": "..."},
  "blueprint_completeness":      {"score": <0-5>, "reason": "..."},
  "guideline_quality":           {"score": <0-5>, "reason": "..."},
  "evaluation_criteria_quality": {"score": <0-5>, "reason": "..."}
}
\end{lstlisting}
\end{mybox}
\label{prompt:judge}

\bigskip

\phantomsection
\refstepcounter{subsection}
\begin{mybox}{Prompt for File Generation (\texttt{llm\_direct} mode)}
\begin{lstlisting}[style=promptstyle]
You are an expert environment builder for AI coding agent benchmarks.

Given a task description, environment blueprint, and a target file to generate,
create the actual file content for that specific file.

# Task Instruction
(*\textcolor{blue}{\{instruction\}}*)

# Environment Blueprint
(*\textcolor{blue}{\{blueprint\}}*)

# Target File to Generate
(*\textcolor{blue}{\{target\_file\}}*)

# Previously Generated Files
(*\textcolor{blue}{\{previous\_files\}}*)

# Instructions
Generate the content for the target file based on:
1. The file's description and filepath
2. The task instruction (what the agent needs to accomplish)
3. The overall environment context (system requirements, pre-setup)
4. Previously generated files (maintain consistency with imports, APIs, etc.)

You MUST end your response with a ```json fenced code block -- this block is required.
The JSON object must have exactly two fields:
- "filepath": the exact path from the target file
- "content": the full file content as a string
\end{lstlisting}
\end{mybox}
\label{prompt:file_direct}

\bigskip

\phantomsection
\refstepcounter{subsection}
\begin{mybox}{Prompt for File Generation (\texttt{local\_tool} mode)}
\begin{lstlisting}[style=promptstyle]
[System]
You are a specialized local artifact generation agent running inside a Linux sandbox.

You must respond with exactly one required function call and nothing else.

Available tool:
  python: {"target_filepath": string, "code": string, "timeout_sec": integer (optional)}

Rules:
- Only use the python tool.
- target_filepath must exactly match the requested target path.
- Use Python to create or repair the target artifact.
- Keep working until the tool observation reports a valid artifact.

[User]
Target file: (*\textcolor{blue}{\{target\_filepath\}}*)
Description: (*\textcolor{blue}{\{file\_description\}}*)

Task context (for reference only -- do NOT solve the whole task, only create the target artifact):
(*\textcolor{blue}{\{instruction\_summary\}}*)
\end{lstlisting}
\end{mybox}
\label{prompt:file_local_tool}

\bigskip

\phantomsection
\refstepcounter{subsection}
\begin{mybox}{Prompt for File Generation (\texttt{remote\_fetch} mode)}
\begin{lstlisting}[style=promptstyle]
[System]
You are a specialized stateless remote fetch agent running inside a Linux sandbox.

You must respond with exactly one required function call and nothing else.

Available tools:
  web_search:    {"query": string, "top_k": integer (optional), "domain_hint": string (optional)}
  fetch_page:    {"url": string, "mode": "http"|"dynamic"|"stealth", "timeout_ms": integer (optional)}
  download_file: {"url": string, "save_as": string, "timeout_ms": integer (optional)}

Rules:
- Stateless: do not assume browser state or session reuse between actions.
- save_as must exactly match the requested target path.
- Use search, fetch, and download as separate steps.
- Keep working until the downloaded artifact validates successfully.

[User]
Target file: (*\textcolor{blue}{\{target\_filepath\}}*)
Description: (*\textcolor{blue}{\{file\_description\}}*)

Task context (for reference only -- do NOT solve the whole task, only fetch the target artifact):
(*\textcolor{blue}{\{instruction\_summary\}}*)
\end{lstlisting}
\end{mybox}
\label{prompt:file_remote_fetch}

\bigskip

\phantomsection
\refstepcounter{subsection}
\begin{mybox}{Prompt for File Verification}
\begin{lstlisting}[style=promptstyle]
[System]
You are a file verifier agent running inside a Linux sandbox.

Your role is verification-only.

Rules:
- You are not a task-solving agent.
- Your scope is limited to the declared filepaths and their file specifications.
- Use shell commands only to inspect the provided files.
- Do NOT modify files, use network access, or evaluate runtime state.
- Prefer status="continue" whenever evidence is incomplete.
- Respond with exactly one JSON object per turn.
- Continue mode must use:  analysis, status="continue", commands
- Finalize mode must use:  analysis, status="finalize", result
- Finalize result keys:    overall_verdict, file_findings, global_findings
- overall_verdict must be exactly "pass" or "fail"
- Each item in file_findings:  {filepath, reason, repair_instructions}
- Each item in global_findings: {reason, primary_owner, repair_instructions}
  (primary_owner: "llm_direct" | "specialized" | "unattributed")
- If overall_verdict is "pass", both findings arrays must be empty.
- If overall_verdict is "fail", at least one finding must be present.
- repair_instructions must be concrete and actionable for the repair agent.

Valid response examples:

  Continue:
  {
    "analysis": "Need to check file content.",
    "status": "continue",
    "commands": ["cat /app/data.csv"]
  }

  Finalize (pass):
  {
    "analysis": "All files match their specifications.",
    "status": "finalize",
    "result": {"overall_verdict": "pass", "file_findings": [], "global_findings": []}
  }

  Finalize (fail):
  {
    "analysis": "Found a mismatch in /app/example.txt.",
    "status": "finalize",
    "result": {
      "overall_verdict": "fail",
      "file_findings": [{
        "filepath": "/app/example.txt",
        "reason": "Content does not match the declared format.",
        "repair_instructions": "Rewrite /app/example.txt to match the declared format."
      }],
      "global_findings": []
    }
  }

[User]
Task instruction:
(*\textcolor{blue}{\{instruction\}}*)

Verification scope:
- Only verify the declared filepaths below.
- Ignore runtime state and broader task completion.

Files to verify:
(*\textcolor{blue}{\{file\_lines\}}*)

File-only verification objective:
- Verify each declared file against its description and the task instruction.
- Only use global_findings for still-file-scoped cross-file issues.
- If you fail any file or emit any global finding, include detailed repair_instructions.

Initial workspace tree (initial hint only):
(*\textcolor{blue}{\{workspace\_tree\}}*)
\end{lstlisting}
\end{mybox}
\label{prompt:file_verify}

\bigskip

\phantomsection
\refstepcounter{subsection}
\begin{mybox}{Prompt for Environment Setup (env build)}
\begin{lstlisting}[style=promptstyle]
You are an expert DevOps engineer setting up a Linux sandbox environment for an AI coding agent. Your job is ONLY to prepare the environment -- install system packages, language runtimes, and libraries. You must NOT write application code or execute the task itself.

# Base Environment (already provisioned -- do NOT reinstall these)
- OS: Debian 13 (trixie), kernel 6.1.x, x86_64
- User: user (uid=1000), non-root. Use sudo for privileged operations.
- Working directory: /app
- Pre-installed: Python 3.12, pip 25.x, Node.js 20.x, npm 10.x, Java 8 (OpenJDK), gcc/g++ 14.x, make 4.x, git 2.x, curl 8.x, wget 1.x
- Harbor directories (perms=777): /logs/agent, /logs/verifier, /tests, /app, /output, /solution

# Task Instruction (for context only -- do NOT execute this task)
(*\textcolor{blue}{\{instruction\}}*)

# Environment Blueprint
(*\textcolor{blue}{\{blueprint\}}*)

# Pre-seeded Assets
The following assets have already been written into the sandbox. Reference them directly (e.g. `pip install -r /app/requirements.txt`). Do NOT recreate or overwrite them.

(*\textcolor{blue}{\{pre\_seeded\_files\}}*)

# Instructions
Generate shell commands to ONLY:
1. Execute each step listed in setup_steps, in order
2. Create necessary directories
3. Install from any dependency manifest in the pre-seeded assets
4. Start or configure required services, environment variables, and permissions

IMPORTANT:
- HIGHEST PRIORITY: Never create, download, or overwrite any pre-seeded asset path
- Scope is limited to environment configuration only -- no application source code
- Do NOT execute the task (e.g. do NOT run `node index.js` or `python main.py`)
- When using sudo with network commands, ALWAYS use `sudo -E` to preserve proxy env vars

End your response with a ```bash fenced code block containing the commands.
\end{lstlisting}
\end{mybox}
\label{prompt:env_build}

\bigskip

\phantomsection
\refstepcounter{subsection}
\begin{mybox}{Prompt for Environment Verification (env verify)}
\begin{lstlisting}[style=promptstyle]
You are an expert DevOps engineer verifying that a Linux sandbox environment has been correctly set up. A setup script has already been executed successfully (exit code 0). Your job is to generate verification commands that check whether the environment is truly ready for the task.

# Base Environment (already provisioned)
- OS: Debian 13 (trixie), kernel 6.1.x, x86_64; Working directory: /app
- Pre-installed: Python 3.12, Node.js 20.x, Java 8, gcc/g++ 14.x, make, git, curl, wget

# Task Instruction (for context -- do NOT execute this task)
(*\textcolor{blue}{\{instruction\}}*)

# Environment Blueprint
(*\textcolor{blue}{\{blueprint\}}*)

# Pre-seeded Assets
(*\textcolor{blue}{\{pre\_seeded\_files\}}*)

# Setup Script That Was Executed
(*\textcolor{blue}{\{setup\_script\}}*)

# Instructions
Generate shell commands that ONLY verify the environment is ready:
1. Check that required packages/libraries are importable (e.g. `python3 -c "import flask"`, `node -e "require('express')"`)
2. Check that required CLI tools are available (e.g. `which gcc`, `java -version`)
3. Check that required directories exist
4. Check that dependency versions meet requirements if specific versions were requested

IMPORTANT:
- Do NOT install anything -- only verify
- Do NOT execute the task itself or write application code
- The script does NOT use `set -e` -- all commands run even if some fail

End your response with a ```bash fenced code block containing the verification commands.
\end{lstlisting}
\end{mybox}
\label{prompt:env_verify}

\bigskip

\phantomsection
\refstepcounter{subsection}
\begin{mybox}{Prompt for Environment Repair (env repair)}
\begin{lstlisting}[style=promptstyle]
You are an expert DevOps engineer fixing a failed sandbox environment setup. Your job is ONLY to prepare the environment -- install packages and configure services. You must NOT write application code or execute the task.

# Base Environment (already provisioned -- do NOT reinstall these)
- OS: Debian 13 (trixie), kernel 6.1.x, x86_64; Working directory: /app
- Pre-installed: Python 3.12, Node.js 20.x, Java 8, gcc/g++ 14.x, make, git, curl, wget

# Task Instruction (for context only -- do NOT execute this task)
(*\textcolor{blue}{\{instruction\}}*)

# Environment Blueprint
(*\textcolor{blue}{\{blueprint\}}*)

# Pre-seeded Assets
The following assets will be re-written before your commands run. Do NOT recreate them.

(*\textcolor{blue}{\{pre\_seeded\_files\}}*)

# Previous Setup Commands
(*\textcolor{blue}{\{commands\}}*)

# Command Failure Details
(*\textcolor{blue}{\{errors\}}*)

# Instructions
The previous attempt failed. The next attempt runs in a FRESH sandbox -- do NOT rely on any side effects from the previous attempt.

Generate a FULL corrected list of setup commands (not a patch/delta). All commands will be combined into a single bash script with `set -euxo pipefail` and executed once.

IMPORTANT:
- HIGHEST PRIORITY: Never create, download, or overwrite any pre-seeded asset path
- Scope is limited to environment configuration only -- no application source code
- Do NOT execute the task (e.g. do NOT run `node index.js` or `python main.py`)
- When using sudo with network commands, ALWAYS use `sudo -E` to preserve proxy env vars

End your response with a ```bash fenced code block containing the full corrected commands.
\end{lstlisting}
\end{mybox}
\label{prompt:env_repair}

\bigskip

\phantomsection
\refstepcounter{subsection}
\begin{mybox}{Prompt for Pytest Verifier Generation}
\begin{lstlisting}[style=promptstyle]
You are generating a pytest-based task verifier for a terminal benchmark task.

Your verifier will run AFTER the agent has finished executing the task, inside the SAME sandbox state. Do not assume the environment is restarted. Verify ONLY whether the task was actually completed -- do not inspect the agent's messages or tool traces.

You must output a JSON object with this exact schema:
{
  "system_packages": ["..."],
  "python_packages": ["..."],
  "helper_files": [{"path": "tests/filename.ext", "content": "..."}],
  "test_outputs_py": "..."
}

Rules:
1. The verifier must use pytest.
2. `test_outputs_py` must contain valid Python source code.
3. Tests must be black-box.
4. Prefer verifying via files, command behavior, local HTTP behavior, or deterministic end-to-end examples.
5. Tests must be deterministic and self-contained.
6. Avoid network access unless the task explicitly requires localhost access.
7. Only add packages required by the verifier itself.
8. Put any extra test assets into `helper_files`.
9. Do not generate explanations outside the JSON.

# Instruction
(*\textcolor{blue}{\{instruction\}}*)

# Evaluation Criteria
(*\textcolor{blue}{\{evaluation\_criteria\}}*)

# Target Output File
(*\textcolor{blue}{\{target\_output\_file\}}*)

# Initial Text Files (ACTUAL CONTENT -- already present in environment)
IMPORTANT: These files already exist when the agent starts. DO NOT test for them. They are provided as context only.

(*\textcolor{blue}{\{initial\_text\_files\}}*)

# Initial Asset Files (METADATA ONLY -- already present in environment)
IMPORTANT: These files exist but only metadata is provided. Do NOT assume actual content.

(*\textcolor{blue}{\{initial\_asset\_files\}}*)

# Validated Environment File Paths
(*\textcolor{blue}{\{validated\_filepaths\}}*)

CRITICAL RULES:
- DO NOT test whether initial files exist (guaranteed to exist)
- DO NOT test initial file content (already validated)
- ONLY test whether the evaluation criteria are met
- Test what the AGENT creates, not what the environment provides
\end{lstlisting}
\end{mybox}
\label{prompt:pytest_generation}

\bigskip

\phantomsection
\refstepcounter{subsection}
\begin{mybox}{System Prompt for Terminus2 Agent (JSON Format)}
\begin{lstlisting}[style=promptstyle]
You are an AI assistant tasked with solving command-line tasks in a Linux environment. You will be given a task description and the output from previously executed commands. Your goal is to solve the task by providing batches of shell commands.

Format your response as JSON with the following structure:

{
  "analysis": "Analyze the current state based on the terminal output. What has been accomplished? What still needs to be done?",
  "plan": "Describe your plan for the next steps. What commands will you run and why? Be specific about what you expect each command to accomplish.",
  "commands": [
    {"keystrokes": "ls -la\n",     "duration": 0.1},
    {"keystrokes": "cd project\n", "duration": 0.1}
  ],
  "task_complete": true
}

Required fields:
- "analysis":  Your analysis of the current situation
- "plan":      Your plan for the next steps
- "commands":  Array of command objects to execute

Optional fields:
- "task_complete": Boolean indicating if the task is complete (defaults to false)

Command object structure:
- "keystrokes": Exact keystrokes to send to the terminal (required, must end with \n)
- "duration":   Seconds to wait before executing the next command (default: 1.0)

IMPORTANT: The text inside "keystrokes" is sent verbatim to the terminal:
- Every command must end with \n or it will not execute.
- Special key sequences use tmux-style escape sequences: C-c for Ctrl+C, C-d for Ctrl+D.

The "duration" attribute controls how long to wait for output before proceeding:
- Immediate commands (cd, ls, echo, cat):  0.1 s
- Standard commands (gcc, find, rustc):    1.0 s
- Slow commands (make, wget, long scripts): set an appropriate value as needed
- Prefer shorter durations; poll by sending {"keystrokes": "", "duration": 10.0} rather than blocking.

Important notes:
- Send keystrokes exactly as written; do not add extra whitespace unless intended.
- Extra text outside the JSON generates warnings but is tolerated.
- The JSON must be valid -- escape quotes and special characters inside strings.
- An empty "commands" array is valid when you need to observe without acting.

Task Description:
(*\textcolor{blue}{\{instruction\}}*)

Current terminal state:
(*\textcolor{blue}{\{terminal\_state\}}*)
\end{lstlisting}
\end{mybox}
\label{prompt:terminus2}

\bigskip

\phantomsection
\refstepcounter{subsection}
\begin{mybox}{Prompt for Environment Quality Evaluation}
\begin{lstlisting}[style=promptstyle]
You are evaluating a Harbor-format terminal benchmark task.

Task directory: (*\textcolor{blue}{\{task\_dir\}}*)

Please read the task contents (instruction.md, environment/ files,
tests/test_outputs.py or tests/test_final_state.py) and score it on
four dimensions. Each dimension is scored 1-3:
  1 = poor   2 = acceptable   3 = good

Important: base your scores ONLY on the file contents you read.
Ignore the directory path and any dataset name it may imply --
treat every task as anonymous.

Scoring dimensions:

1. terminal_nativeness
   Does the task genuinely require terminal CLI operations (compilers,
   package managers, system commands, build tools, network tools)?
   High score = real CLI toolchain required.
   Low score  = just writing files or trivial echo commands.

2. env_task_consistency
   Do the pre-placed environment files and setup precisely match what
   the instruction requires?
   High score = environment provides exactly the right scaffolding,
                no excess or gap.
   Low score  = environment is empty, irrelevant, or contradicts
                the task.

3. env_quality
   Are the environment files (Dockerfile, setup.sh, initial files)
   well-formed and credible?
   High score = files are realistic, complete, and executable.
   Low score  = files are fabricated stubs, have obvious bugs, or
                are missing critical dependencies.

4. verifier_robustness
   Do the pytest assertions in tests/ accurately distinguish
   task-complete from task-incomplete?
   High score = assertions are specific, cover all key acceptance
                criteria from the instruction, low false-positive risk.
   Low score  = only checks file existence, or assertions are
                unrelated to the instruction requirements.

Important: if a dimension is not applicable (e.g. no tests/ directory
exists), score it 1.

Respond with ONLY a JSON object, no other text:
{
  "terminal_nativeness":         <1|2|3>,
  "terminal_nativeness_reason":  "<one sentence>",
  "env_task_consistency":        <1|2|3>,
  "env_task_consistency_reason": "<one sentence>",
  "env_quality":                 <1|2|3>,
  "env_quality_reason":          "<one sentence>",
  "verifier_robustness":         <1|2|3>,
  "verifier_robustness_reason":  "<one sentence>"
}
\end{lstlisting}
\end{mybox}
\label{prompt:env_quality}


\definecolor{DExBlueDark}{HTML}{1A3A5C}
\definecolor{DExBlueMid}{HTML}{2E6DA4}
\definecolor{DExBlueBg}{HTML}{EAF2FB}
\definecolor{DExGreenDark}{HTML}{1D6A3A}
\definecolor{DExGreenBg}{HTML}{E6F4EC}
\definecolor{DExOrangeDark}{HTML}{8B4500}
\definecolor{DExOrangeBg}{HTML}{FDF3E7}
\definecolor{DExPurpleDark}{HTML}{5B2C8D}
\definecolor{DExPurpleBg}{HTML}{F5EEF8}
\definecolor{DExGrayBg}{HTML}{F4F6F7}
\definecolor{DExTealDark}{HTML}{0E6655}
\definecolor{DExTermBg}{HTML}{1E1E2E}
\definecolor{DExTermFg}{HTML}{CDD6F4}
\definecolor{DExTermGreen}{HTML}{A6E3A1}
\definecolor{DExTermYellow}{HTML}{F9E2AF}
\definecolor{DExTermRed}{HTML}{F38BA8}
\definecolor{DExPassGreen}{HTML}{27AE60}

\newtcolorbox{casewrapper}[2][]{%
  enhanced, breakable,
  colback=DExBlueBg,
  colframe=DExBlueDark,
  coltitle=white,
  fonttitle=\bfseries\normalsize,
  attach boxed title to top left={yshift=-2.8mm, xshift=4mm},
  boxed title style={colback=DExBlueDark, sharp corners, boxrule=0pt},
  sharp corners=south,
  rounded corners=north,
  boxrule=1pt,
  top=5mm, left=4mm, right=4mm, bottom=4mm,
  title={#2},
  #1
}

\newtcolorbox{moduleblock}[2][]{%
  enhanced, breakable,
  colback=white,
  colframe=#1,
  coltitle=white,
  fonttitle=\bfseries\small,
  boxed title style={colback=#1, sharp corners, boxrule=0pt},
  attach boxed title to top left={yshift=-2.5mm, xshift=3mm},
  sharp corners,
  boxrule=0.6pt,
  top=4mm, left=3.5mm, right=3.5mm, bottom=3mm,
  title={\strut #2}
}

\newtcolorbox{fileblock}[2][DExTealDark]{%
  enhanced, breakable,
  colback=white,
  colframe=#1,
  coltitle=white,
  fonttitle=\bfseries\footnotesize\ttfamily,
  boxed title style={colback=#1, sharp corners, boxrule=0pt},
  attach boxed title to top left={yshift=-2mm, xshift=3mm},
  sharp corners,
  boxrule=0.4pt,
  top=3.5mm, left=3mm, right=3mm, bottom=2.5mm,
  fontupper=\small,
  title={\strut #2}
}

\makeatletter
\@ifundefined{lst@style@dexpycode}{%
  \lstdefinestyle{dexpycode}{%
    language=Python,
    basicstyle=\ttfamily\scriptsize,
    breaklines=true,
    keepspaces=true,
    showstringspaces=false,
    frame=single,
    numbers=none,
    keywordstyle=\color{DExBlueDark}\bfseries,
    commentstyle=\color{DExTealDark},
    stringstyle=\color{DExPurpleDark},
    rulecolor=\color{DExBlueDark!30!white},
    framesep=4pt,
    columns=fullflexible,
  }%
}{}
\makeatother

\lstdefinestyle{terminalstyle}{%
  basicstyle=\ttfamily\scriptsize\color{DExTermFg},
  backgroundcolor=\color{DExTermBg},
  breaklines=true,
  frame=none,
  numbers=none,
  keepspaces=true,
  columns=fullflexible,
  showstringspaces=false,
  moredelim=[is][\color{DExTermGreen}]{<G>}{</G>},
  moredelim=[is][\color{DExTermYellow}]{<Y>}{</Y>},
  moredelim=[is][\color{DExTermRed}]{<R>}{</R>},
  literate={\ }{{ }}{1},
}

\makeatletter
\@ifundefined{lst@style@dexrawcode}{%
  \lstdefinestyle{dexrawcode}{%
    basicstyle=\ttfamily\scriptsize,
    breaklines=true,
    keepspaces=true,
    showstringspaces=false,
    frame=single,
    numbers=none,
    rulecolor=\color{DExBlueDark!30!white},
    framesep=4pt,
    columns=fullflexible,
    literate={~}{{\textasciitilde}}{1},
  }%
}{}
\makeatother

\makeatletter
\@ifundefined{lst@style@dexshcode}{%
  \lstdefinestyle{dexshcode}{%
    language=bash,
    basicstyle=\ttfamily\scriptsize,
    breaklines=true,
    keepspaces=true,
    showstringspaces=false,
    frame=single,
    numbers=none,
    keywordstyle=\color{DExBlueDark}\bfseries,
    commentstyle=\color{DExTealDark}\itshape,
    stringstyle=\color{DExPurpleDark},
    rulecolor=\color{DExGreenDark!30!white},
    framesep=4pt,
    columns=fullflexible,
  }%
}{}
\makeatother

\makeatletter
\@ifundefined{kvtag}{%
\newcommand{\kvtag}[2]{%
  \colorbox{DExGrayBg}{\texttt{\textcolor{DExBlueMid}{#1}}\texttt{:~}\texttt{\textcolor{DExTealDark}{#2}}}%
}}{}
\makeatother

\makeatletter
\@ifundefined{filebadge}{%
\newcommand{\filebadge}[2]{%
  \colorbox{#1!18!white}{\textcolor{#1!80!black}{\texttt{\small #2}}}%
}}{}
\makeatother

\newcommand{\DExCheck}{\textcolor{DExGreenDark}{\ding{51}}}
\newcommand{\DExPass}{\textcolor{DExPassGreen}{\ding{51}}}

\clearpage

\section{Data Examples}\label{sec:appendix-data-examples}

To complement the textual pipeline description in Section~\ref{sec:method}, this section presents \emph{three stage-focused examples} drawn from the \textbf{Terminal-World} dataset. Rather than compressing an end-to-end rollout for every task, each example zooms into a single synthesis stage and displays its inputs and outputs verbatim so the reader can see the concrete shape of each artifact:

\begin{enumerate}[leftmargin=1.8em, itemsep=3pt]
  \item \textbf{Example~\ref{sec:example-task-gen}} illustrates \textbf{Task Generation} (\S\ref{sec:task_collection}): from a single (Skill $S$, Persona $U$) pair to the synthesized quadruple $(\mathcal{I}, \mathcal{E}, \mathcal{V}, \mathcal{G})$. We show all six artifacts verbatim.
  \item \textbf{Example~\ref{sec:example-env-build}} illustrates \textbf{Environment Building} (\S\ref{sec:env_construction}): a three-file blueprint is routed through three different sub-agents of the multi-agent GVR architecture, producing the initial files $F$, setup script $B_\text{env}$, and pytest verifier $T_\text{test}$.
  \item \textbf{Example~\ref{sec:example-traj}} illustrates \textbf{Trajectory Collection} (\S\ref{sec:traj_collection}): a multi-turn teacher-model rollout with verbatim \texttt{analysis}/\texttt{plan}/\texttt{commands}/\texttt{observation} for four representative steps.
\end{enumerate}

\appsection{Example 1: Task Generation --- ELF Binary Parsing (Astrophysics)}\label{sec:example-task-gen}

\begin{casewrapper}{%
  \faCode\enspace
  Stage-Focused Example 1 \texttt{|} Stage: \textbf{Task Generation} (\S\ref{sec:task_collection})}

\begin{moduleblock}[DExBlueDark]{%
  \raisebox{0.1ex}{\ding{182}}\enspace Input $S$ --- Agent Skill}

\begin{lstlisting}[style=dexrawcode]
---
name: elf-binary-analysis
description: >
  Analyze ELF binary files for reverse engineering, security research, and
  exploitation. Use this skill whenever the user needs to understand ELF
  structure, analyze program headers, section headers, symbols, relocations,
  GOT/PLT, or identify binary protections like RELRO, stack canaries, and PIE.
  Trigger on any request involving ELF files, binary analysis, readelf output
  interpretation, or exploitation reconnaissance.
tags: ['reverse-engineering', 'binary-analysis', 'elf']
source: github.com/abelrguezr/hacktricks-skills
task_slug: binary-exploitation/elf-tricks
route_label: segment-buffer-allocation
---
\end{lstlisting}
\end{moduleblock}

\bigskip

\begin{moduleblock}[DExBlueDark]{%
  \raisebox{0.1ex}{\ding{183}}\enspace Input $U$ --- User Persona (FinePersonas)}

\begin{lstlisting}[style=dexrawcode]
An astrophysicist or a science writer covering recent developments in
astrophysics, particularly focused on gamma-ray bursts, gravitational
waves, and astronomical research.

labels: [Science, Academia, Space Exploration]
\end{lstlisting}
\end{moduleblock}

\bigskip

\begin{moduleblock}[DExGreenDark]{%
  \raisebox{0.1ex}{\ding{184}}\enspace Output $\mathcal{I}$ --- Task Instruction}

\begin{tcolorbox}[colback=DExBlueBg, colframe=DExBlueMid!60!white, boxrule=0.5pt,
  sharp corners, left=3mm, right=3mm, top=2mm, bottom=2mm, fontupper=\small, breakable]
As an astrophysicist, I have a compiled binary tool \texttt{grb\_processor.elf} used for gamma-ray burst data analysis. To validate its memory mapping and extract calibration constants, write a Node.js script that parses the ELF file. Your script should identify the \texttt{PT\_LOAD} segment with \texttt{RW} flags (the data and BSS region), allocate a buffer with zero-padding for the BSS section, and extract 32-bit integer values from all 4-byte aligned virtual addresses within that segment. Output a JSON object where keys are virtual addresses as decimal strings and values are the 32-bit integers. Save the JSON to \texttt{/app/memory\_map.json}.
\end{tcolorbox}
\end{moduleblock}

\bigskip

\begin{moduleblock}[DExGreenDark]{%
  \raisebox{0.1ex}{\ding{185}}\enspace Output $\mathcal{E}$ --- Environment Blueprint}

\begin{fileblock}{Initial Files: /app/grb\_processor.elf \quad\textnormal{\textbullet}\quad mode: local\_tool}
ELF 64-bit LSB executable binary file. To generate, compile the following C program with \texttt{gcc -o /app/grb\_processor.elf /tmp/source.c}. C source code:
\begin{lstlisting}[style=dexrawcode]
#include <stdint.h>
int calibration = 271828;          // Euler's number * 100000
int flux_readings[255] = {0};      // 255 x 4 = 1020 bytes of initialized data
char photon_buffer[4096];          // uninitialized BSS (zero-padded at load time)
int main() { return 0; }
\end{lstlisting}
After compilation, \texttt{readelf -l} shows two LOAD segments: LOAD[0] (text, \texttt{R E}) with \texttt{VirtAddr=0x400000}, \texttt{FileSiz=0x238}, \texttt{MemSiz=0x238} --- code only, not used for extraction; LOAD[1] (data+BSS, \texttt{RW}) with \texttt{VirtAddr=0x601000}, \texttt{FileSiz=0x400}, \texttt{MemSiz=0x1400}. Extraction targets only the \texttt{RW} LOAD segment: MemSiz \texttt{0x1400} = 5120 bytes, yielding 1280 key-value pairs (5120/4=1280). \texttt{calibration} occupies the first 4 bytes of LOAD[1] at \texttt{VirtAddr=0x601000}.
\end{fileblock}

\medskip

\begin{fileblock}[DExOrangeDark]{Setup Steps}
\begin{itemize}[leftmargin=1.2em, itemsep=1pt, topsep=2pt]
  \item Install \texttt{binutils} package via \texttt{apt-get} to ensure \texttt{readelf} command is available for ELF inspection.
\end{itemize}
\end{fileblock}
\end{moduleblock}

\bigskip

\begin{moduleblock}[DExGreenDark]{%
  \raisebox{0.1ex}{\ding{186}}\enspace Output $\mathcal{V}$ --- Evaluation Scope}

\begin{itemize}[leftmargin=1.4em, itemsep=2pt, topsep=2pt]
  \item[\DExCheck] File \texttt{/app/memory\_map.json} exists and contains valid JSON.
  \item[\DExCheck] JSON object has exactly \textbf{1280} key-value pairs, matching total MemSiz \texttt{0x1400} bytes divided by 4.
  \item[\DExCheck] Key \texttt{'6295552'} (decimal for virtual address \texttt{0x601000}, start of the RW LOAD segment) has value \textbf{271828}.
  \item[\DExCheck] All keys in the JSON are strings that can be parsed as integers, and all values are numbers.
\end{itemize}
\end{moduleblock}

\bigskip

\begin{moduleblock}[DExGreenDark]{%
  \raisebox{0.1ex}{\ding{187}}\enspace Output $\mathcal{G}$ --- Execution Guideline}

\begin{enumerate}[leftmargin=1.9em, itemsep=2pt, topsep=2pt, label=\small Step \arabic*.]
  \item Install \texttt{binutils} for \texttt{readelf} command --- Run \texttt{apt-get update \&\& apt-get install -y binutils} --- Verify with \texttt{readelf --version}.
  \item Inspect ELF header to confirm binary properties --- Execute \texttt{readelf -h /app/grb\_processor.elf} --- Confirm output shows \texttt{Class:~ELF64} and little-endian.
  \item Identify the data LOAD segment --- Run \texttt{readelf -l /app/grb\_processor.elf} --- Locate the \texttt{PT\_LOAD} entry with \texttt{RW} flags; note its VirtAddr (\texttt{0x601000}), FileSiz (\texttt{0x400}), and MemSiz (\texttt{0x1400}); confirm two LOAD segments exist.
  \item Verify Node.js runtime --- Run \texttt{node --version} --- Ensure version is v14.x or higher for BigInt support.
  \item \textbf{[IMPORTANT]} Create Node.js script with BSS handling --- Write \texttt{/app/extract.js} implementing buffer allocation based on MemSiz and 4-byte aligned reads --- Verify file creation with \texttt{ls -l /app/extract.js}.
  \item Run extraction script --- \texttt{node /app/extract.js /app/grb\_processor.elf > /app/memory\_map.json 2>\&1} --- Check exit code with \texttt{echo \$?}; it should be 0.
  \item Validate JSON output format --- \texttt{head -c 200 /app/memory\_map.json} --- Confirm it starts with \texttt{\{} and appears as valid JSON.
  \item Final verification --- use a Node.js one-liner to parse the JSON, verify key count is 1280 and value at \texttt{'6295552'} (VirtAddr \texttt{0x601000}) is 271828.
\end{enumerate}
\end{moduleblock}

\end{casewrapper}

\appsection{Example 2: Environment Building --- Multi-Format Data Merger}\label{sec:example-env-build}

\begin{casewrapper}{%
  \faCode\enspace
  Stage-Focused Example 2 \texttt{|} Stage: \textbf{Environment Building} (\S\ref{sec:env_construction})}

\begin{moduleblock}[DExBlueDark]{%
  \raisebox{0.1ex}{\ding{182}}\enspace Input --- Environment Blueprint $\mathcal{E}_\text{files}$}

\begin{fileblock}{Initial Files: (1) /app/quiz\_data.json \quad\textnormal{\textbullet}\quad mode: llm\_direct}
JSON array of objects with keys: \texttt{student\_id} (int), \texttt{topic} (string, either \texttt{'algebra'} or \texttt{'scientific\_notation'}), \texttt{score} (float or null), \texttt{timestamp} (ISO8601 string). 50 objects. Example objects:
\begin{lstlisting}[style=dexrawcode]
{"student_id": 101, "topic": "algebra",             "score": 88.5, ...}
{"student_id": 102, "topic": "scientific_notation", "score": 85.5, ...}
\end{lstlisting}
Student\_id 103 has a null score for algebra (missing data). Student\_id 101 appears twice with different scores for algebra (88.5) and scientific\_notation (92.0) --- agent must map these to separate fields.
\end{fileblock}

\medskip

\begin{fileblock}{Initial Files: (2) /app/gradebook.csv \quad\textnormal{\textbullet}\quad mode: llm\_direct}
CSV file with columns: \texttt{id} (int), \texttt{name} (string), \texttt{algebra\_grade} (float), \texttt{sci\_not\_grade} (float), \texttt{updated} (ISO8601 string). 50 rows. Example rows:
\begin{lstlisting}[style=dexrawcode]
101,Alice Johnson,90.0,88.5,2024-10-17T09:00:00
102,Bob Smith,76.5,89.0,2024-10-17T09:05:00
\end{lstlisting}
Row for \texttt{id 104} has \texttt{algebra\_grade=-1.0} (sentinel for missing data). Column \texttt{id} corresponds to \texttt{student\_id} in other sources.
\end{fileblock}

\medskip

\begin{fileblock}{Initial Files: (3) /app/project\_scores.parquet \quad\textnormal{\textbullet}\quad mode: local\_tool}
Parquet file with columns: \texttt{student\_id} (int), \texttt{project\_type} (string, \texttt{'algebra'} or \texttt{'scientific\_notation'}), \texttt{project\_score} (int or string), \texttt{date} (string in YYYY-MM-DD format). 30 rows. Example: \texttt{student\_id=105, project\_type='algebra', project\_score=95, date='2024-10-18'}. Three rows have \texttt{project\_score} as string \texttt{'N/A'} instead of integer (type inconsistency to handle). Scores are out of 100. The final merged output must store all score columns (\texttt{algebra\_score}, \texttt{sci\_not\_score}) as \texttt{float64} dtype.
\end{fileblock}

\medskip

\begin{fileblock}[DExOrangeDark]{Setup Steps}
\begin{itemize}[leftmargin=1.2em, itemsep=1pt, topsep=2pt]
  \item Install \texttt{pandas} and \texttt{pyarrow} via \texttt{pip} to handle CSV, JSON, and Parquet file operations.
\end{itemize}
\end{fileblock}
\end{moduleblock}

\bigskip

\begin{moduleblock}[DExGreenDark]{%
  \raisebox{0.1ex}{\ding{183}}\enspace Artifact $F$ --- Generated Initial Files (head of each)}

\noindent\textbf{\small Routed to $\mathcal{A}_{llm\_direct}$ --- \texttt{(1) /app/quiz\_data.json}}:
\begin{lstlisting}[style=dexrawcode]
[
  {"student_id": 101, "topic": "algebra",             "score": 88.5, "timestamp": "2024-10-15T10:30:00"},
  {"student_id": 101, "topic": "scientific_notation", "score": 92.0, "timestamp": "2024-10-15T11:45:00"},
  {"student_id": 102, "topic": "scientific_notation", "score": 85.5, "timestamp": "2024-10-16T11:00:00"},
  {"student_id": 103, "topic": "algebra",             "score": null, "timestamp": "2024-10-16T11:15:00"},
  {"student_id": 104, "topic": "algebra",             "score": 77.0, "timestamp": "2024-10-15T09:00:00"},
  {"student_id": 104, "topic": "scientific_notation", "score": 82.5, "timestamp": "2024-10-15T10:00:00"},
  ... (44 more entries)
]
\end{lstlisting}

\medskip
\noindent\textbf{\small Routed to $\mathcal{A}_{llm\_direct}$ --- \texttt{(2) /app/gradebook.csv}}:
\begin{lstlisting}[style=dexrawcode]
id,name,algebra_grade,sci_not_grade,updated
101,Alice Johnson,90.0,88.5,2024-10-17T09:00:00
102,Bob Smith,76.5,89.0,2024-10-17T09:05:00
103,Charlie Brown,82.0,81.5,2024-10-17T09:10:00
104,David Williams,-1.0,90.0,2024-10-17T09:15:00
105,Eva Martinez,91.0,89.5,2024-10-17T09:20:00
106,Frank Miller,85.0,84.0,2024-10-17T09:25:00
... (44 more rows)
149,Will Morgan,82.0,79.5,2024-10-17T12:35:00
150,Zoe Carter,88.5,91.0,2024-10-17T12:40:00
\end{lstlisting}

\medskip
\noindent\textbf{\small Routed to $\mathcal{A}_{local\_tool}$ --- \texttt{(3) /app/project\_scores.parquet}}:
\begin{lstlisting}[style=dexrawcode]
>>> import pandas as pd; df = pd.read_parquet('/app/project_scores.parquet')
>>> df.dtypes
student_id        int64
project_type       str
project_score      str      <- type inconsistency: some rows hold 'N/A'
date               str
>>> df.head(8)
 student_id         project_type  project_score         date
        101              algebra             70   2024-10-16
        102  scientific_notation             81   2024-10-13
        103              algebra             95   2024-10-22
        104  scientific_notation             91   2024-10-24
        105              algebra             95   2024-10-18
        106  scientific_notation            N/A   2024-10-17
        107              algebra             94   2024-10-22
        108  scientific_notation             86   2024-10-14
>>> df.shape
(30, 4)
\end{lstlisting}
\end{moduleblock}

\bigskip

\begin{moduleblock}[DExGreenDark]{%
  \raisebox{0.1ex}{\ding{184}}\enspace Artifact $B_\text{env}$ --- Generated \texttt{setup.sh}}

\begin{lstlisting}[style=dexshcode]
#!/usr/bin/env bash
set -euxo pipefail
cd /app

sudo -E pip install pandas pyarrow
\end{lstlisting}
\end{moduleblock}

\bigskip

\begin{moduleblock}[DExGreenDark]{%
  \raisebox{0.1ex}{\ding{185}}\enspace Artifact $T_\text{test}$ --- Generated pytest Verifier}

\begin{lstlisting}[style=dexpycode,
  basicstyle=\ttfamily\tiny,
  breaklines=true, breakatwhitespace=false]
import pandas as pd, json, os, pytest
import numpy as np

def test_output_files_exist():
    assert os.path.exists("/app/merged_students.parquet")
    assert os.path.exists("/app/conflicts.json")

def test_merged_parquet_structure():
    df = pd.read_parquet("/app/merged_students.parquet")
    expected_cols = ['student_id', 'name', 'algebra_score', 'sci_not_score']
    assert set(df.columns) == set(expected_cols)
    assert pd.api.types.is_integer_dtype(df['student_id'])
    assert pd.api.types.is_float_dtype(df['algebra_score'])
    assert pd.api.types.is_float_dtype(df['sci_not_score'])
    assert df['student_id'].dtype == 'int64'

def test_row_count():
    quiz      = pd.read_json   ("/app/quiz_data.json")
    gradebook = pd.read_csv    ("/app/gradebook.csv")
    project   = pd.read_parquet("/app/project_scores.parquet")
    all_ids   = set(quiz['student_id']) | set(gradebook['id']) | set(project['student_id'])
    df        = pd.read_parquet("/app/merged_students.parquet")
    assert len(df) == len(all_ids)
    assert set(df['student_id']) == all_ids

def test_priority_and_data_logic():
    df = pd.read_parquet("/app/merged_students.parquet")
    # Student 101: Quiz(alg=88.5, sci=92.0) beats Gradebook(alg=90.0, sci=88.5) by priority.
    s101 = df[df['student_id'] == 101].iloc[0]
    assert s101['algebra_score'] == 88.5
    assert s101['sci_not_score'] == 92.0
    # Student 103: Quiz(alg=null) -> fallback to Gradebook(alg=82.0).
    s103 = df[df['student_id'] == 103].iloc[0]
    assert s103['algebra_score'] == 82.0
    # Student 104: Quiz(alg=77.0) beats Gradebook(alg=-1.0 sentinel).
    s104 = df[df['student_id'] == 104].iloc[0]
    assert s104['algebra_score'] == 77.0
    assert s101['name'] == "Alice Johnson"
    # Student only in Gradebook.
    s149 = df[df['student_id'] == 149].iloc[0]
    assert s149['name'] == "Will Morgan"
    assert s149['algebra_score'] == 82.0

def test_conflicts_json_content():
    with open("/app/conflicts.json", 'r') as f:
        data = json.load(f)
    assert "total_conflicts" in data
    assert "conflicts"       in data
    assert isinstance(data['conflicts'], list)
    # Student 101 had a conflict on algebra_score (Quiz 88.5 vs Gradebook 90.0).
    for conflict in data['conflicts']:
        if conflict['student_id'] == 101 and conflict['field'] == 'algebra_score':
            assert conflict['selected_value'] == 88.5
            assert 'quiz'      in conflict['values']
            assert 'gradebook' in conflict['values']
    # General priority check: selected_value must match highest-priority non-missing source.
    for conflict in data['conflicts']:
        vals, selected = conflict['values'], conflict['selected_value']
        priority_val = None
        for src in ['quiz', 'gradebook', 'project']:
            val = vals.get(src)
            if val is not None and val != -1.0 and val != 'N/A' \
               and not (isinstance(val, float) and np.isnan(val)):
                priority_val = val
                break
        if priority_val is not None:
            assert float(selected) == float(priority_val)
\end{lstlisting}
\end{moduleblock}

\end{casewrapper}

\appsection{Example 3: Trajectory Collection --- Video OCR Extraction}\label{sec:example-traj}

\begin{casewrapper}{%
  \faCode\enspace
  Stage-Focused Example 3 \texttt{|} Stage: \textbf{Trajectory Collection} (\S\ref{sec:traj_collection})}

\begin{center}
\includegraphics[width=\linewidth]{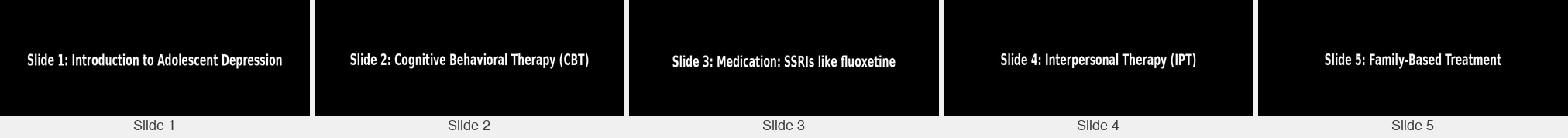}
\end{center}

\bigskip

\begin{moduleblock}[DExOrangeDark]{%
  \raisebox{0.1ex}{\ding{182}}\enspace Step 2 (of 11) --- Install \texttt{ffmpeg}}

\noindent\textbf{\footnotesize analysis}\,\ \ {\footnotesize We're at /app with a fresh terminal. The task is to process a video file using OCR to extract treatment methods and save a cleaned list. A checklist is provided with 10 steps. We need to start with step 1: update package lists.}

\noindent\textbf{\footnotesize plan}\,\ \ {\footnotesize Execute step 1: run apt-get update to update package lists. Then proceed to step 2: install ffmpeg.}

\noindent\textbf{\footnotesize commands}
\begin{lstlisting}[style=dexshcode]
apt-get update
apt-get install -y ffmpeg
\end{lstlisting}

\noindent\textbf{\footnotesize observation}
\begin{tcolorbox}[colback=DExTermBg, colframe=DExOrangeDark!70!black, boxrule=0.5pt,
  sharp corners, left=3mm, right=3mm, top=2.5mm, bottom=2.5mm,
  fontupper=\color{DExTermFg}\ttfamily\scriptsize, breakable]
/app \$ apt-get update\\
Hit:1 http://mirrors.tuna.tsinghua.edu.cn/debian trixie InRelease\\
Reading package lists... Done\\
/app \$ apt-get install -y ffmpeg\\
Reading package lists... Done\\
Building dependency tree... Done\\
\textcolor{DExTermGreen}{ffmpeg is already the newest version (7:7.1.3-0+deb13u1).}\\
0 upgraded, 0 newly installed, 0 to remove and 79 not upgraded.
\end{tcolorbox}
\end{moduleblock}

\bigskip

\begin{moduleblock}[DExOrangeDark]{%
  \raisebox{0.1ex}{\ding{183}}\enspace Step 6 (of 11) --- Extract video frames with \texttt{ffmpeg}}

\noindent\textbf{\footnotesize analysis}\,\ \ {\footnotesize All prerequisites installed. Proceed to step 6: extract video frames. Create /app/frames and run ffmpeg at \texttt{fps=1}.}

\noindent\textbf{\footnotesize plan}\,\ \ {\footnotesize \texttt{mkdir -p /app/frames}, then ffmpeg with \texttt{fps=1}. Verify with \texttt{find ... | wc -l}.}

\noindent\textbf{\footnotesize commands}
\begin{lstlisting}[style=dexshcode]
mkdir -p /app/frames
ffmpeg -i /app/education_video.mp4 -vf "fps=1" /app/frames/frame_%04d.png
find /app/frames -name "*.png" | wc -l
\end{lstlisting}

\noindent\textbf{\footnotesize observation}
\begin{tcolorbox}[colback=DExTermBg, colframe=DExOrangeDark!70!black, boxrule=0.5pt,
  sharp corners, left=3mm, right=3mm, top=2.5mm, bottom=2.5mm,
  fontupper=\color{DExTermFg}\ttfamily\scriptsize, breakable]
/app \$ ffmpeg -i /app/education\_video.mp4 -vf "fps=1" /app/frames/frame\_\%04d.png\\
ffmpeg version 7.1.3-0+deb13u1 Copyright (c) 2000-2025 the FFmpeg developers\\
\textcolor{DExTermYellow}{\ldots}\\
Input \#0, mov,mp4,m4a,3gp,3g2,mj2, from '/app/education\_video.mp4':\\
\ \ Duration: 00:00:15.00, start: 0.000000, bitrate: 21 kb/s\\
\ \ Stream \#0:0: Video: h264 (High) (avc1/0x31637661), yuv420p, 1280x720, 20 kb/s, 1 fps\\
Stream mapping:\\
\ \ Stream \#0:0 -> \#0:0 (h264 (native) -> png (native))\\
Output \#0, image2, to '/app/frames/frame\_\%04d.png':\\
\ \ Stream \#0:0: Video: png, rgb24, 1280x720, q=2-31, 1 fps\\
\textcolor{DExTermGreen}{frame=\ \ 15 fps=0.0 q=-0.0 Lsize=N/A time=00:00:15.00 bitrate=N/A speed= 157x}\\
/app \$ find /app/frames -name "*.png" | wc -l\\
\textcolor{DExTermGreen}{15}
\end{tcolorbox}
\end{moduleblock}

\bigskip

\begin{moduleblock}[DExOrangeDark]{%
  \raisebox{0.1ex}{\ding{184}}\enspace Step 7 (of 11) --- Author \texttt{ocr.py}, run OCR}

\noindent\textbf{\footnotesize analysis}\,\ \ {\footnotesize Frame extraction successful: 15 frames. Now step 7: create OCR script and run it.}

\noindent\textbf{\footnotesize plan}\,\ \ {\footnotesize Create \texttt{/app/ocr.py} that globs frames, runs \texttt{pytesseract} on each, and appends to \texttt{all\_text.txt}.}

\noindent\textbf{\footnotesize commands}
\begin{lstlisting}[style=dexpycode]
import cv2, pytesseract, glob

frames_dir  = '/app/frames'
output_file = '/app/all_text.txt'

frame_paths = sorted(glob.glob(f'{frames_dir}/frame_*.png'))
all_text = []
for path in frame_paths:
    img = cv2.imread(path)
    if img is None:
        continue
    text = pytesseract.image_to_string(img)
    all_text.append(text)

with open(output_file, 'w') as f:
    f.write('\n'.join(all_text))
\end{lstlisting}

\noindent\textbf{\footnotesize observation}
\begin{tcolorbox}[colback=DExTermBg, colframe=DExOrangeDark!70!black, boxrule=0.5pt,
  sharp corners, left=3mm, right=3mm, top=2.5mm, bottom=2.5mm,
  fontupper=\color{DExTermFg}\ttfamily\scriptsize, breakable]
/app \$ python3 /app/ocr.py\\
/app \$ wc -l /app/all\_text.txt\\
\textcolor{DExTermGreen}{29 /app/all\_text.txt}\\
/app \$ head -c 200 /app/all\_text.txt\\
\textcolor{DExTermGreen}{Slide 1: Introduction to Adolescent Depression}\\
\textcolor{DExTermGreen}{Slide 1: Introduction to Adolescent Depression}\\
\textcolor{DExTermGreen}{Slide 2: Cognitive Behavioral Therapy (CBT)}\\
\textcolor{DExTermGreen}{Slide 2: Co}\ldots
\end{tcolorbox}
\end{moduleblock}

\bigskip

\begin{moduleblock}[DExOrangeDark]{%
  \raisebox{0.1ex}{\ding{185}}\enspace Step 9 (of 11) --- Refine \texttt{parse.py} to filter out the intro slide}

\noindent\textbf{\footnotesize analysis}\,\ \ {\footnotesize Step~8 produced an initial parse with 5 unique lines, including the non-treatment \textit{Introduction to Adolescent Depression}. Filter by treatment-related keywords.}

\noindent\textbf{\footnotesize plan}\,\ \ {\footnotesize Rewrite \texttt{/app/parse.py} with a keyword whitelist [\texttt{therapy, treatment, medication, cbt, ssri, antidepressant}], dedup, re-run, verify $\geq 4$ lines and CBT presence.}

\noindent\textbf{\footnotesize commands}
\begin{lstlisting}[style=dexpycode]
input_file  = '/app/all_text.txt'
output_file = '/app/treatment_list.txt'
keywords    = ['therapy', 'treatment', 'medication', 'cbt', 'ssri', 'antidepressant']

with open(input_file, 'r') as f:
    lines = f.read().splitlines()

treatments = set()
for line in lines:
    line = line.strip()
    if not line:
        continue
    if line.startswith('Slide'):
        parts = line.split(':', 1)
        if len(parts) == 2:
            treatment = parts[1].strip()
            if treatment and any(kw in treatment.lower() for kw in keywords):
                treatments.add(treatment)

with open(output_file, 'w') as f:
    for t in sorted(treatments):
        f.write(t + '\n')
\end{lstlisting}

\noindent\textbf{\footnotesize observation}
\begin{tcolorbox}[colback=DExTermBg, colframe=DExOrangeDark!70!black, boxrule=0.5pt,
  sharp corners, left=3mm, right=3mm, top=2.5mm, bottom=2.5mm,
  fontupper=\color{DExTermFg}\ttfamily\scriptsize, breakable]
/app \$ python3 /app/parse.py\\
\textcolor{DExTermGreen}{Extracted 4 unique treatments.}\\
/app \$ cat /app/treatment\_list.txt\\
\textcolor{DExTermGreen}{Cognitive Behavioral Therapy (CBT)}\\
\textcolor{DExTermGreen}{Family-Based Treatment}\\
\textcolor{DExTermGreen}{Interpersonal Therapy (IPT)}\\
\textcolor{DExTermGreen}{Medication: SSRIs like fluoxetine}\\
/app \$ wc -l /app/treatment\_list.txt\\
\textcolor{DExTermGreen}{4 /app/treatment\_list.txt}\\
/app \$ grep -q "Cognitive Behavioral Therapy\textbackslash|CBT" /app/treatment\_list.txt \&\& echo "Contains CBT"\\
\textcolor{DExTermGreen}{Contains CBT}
\end{tcolorbox}
\end{moduleblock}

\bigskip

\begin{moduleblock}[DExPurpleDark]{%
  \raisebox{0.1ex}{\ding{186}}\enspace Verification Result (downstream pytest)}

{\footnotesize\setlength{\tabcolsep}{4pt}%
\begin{tabular}{@{}llll@{}}
\toprule
\textbf{Test Case} & \textbf{Status} & \textbf{Metric} & \textbf{Value}\\
\midrule
\texttt{test\_treatment\_list\_exists}      & \textcolor{DExPassGreen}{\DExPass\ PASSED} & file present       & \texttt{/app/treatment\_list.txt}\\
\texttt{test\_treatment\_list\_min\_lines}  & \textcolor{DExPassGreen}{\DExPass\ PASSED} & line count         & $\geq 4$ (actual: 4)\\
\texttt{test\_cbt\_present}                 & \textcolor{DExPassGreen}{\DExPass\ PASSED} & CBT keyword        & found\\
\texttt{test\_ssri\_present}                & \textcolor{DExPassGreen}{\DExPass\ PASSED} & SSRI keyword       & found\\
\texttt{test\_ipt\_present}                 & \textcolor{DExPassGreen}{\DExPass\ PASSED} & IPT keyword        & found\\
\texttt{test\_family\_based\_present}       & \textcolor{DExPassGreen}{\DExPass\ PASSED} & FBT keyword        & found\\
\midrule
\multicolumn{2}{@{}l}{\textbf{Reward}} & \multicolumn{2}{l}{\textbf{1.0} \quad (6/6 passed, 11 steps)}\\
\bottomrule
\end{tabular}}
\end{moduleblock}

\end{casewrapper}

\end{document}